\documentclass[10pt, twocolumn]{elsarticle}
\usepackage{verbatim}
\usepackage{subfigure}
\usepackage{float}
\usepackage{amsmath}
\usepackage{graphicx}
\usepackage{amssymb}
\usepackage{amsfonts}
\usepackage{subfigure}
\usepackage{algorithm}
\usepackage{algpseudocode}
\usepackage{times}
\usepackage{fullpage}
\usepackage{placeins}
\usepackage[figuresright]{rotating}
\usepackage{epstopdf}
\usepackage{bbm}
\usepackage{xfrac}
\usepackage{multicol}
\usepackage{color}
\usepackage{nicefrac}
\usepackage{multicol,lipsum}
\usepackage{geometry}
\usepackage{wrapfig}
\usepackage{romannum}
\graphicspath{{./figures/}}

\usepackage{lineno,hyperref}
\modulolinenumbers[5]

\journal{Journal of Structural Biology}

\biboptions{authoryear}

\begin{document}
\pagenumbering{arabic}
\begin{frontmatter}
\title{APPLE Picker: Automatic Particle Picking, a Low-Effort  Cryo-EM Framework}
\author[add1]{Ayelet Heimowitz}
\ead{aheimowitz@math.princeton.edu}
\author[2]{Joakim And\'{e}n}
\ead{janden@flatironinstitute.org}
\author[add1,add3]{Amit Singer}
\ead{amits@math.princeton.edu}

\address[add1]{The Program in Applied and Computational Mathematics, Princeton University, Princeton, NJ}
\address[2]{Center for Computational Biology, Flatiron Institute, New York, NY}
\address[add3]{Department of Mathematics, Princeton University, Princeton, NJ}
\onecolumn
\begin{abstract}
Particle picking is a crucial first step in the computational pipeline of single-particle cryo-electron microscopy (cryo-EM). Selecting particles from the micrographs is difficult especially for small particles with low contrast. As high-resolution reconstruction typically requires hundreds of thousands of particles, manually picking that many particles is often too time-consuming.  While semi-automated particle picking is currently a popular approach, it may suffer from introducing manual bias into the selection process. In addition, semi-automated particle picking is still somewhat time-consuming.
This paper presents the APPLE (\textbf{A}utomatic \textbf{P}article \textbf{P}icking with \textbf{L}ow user \textbf{E}ffort) picker,  a simple and novel approach for fast, accurate, and fully automatic particle picking. While our approach was inspired by template matching, it is completely template-free. 
This approach is evaluated on publicly available datasets 
containing micrographs of $\beta$-Galactosidase, T20S proteasome, 70S ribosome and keyhole limpet hemocyanin projections.
\end{abstract}

\begin{keyword}
cryo-electron microscopy, single-particle reconstruction, particle picking, template-free, cross-correlation, micrographs, support vector machines.
\end{keyword}

\end{frontmatter}
\begin{multicols}{2}
\section{Introduction}
\label{sec:introduction}

\begin{figure*}[t]
\centering
{\includegraphics[width=0.25\linewidth]{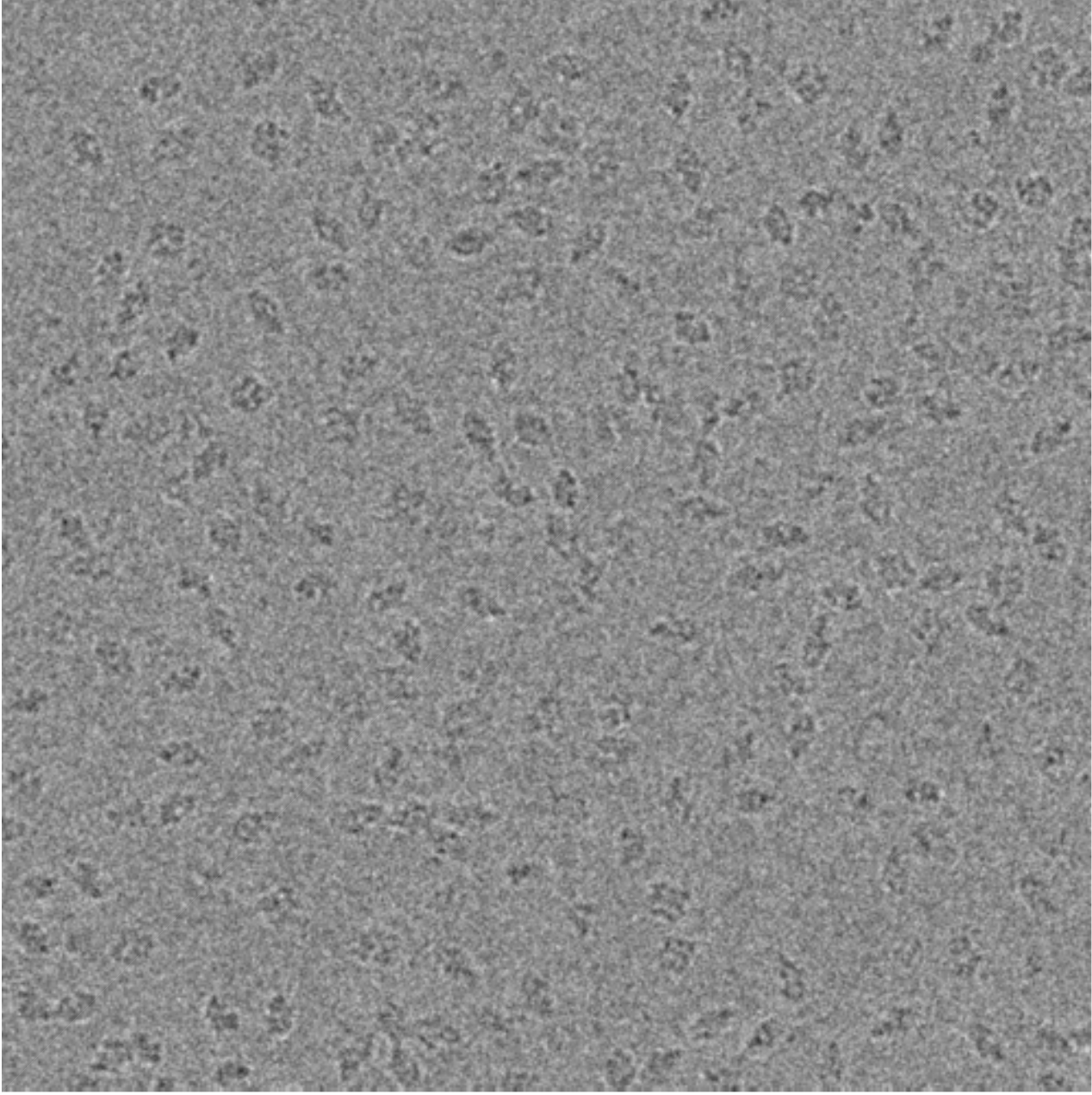}}
{\includegraphics[width=0.25\linewidth]{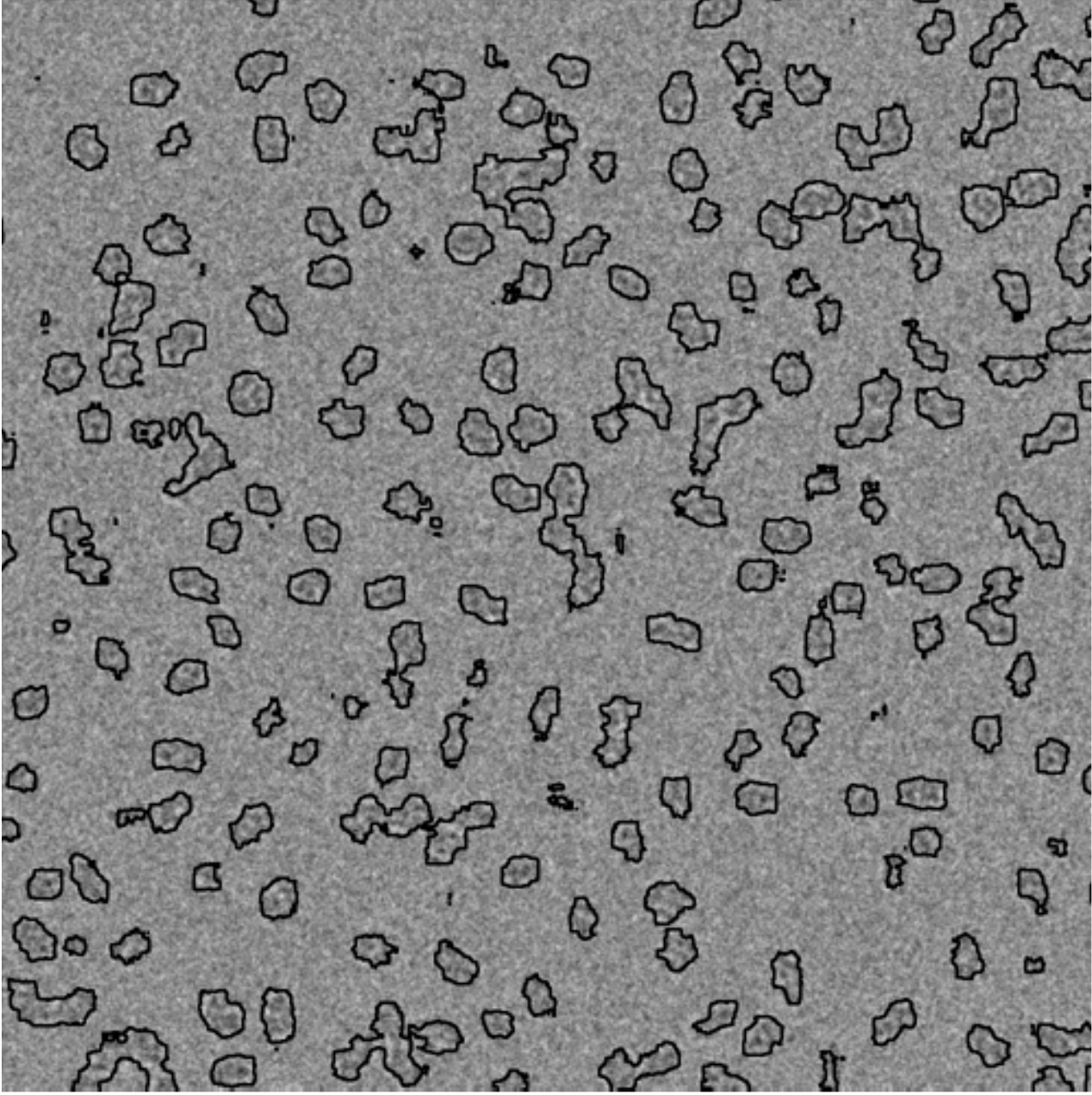}}
{\includegraphics[width=0.25\linewidth]{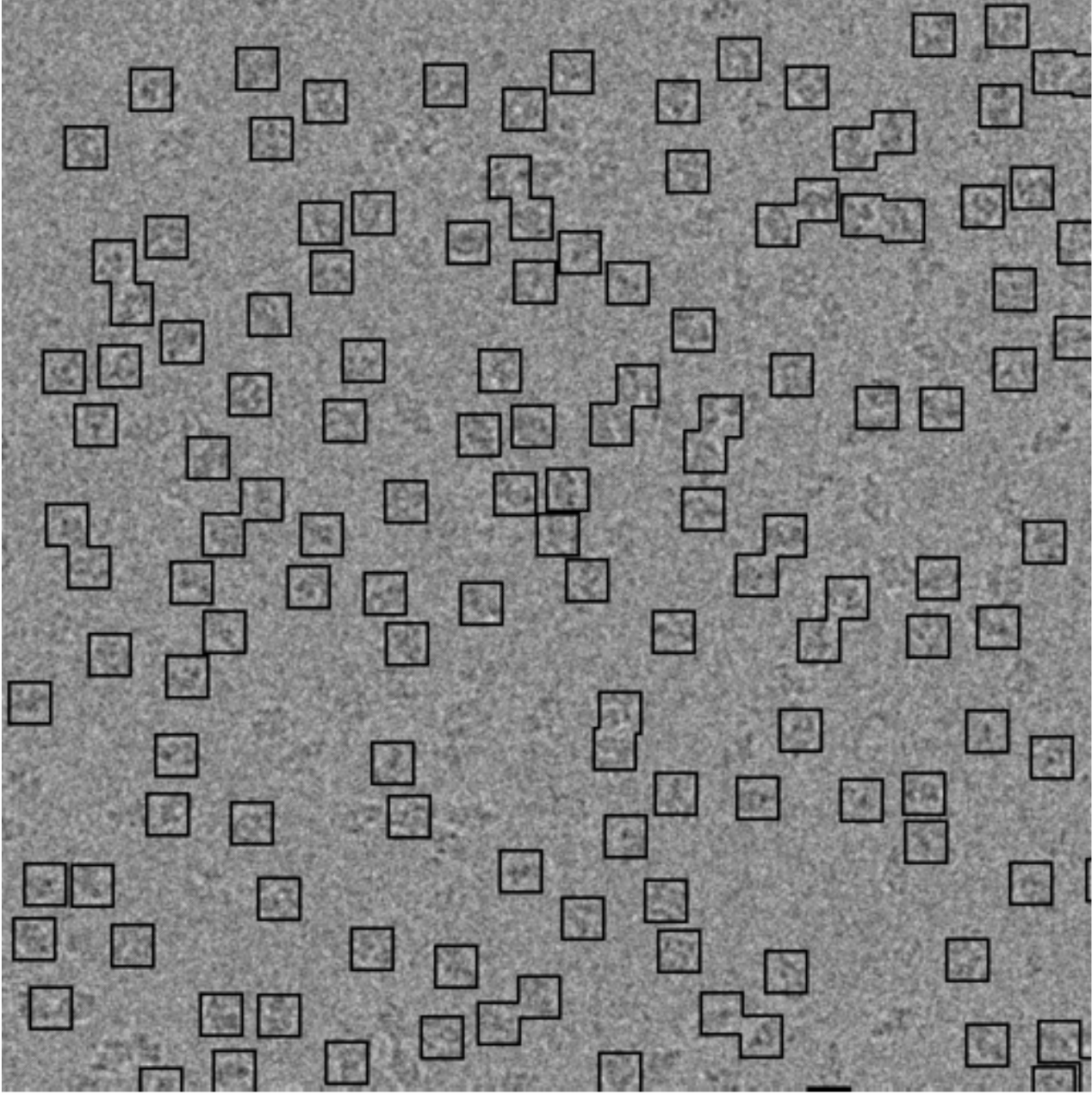}}\\
\subfigure[]{\label{subfig:micro}{\includegraphics[width=0.25\linewidth]{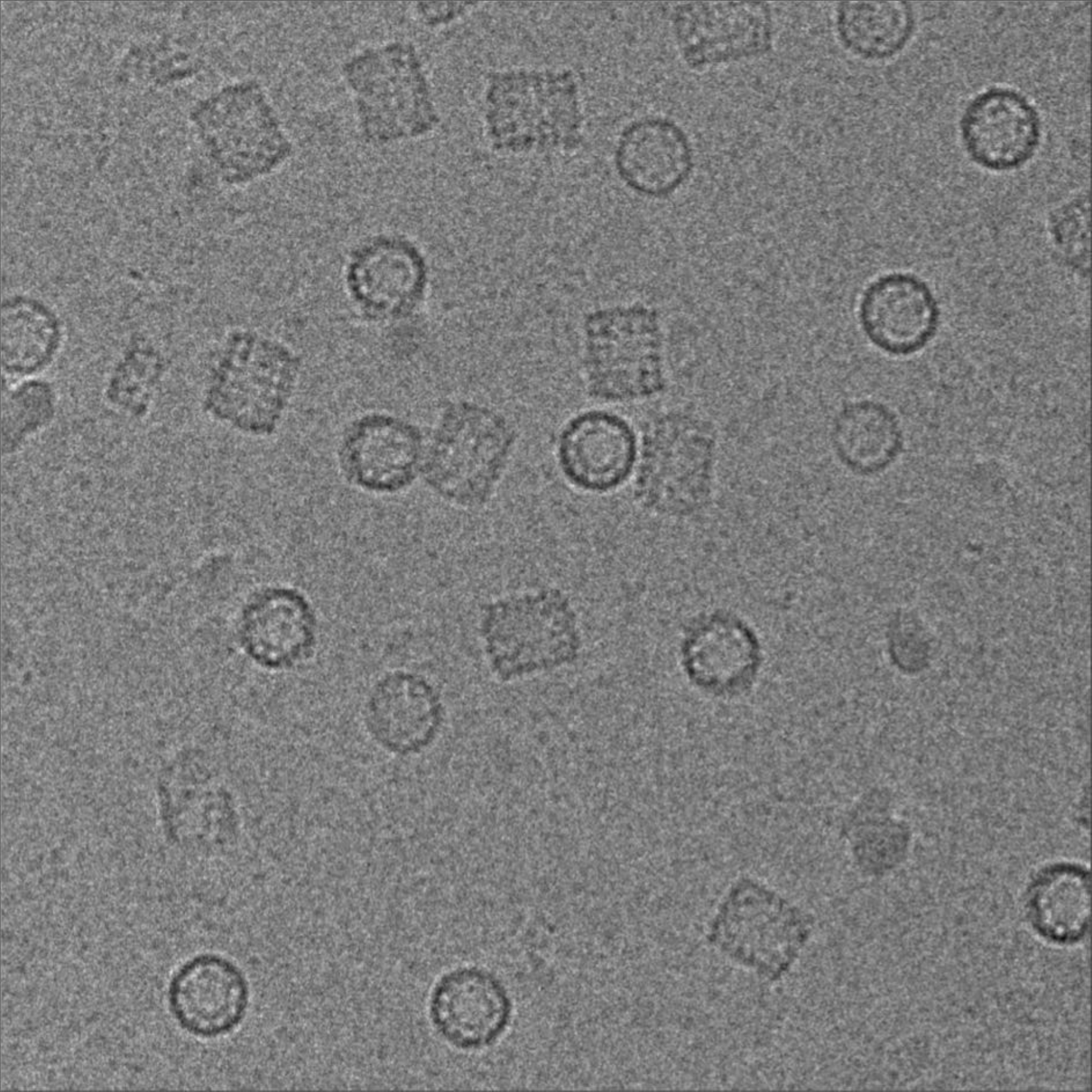}}}
\subfigure[]{\label{subfig:particle}{\includegraphics[width=0.25\linewidth]{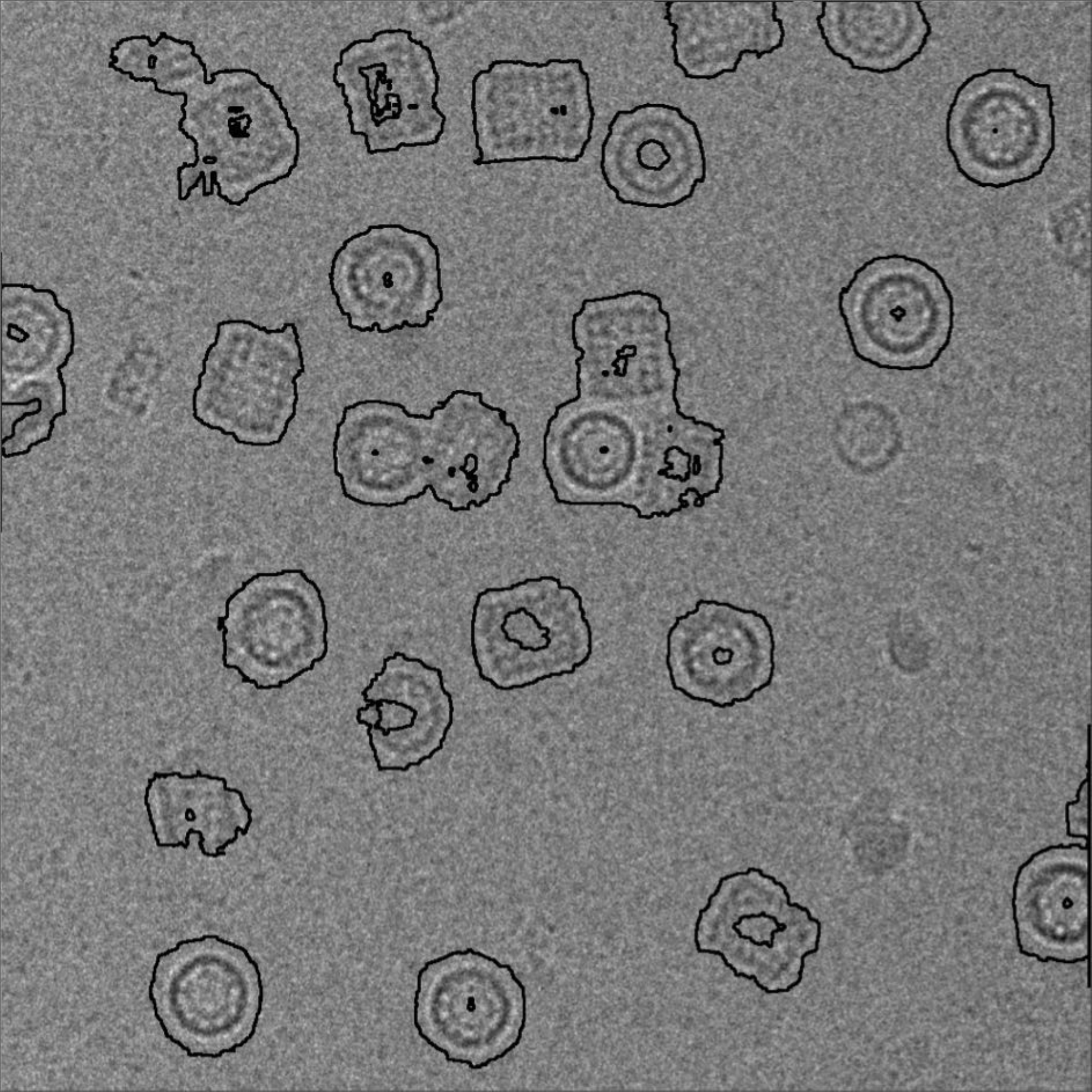}}}
\subfigure[]{\label{subfig:picked} {\includegraphics[width=0.25\linewidth]{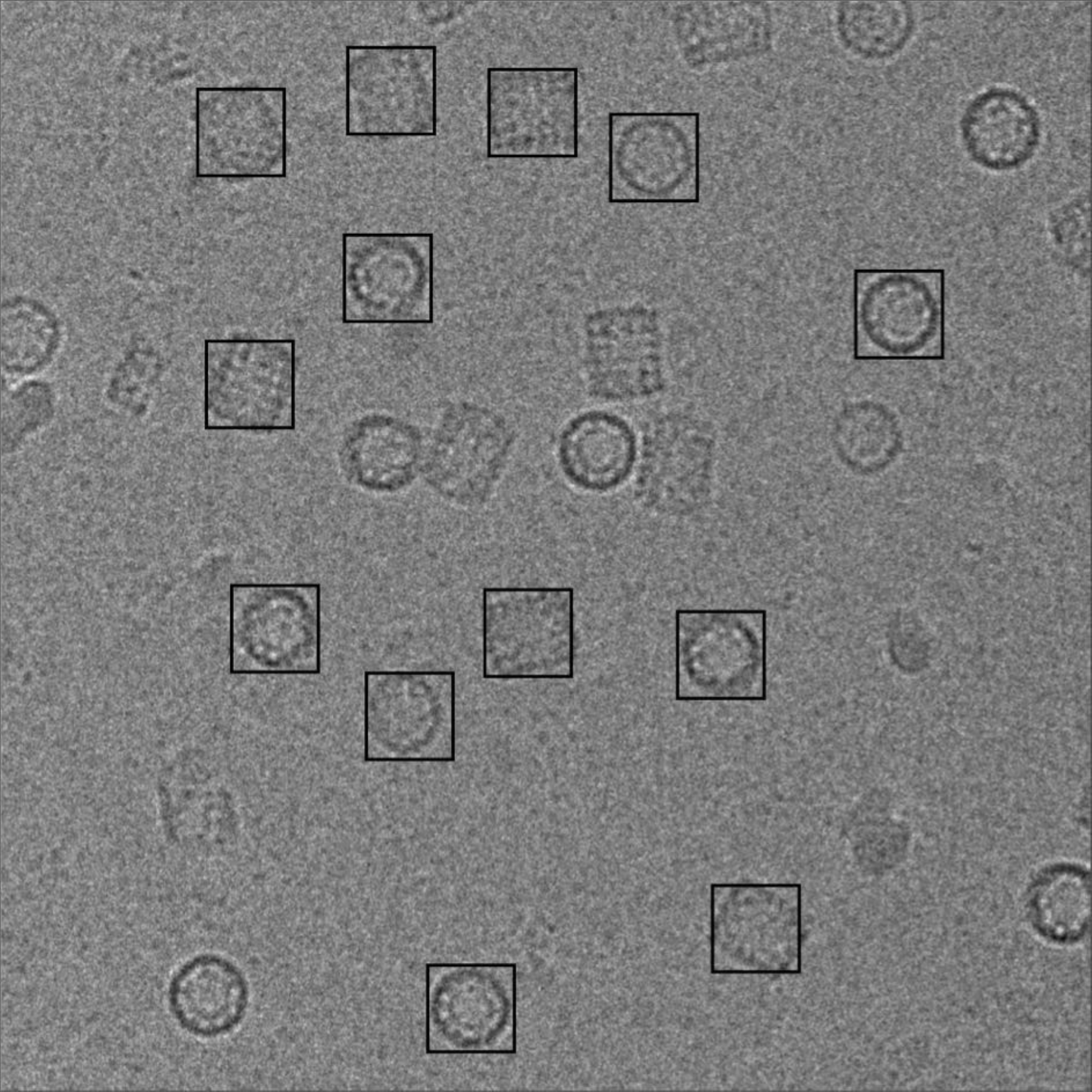}}}
\caption{Result of our suggested framework. The left column contains micrographs. The middle column contains the output of the classifier. 
 The right column contains the picked particles. Top row contains a  $\beta$-Galactosidase  micrograph. Bottom row contains a KLH micrograph.}
 \label{fig:beta1}
\end{figure*}

Single-particle cryo-electron microscopy (cryo-EM) aims to determine the structure of 3D specimens
(macromolecules) from multiple 2D projections. 
In order to acquire these 2D projections,  
a solution containing 
the macromolecules is 
frozen in vitreous ice on carbon film, thus creating a sample grid. An electron beam then passes through the ice and the macromolecules
frozen within,  creating 2D projections.  

Unfortunately, due to radiation damage only a small number of imaging electrons can be used in the creation of the micrograph. 
As a result,  micrographs have a low signal-to-noise ratio (SNR).  
An elaboration on the noise model can be found in \citep{sigworth1}.  

Since micrographs typically have low SNR, each micrograph consists of regions of noise and regions of noisy 2D projections 
of the macromolecule. In addition to these,  micrographs also contain   
regions of non-significant information stemming from 
contaminants such as  carbon film. 

Different types of regions have different typical intensity values. 
The 
regions of the micrograph that contain only noise will typically have 
higher intensity values than  other regions. 
In addition, regions containing a
particle typically have higher variance
than regions containing noise alone \citep{review, VanHeel}. 
Due to this, two cues that can be used for projection image identification are the 
mean and variance of the image.

In order to determine the 3D structure at high resolution, many projection images are needed, 
often in the hundreds of thousands. 
Thus, the first step towards 3D reconstruction of macromolecules consists of determining 
 regions of the micrograph that contain a particle as opposed to 
 regions that contain noise or contaminants. This is the particle picking step.

 A fully manual selection of  hundreds of thousands of 2D
projections is tedious and time-consuming. For this reason, semi-automatic and automatic particle picking is a much
researched problem for which numerous  frameworks have been suggested. 
Solutions to the particle picking problem include  edge detection
\citep{edgeDetection}, deep learning \citep{deep04, deepPickler, deep2},   
support vector machine classifiers 
\citep{textons}, and template matching \citep{azimut}. 

Template matching is a  popular approach to particle picking.  
The input to template matching schemes consists of a micrograph and  images containing 2D templates to match.  
These templates can be, for example, generated from manually selected particle projections. 
The aim is to output the regions in the micrograph that contain the sought-after templates. 

The basic idea behind this approach \citep{signature, azimut, templateDisk, eman, relion} is that the cross-correlation\footnote{Cross-correlation  is not the only possible function to use for template matching methods. For a review of other possibilities see \citep{review}.}
between 
a template image and a micrograph is larger 
in the presence of the template. An issue with this method is the high rate of false detection.  
This issue stems from the fact that given enough random data, meaningless noise can be perceived as a pattern. 
This problem was exemplified in \citep{pitfall, einsteinFromNoise}, where an image of Einstein was used as the template 
and matched to random noise.  
Even though the image was not present in the noise images, a reconstruction from the best-matched images yielded the original Einstein image.

One example  of a template-based framework is provided in RELION \citep{relion, relion3, relion2}. In this framework, 
the user  manually selects approximately one thousand particles from a small number of micrographs. These particle images are
then 2D classified 
to generate a smaller number of template images that are used to automatically select particles from all
micrographs. These particle images are then classified in order to identify non-particles.  
Additional examples of template-based frameworks include SIGNATURE \citep{signature} which employs a post-processing step that ensures  
the locations of any two picked particles cannot overlap, and gEMpicker \citep{gEMpicker} which employs several 
strategies to speed up template matching.

Template matching can also be performed without the input of template images. 
For example, see  
 \citep{DoGPicker} which is based on difference of Gaussians and  is 
 suitable for identifying blobs of a certain size 
  in the micrograph. Another template-free particle picking framework is gautomatch \citep{gautomatch}.

In this paper we propose a particle picking framework that is fully automatic and data-adaptive in the sense that no manual selection is used and no  templates are involved. 
Instead of templates we use a set of automatically selected reference windows. This set includes both particle and noise windows. 
We show that it is possible to determine   
the presence of a particle 
in any query image (\textit{i.e.}, region of the micrograph) 
through cross-correlation with each window of the reference set. Specifically,  
in the case where the query image contains noise alone, since there is 
no template to match, 
the cross-correlation coefficients should 
not indicate the presence of a template regardless of the actual content of each reference window. 
On the other hand, in the case where the query 
image contains a particle, the coefficients will depend on the content of each reference window. 

Once their content is determined, the 
query images most likely to contain a particle and  those  most 
likely to contain noise can be used to train a classifier. The output of this classifier 
is used for particle picking. 

We test our framework on publicly available datasets of $\beta$-Galactosidase dataset  \citep{scheres1, relion, scheres2}, 
T20S proteasome \citep{e10057}, 70S ribosome \citep{n10077} and 
keyhole limpet hemocyanin  \citep{KLH, bakeoff}. 
Some sample results are presented in Fig \ref{fig:beta1}. 
Code for our framework is publicly available.\footnote{https://www.github.com/PrincetonUniversity/APPLEpicker}

We note that our formulation can ignore the contrast transfer function (CTF). This is because the CTF is roughly the same throughout the micrograph and our particle selection procedure performs on the individual micrograph level. Thus, while CTF-correction is not strictly necessary, we discuss the advantage of applying our framework to CTF-corrected micrographs in  Section \ref{subsec:ctf}

\section{Material and Methods}
\label{sec:tm}

In Section 
\ref{subsec:general concept} we  detail our method for 
 determining the content of a single query image $g \in \mathbb{R}^{n \times n}$, where the 
 query image is a window extracted from the micrograph and $n$ is chosen such that the 
 window size is slightly smaller than the particle size (which we assume is known).\footnote{The notation $g \in \mathbb{R}^{n \times n}$ 
 simply means that the size of a query image is $n \times n$ and its content is real-valued.}
This method necessitates the use of a reference set $\{f_m \in \mathbb{R}^{n \times n}\}_{m=1}^{B}$ 
selected from the micrograph in the automatic manner detailed in Section \ref{subsec:windows}.  
We generalize our method to particle picking 
from the full micrograph in Section \ref{subsec:generalize}. Section \ref{subsubsec:class} improves localization through the use of a fast classification step. The complete method, known as the APPLE picker, is described in Section \ref{subsubsec:pick}.

\subsection{Determining the Content of a Query Image}
\label{subsec:general concept}

The idea behind traditional template matching methods is that the cross-correlation score of two similar images is high. Specifically, a  
template image known to contain a particle can  be used in order to identify similar patterns in the micrograph using  cross-correlation. 
In this section we show that
the same idea can be used to determine the content of regions of the micrograph even when no templates are 
available. To this end we use the cross-correlation between  
a query image $g$ 
and a set of reference images  $\{f_m\}_{m=1}^{B}$. The cross-correlation function is \citep{review}
\begin{equation}
c_{f_m,g} \left(x, y \right) = \sum_{x'} \sum_{y'} f _m\left(x', y' \right) g \left(x + x', y + y' \right).%
\label{equ:cross_time}
\end{equation}
This function can be thought of as a score associated with $f_m$, $g$ and an offset $\left(x, y \right)$. 

The cross-correlation score at a certain offset does not in itself have much meaning without  the context of the score in nearby offsets.  
For this reason we define the following normalization on the cross-correlation function
\begin{equation}
\hat{c}_{f_m,g} \left(x, y\right) = c_{f_m,g} \left(x, y\right) - \frac{1}{n^2} \sum_{x'} \sum_{y'} c_{f_m, g} \left(x', y'\right),%
\label{equ:templateMatchingIdea}%
\end{equation}%
where the second term is the mean of $c_{f_m,g} \in \mathbb{R}^{n \times n}$. 
We call (\ref{equ:templateMatchingIdea}) a normalization since it shifts all cross-correlations to a common baseline. 

Consider the case where query image $g$ contains a particle. The score 
$c_{f_m, g} \left(x, y\right)$ is expected to be maximized when $f_m$ contains a particle
with a similar view. In this case there will be some offset $\left(x, y\right)$ such that the images $f_m$ and $g$ match best,
and $c_{f_m, g} \left(x, y\right) > c_{f_m, g} \left(x', y'\right)$  
for all other offsets $\left(x', y'\right)$.  Thus,
\begin{equation}
c_{f_m, g} \left(x, y\right) > \frac{1}{n^2} \sum_{x'} \sum_{y'} c_{f_m, g} \left(x', y'\right).
\label{equ:templateMatchingIdea2}
\end{equation}In other words, $\hat{c}_{f_m, g} \left(x, y \right)$ is expected to be large and positive. 
In this case, we say  $g$ has a strong response to $f_m$.

Next, consider the case where query image $g$ contains no particle. In this
case there should
not exist any offset $\left(x, y \right)$ that greatly increases the match
for any $f_m$. Thus
typically $\hat{c}_{f_m, g} \left(x, y \right)$ is comparatively small in magnitude. 
In other words, $g$ has a weak response to $f_m$.

We define a response signal $\mathbf{s}_g$ such that
\begin{equation}
\mathbf{s}_g \left(m \right) =  \underset{x, y}{\max} \quad \hat{c}_{f_m, g} \left(x, y \right), \quad m = 1, \dots, B.
\label{equ:vectorS} \end{equation}
This signal is associated with a single query image $g$. Each entry $\mathbf{s}_g \left(m \right)$ 
contains the maximal normalized cross-correlation with a single reference image $f_m$. Thus, the response signal captures the
strength of the response of the query image to each of the reference images. 

\begin{figure}[H]
\centering
{\includegraphics[width=0.45\linewidth]{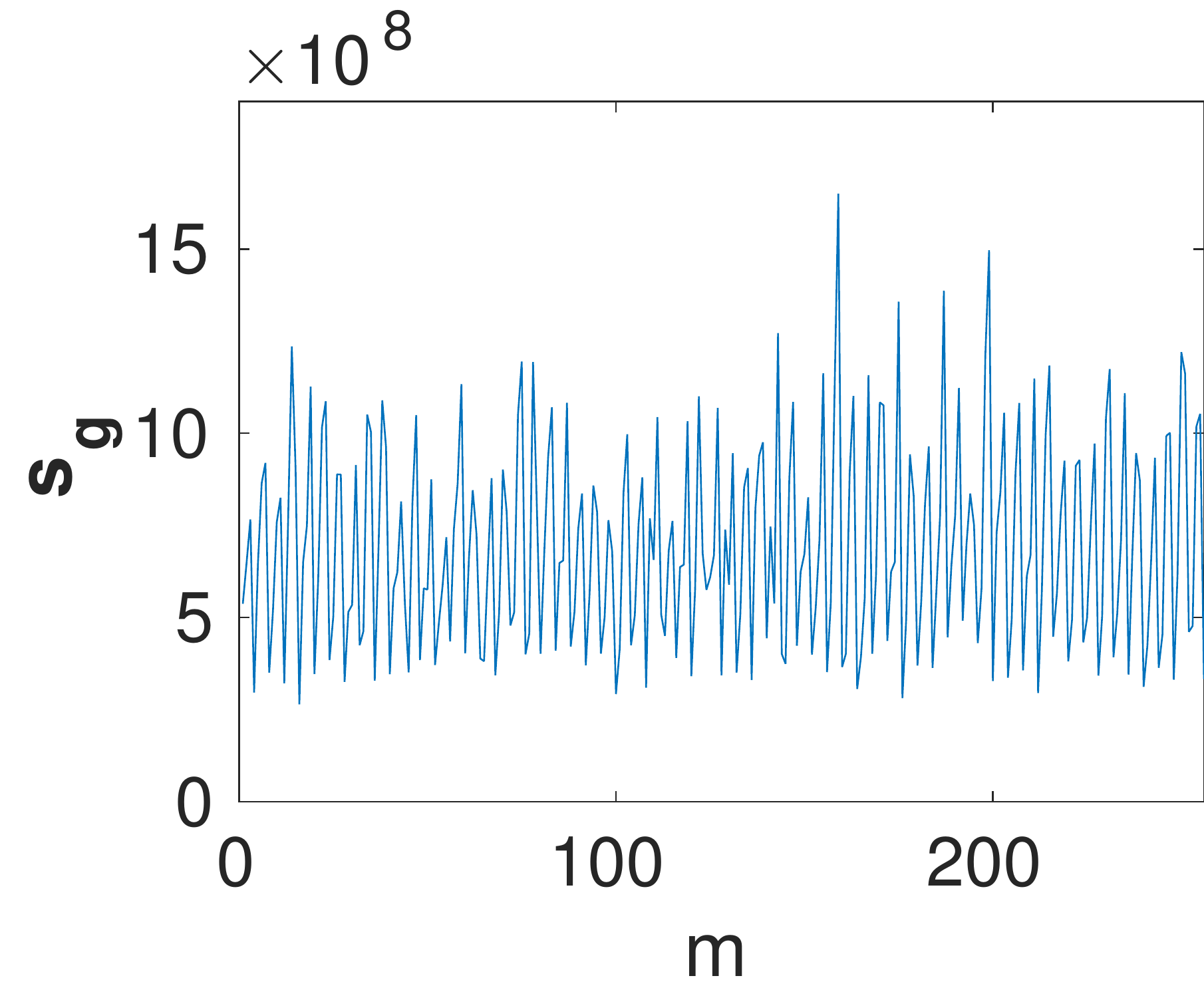}} \hspace{0.25cm}
{\includegraphics[width=0.415\linewidth]{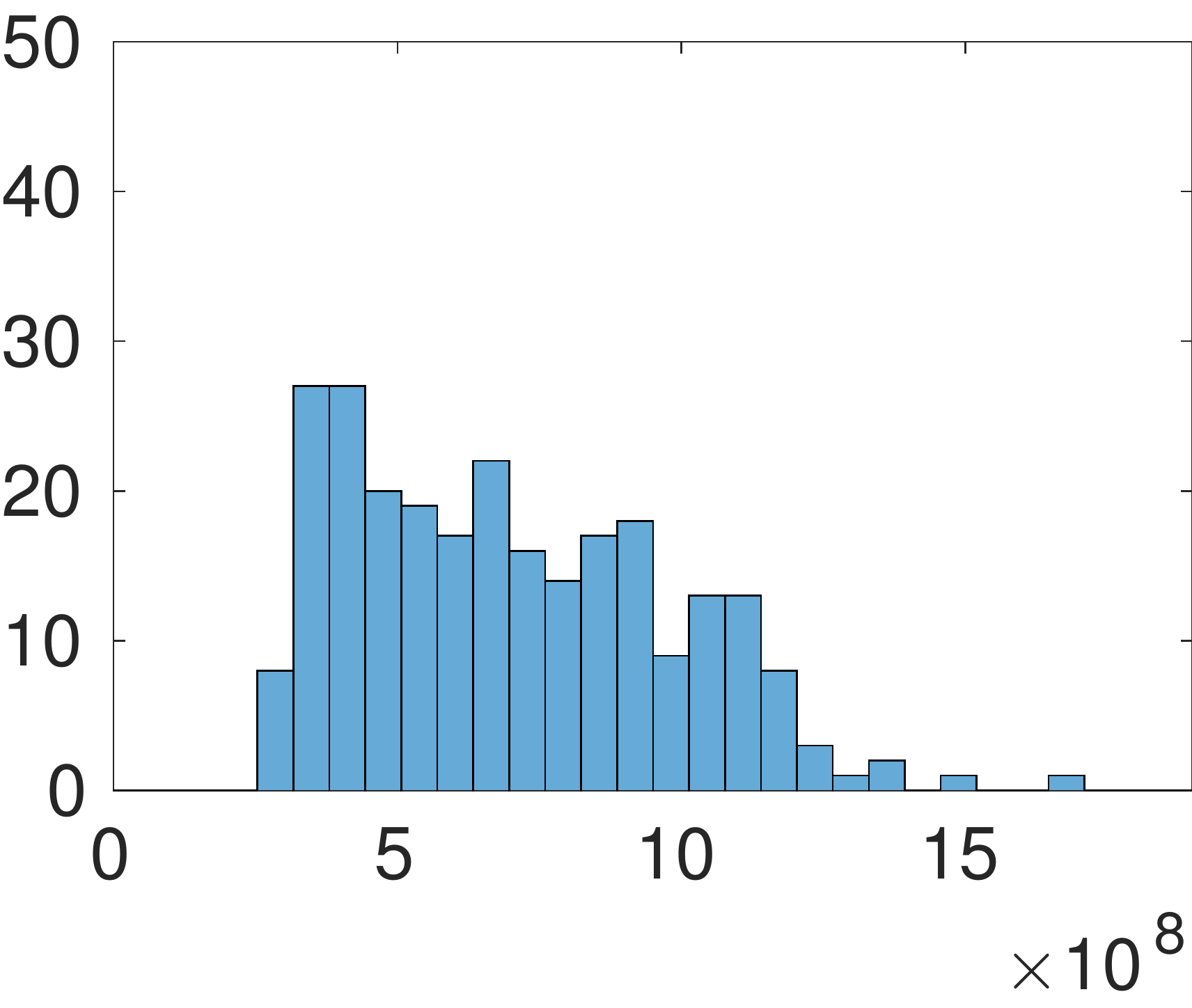}}\\
{\includegraphics[width=0.45\linewidth]{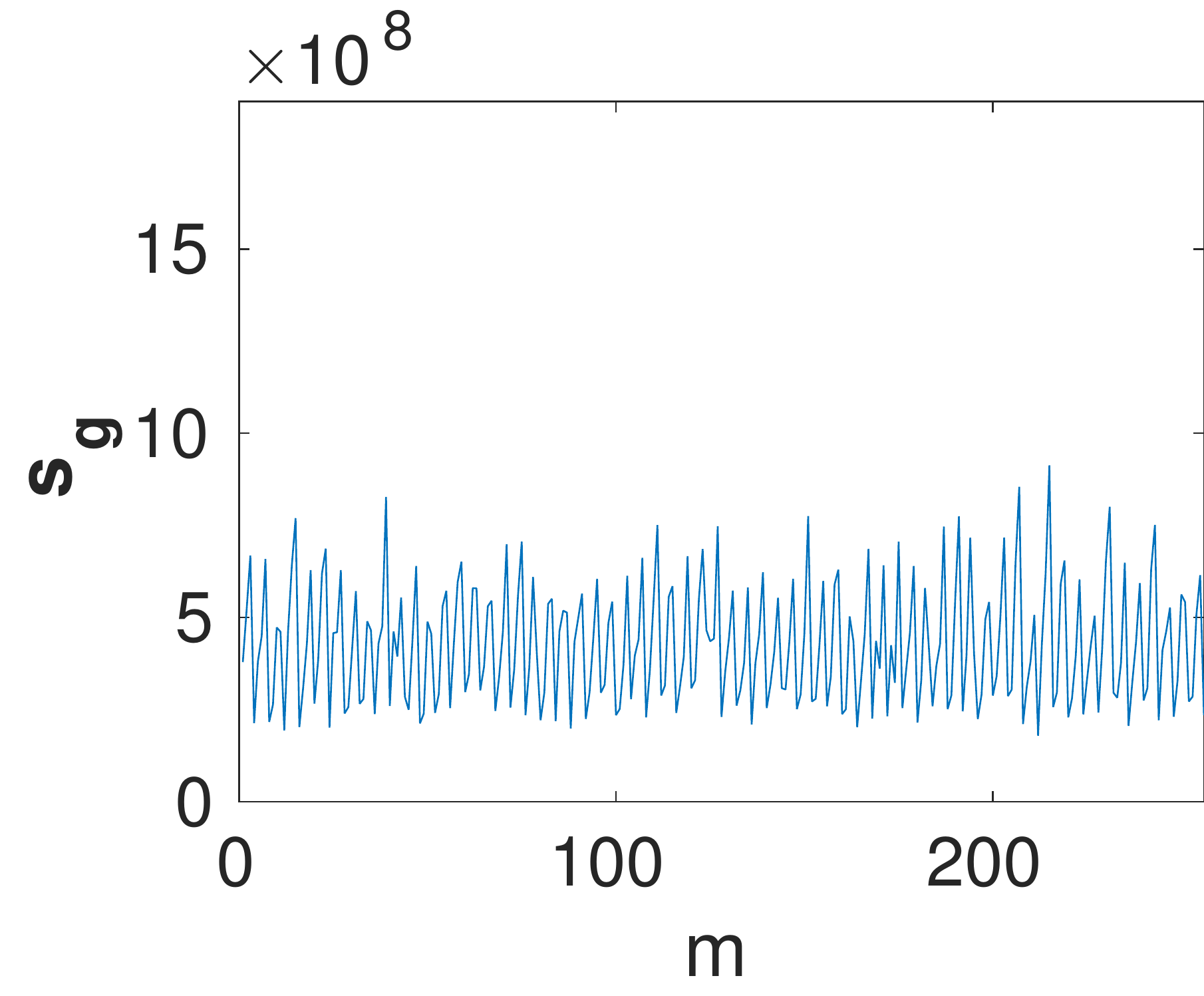}}\hspace{0.25cm}
{\includegraphics[width=0.415\linewidth]{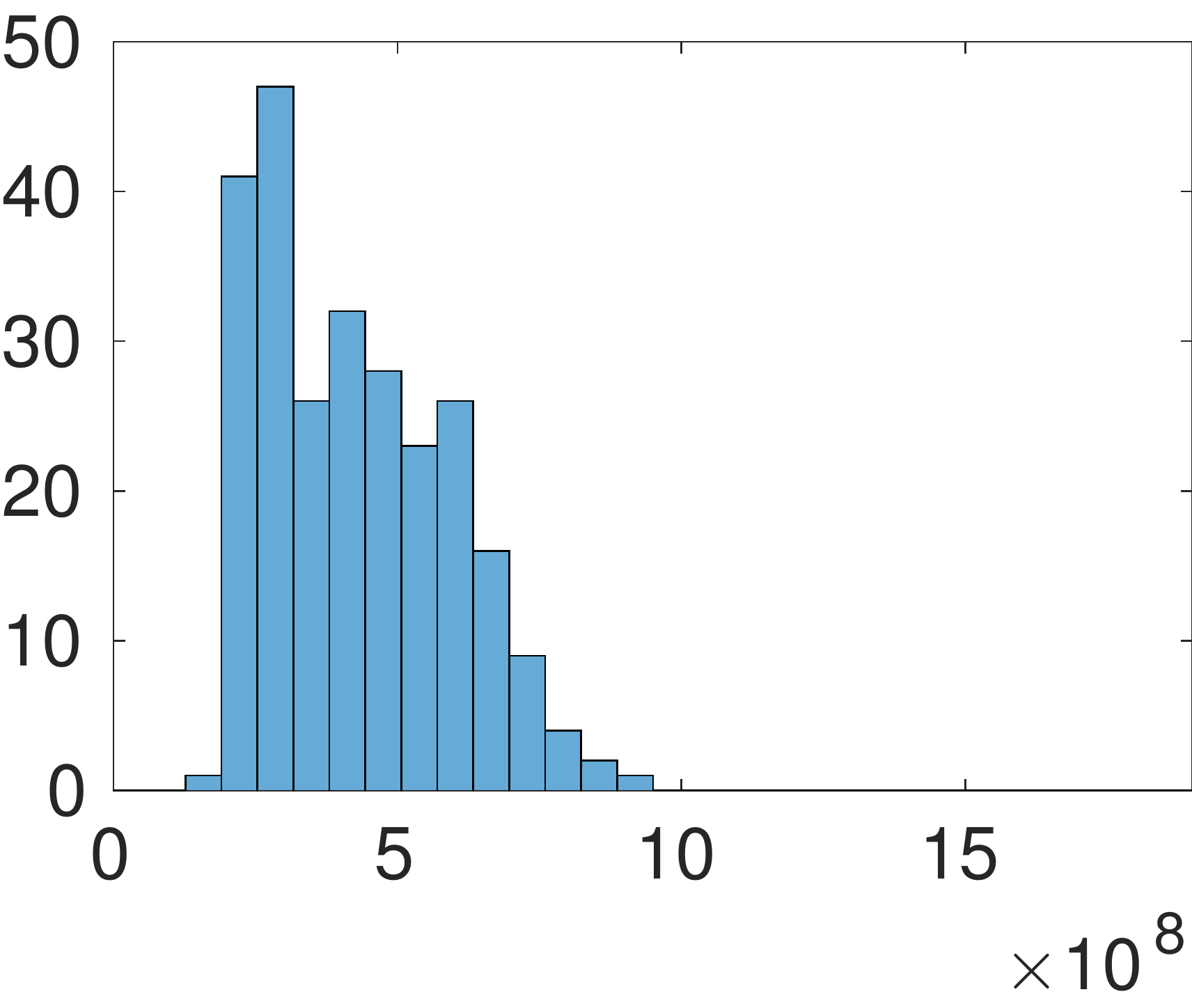}}
\caption{Response signal of a particle image (top) and a noise image (bottom). The left column contains the 
response signals. The right column contains histograms of the response signals.}
\label{fig:signal}
\end{figure}

We suggest that 
 $\mathbf{s}_g$ can be used to determine  
the content of $g$.  
If the query image contains a particle, $\mathbf{s}_g$ will show a
 high response to reference images containing a particle with similar view  
and a comparatively low response to other images.
As a consequence, $\mathbf{s}_g$ will have several high peaks. 
On the other hand, if the query image contains noise alone, $\mathbf{s}_g$ will have relatively uniform content. 
This idea is shown in Figure \ref{fig:signal}.

The above is true despite the high rate of false positives in cross-correlation-based methods. 
This is due to the comparison of each query image to multiple reference windows. The redundancy 
causes robustness to false positives.

\subsection{Reference Set Selection}
\label{subsec:windows}
The set of reference images $\{ f_m \}_{m=1}^{B}$ could contain all possible windows in
the micrograph. However, this would lead to unnecessarily long runtimes. 
Thus, we suggest to choose
a subset of $B$  windows from the micrograph, where each of these windows is either likely to contain a particle or  
likely to contain noise alone. 

\begin{figure}[H]
\centering
\subfigure[]{\label{fig:containers} \includegraphics[width=0.45\linewidth]{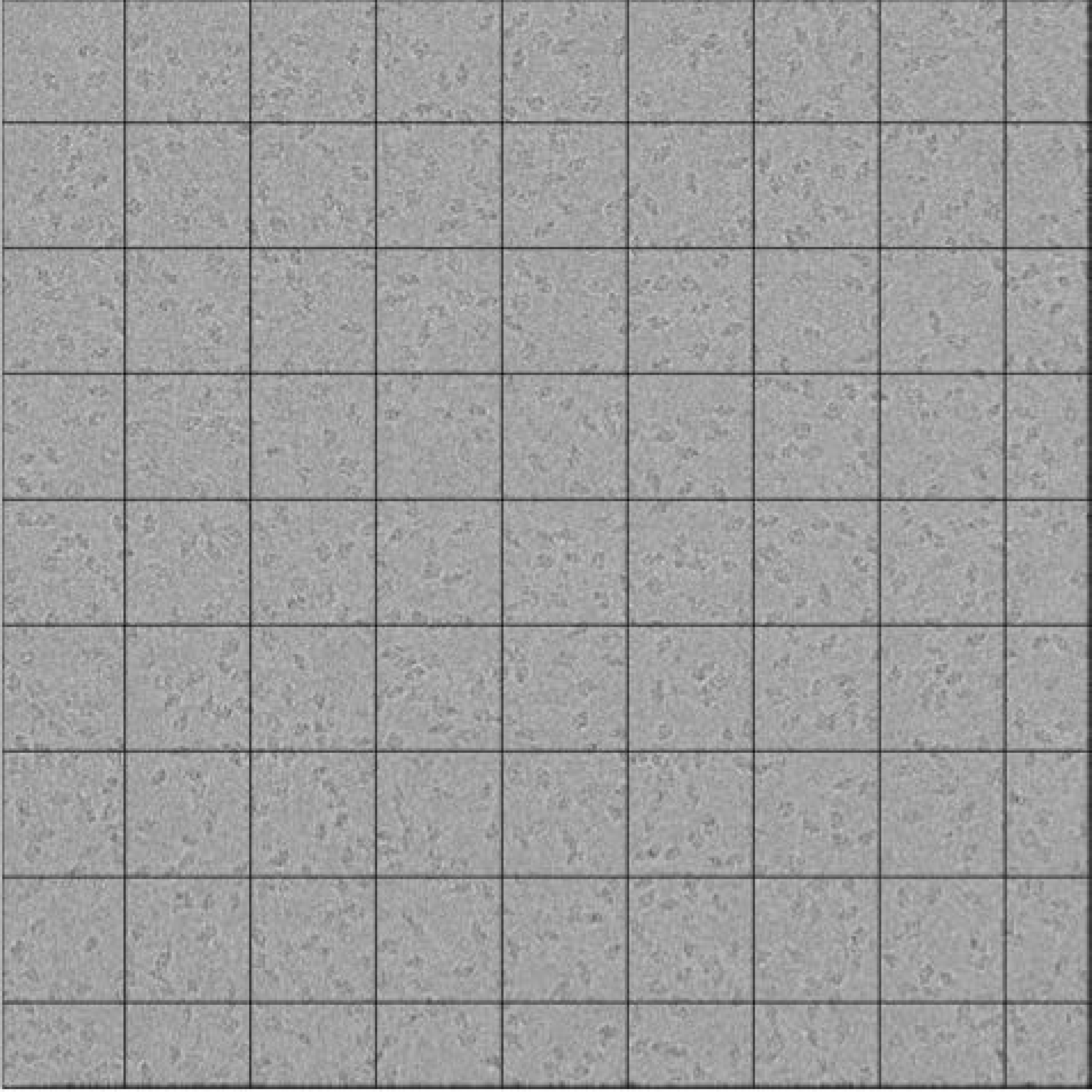}} 
\subfigure[]{\label{fig:interestingWindows} \includegraphics[width=0.45\linewidth]{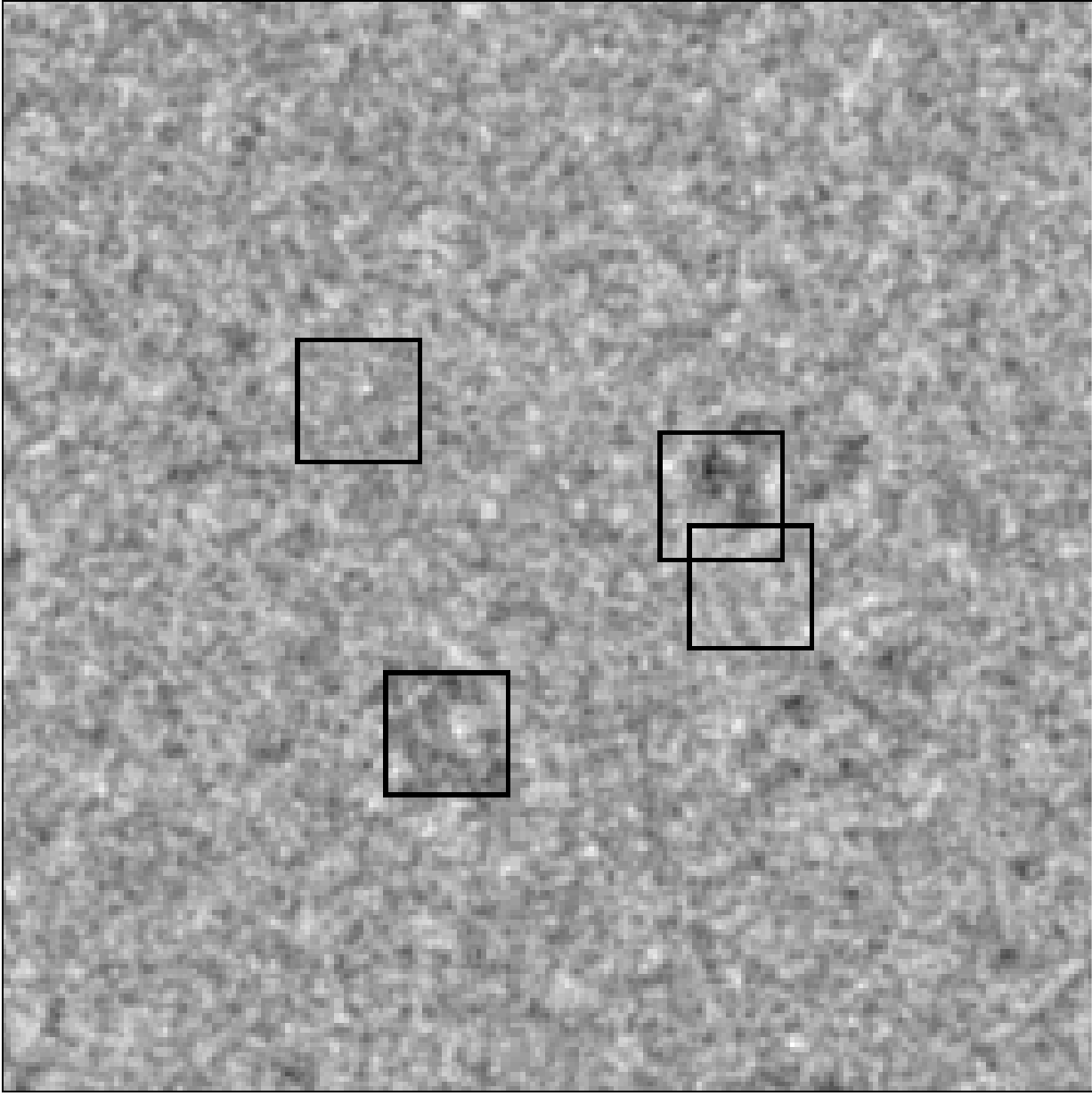}}
\caption{(a) Containers of a micrograph of the $\beta$-Galactosidase dataset. (b) Single container with four windows of interest.}
\end{figure}

In order to automatically select this subset, we first divide the 
micrograph into ${B}/{4}$ non-overlapping \textit{containers}. A container is some 
rectangular portion of the micrograph. Each container holds many $n \times n$ windows. 
Figure \ref{fig:containers} is an example of the division of a micrograph into containers.

As mentioned in Section \ref{sec:introduction} (fourth paragraph), regions containing noisy projections of particles typically have lower
intensity values and higher variance than regions containing noise alone.
Thus we find that the window with the lowest mean intensity in each container likely contains a particle and the
window with the highest mean intensity likely does not. We extract these windows from each container and include them 
in the reference set. We do this also for 
the windows that have the
highest and lowest variance in each container.  
This procedure provides a set of $B$  reference windows. Figure \ref{fig:interestingWindows}
presents the  reference windows extracted from a single container. We suggest setting $B$ to approximately $300$. 

\begin{figure*}[t]
\centering
\subfigure[]{\includegraphics[width=0.28\linewidth]{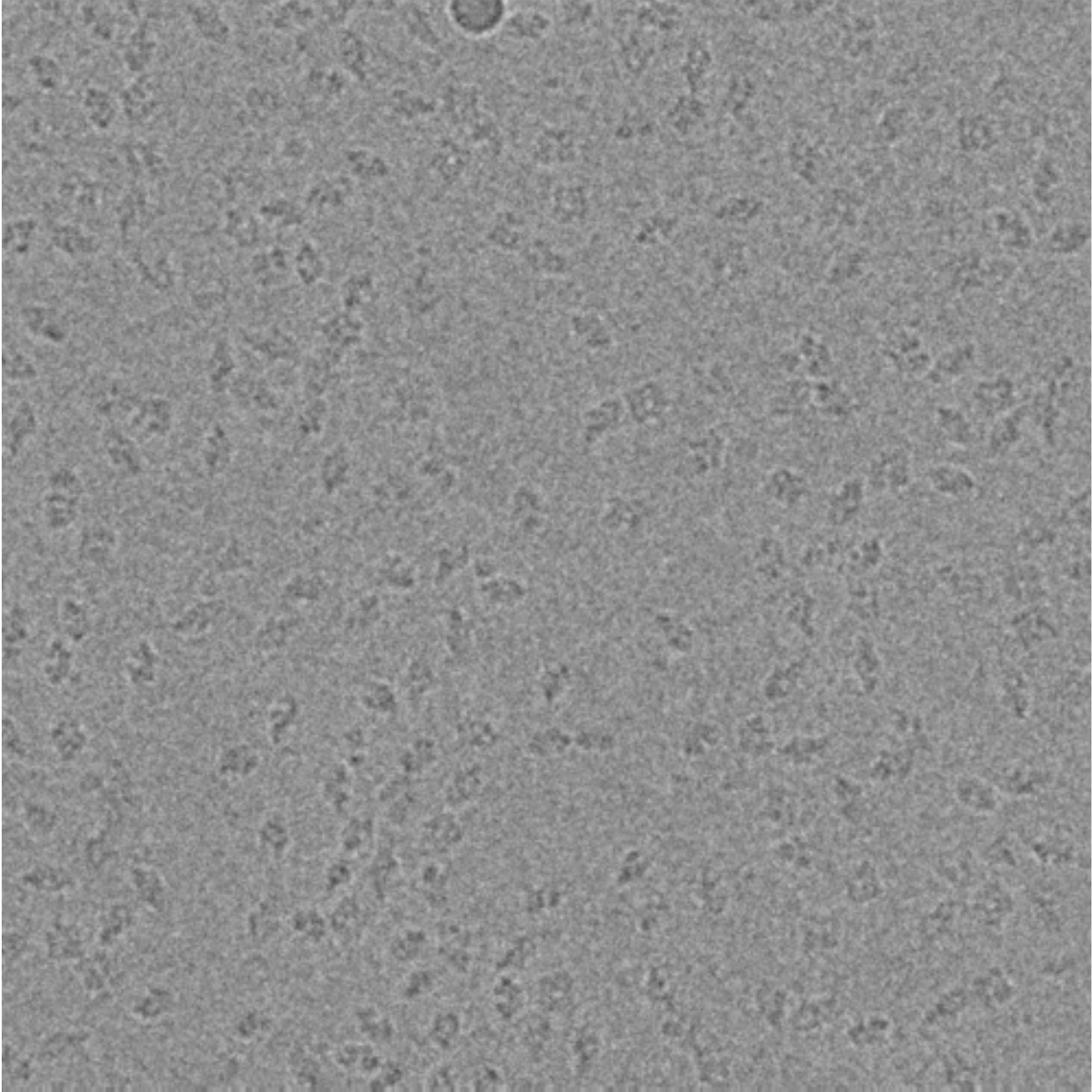}}
\subfigure[]{\label{fig:high}\includegraphics[width=0.28\linewidth]{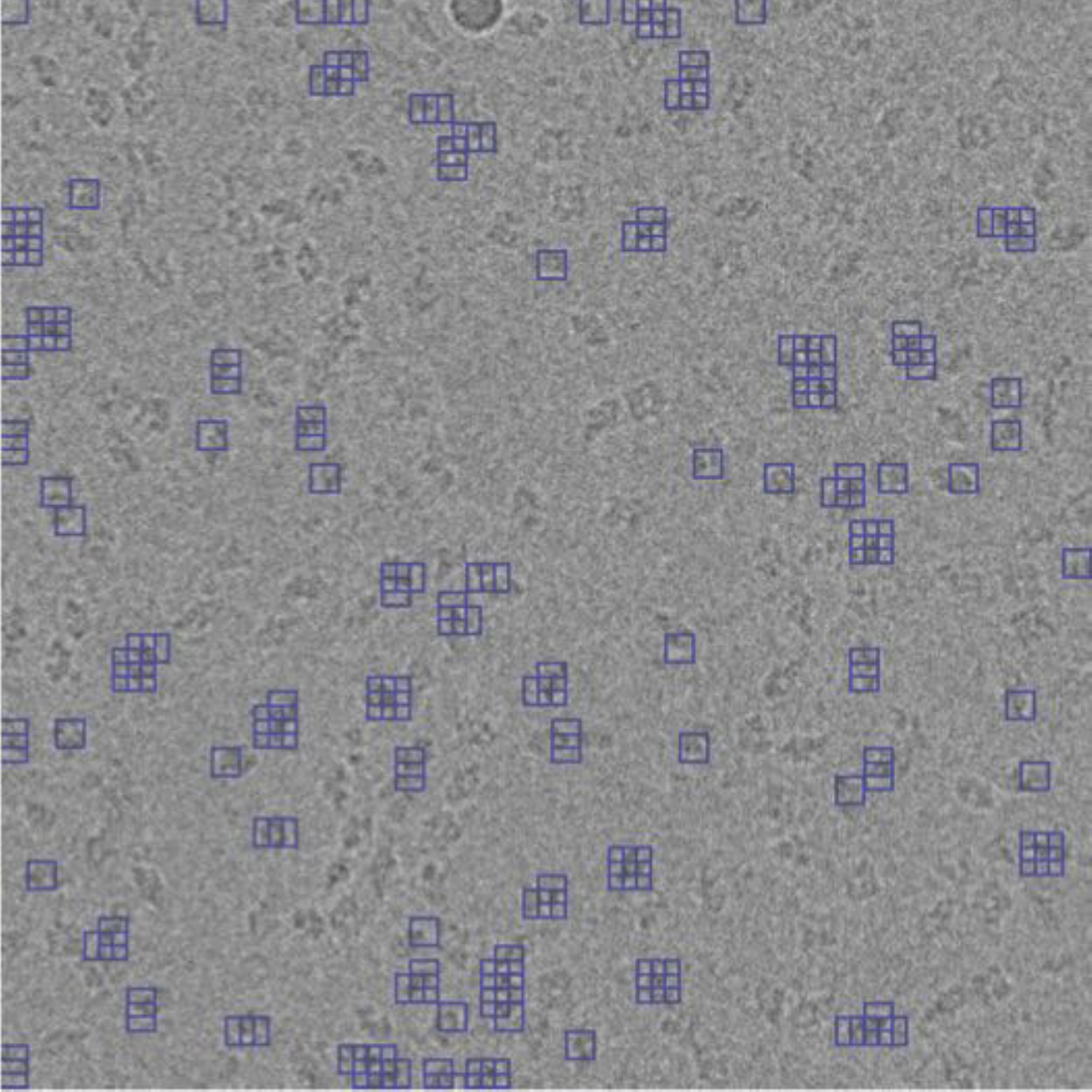}}
\subfigure[]{\label{fig:low}\includegraphics[width=0.28\linewidth]{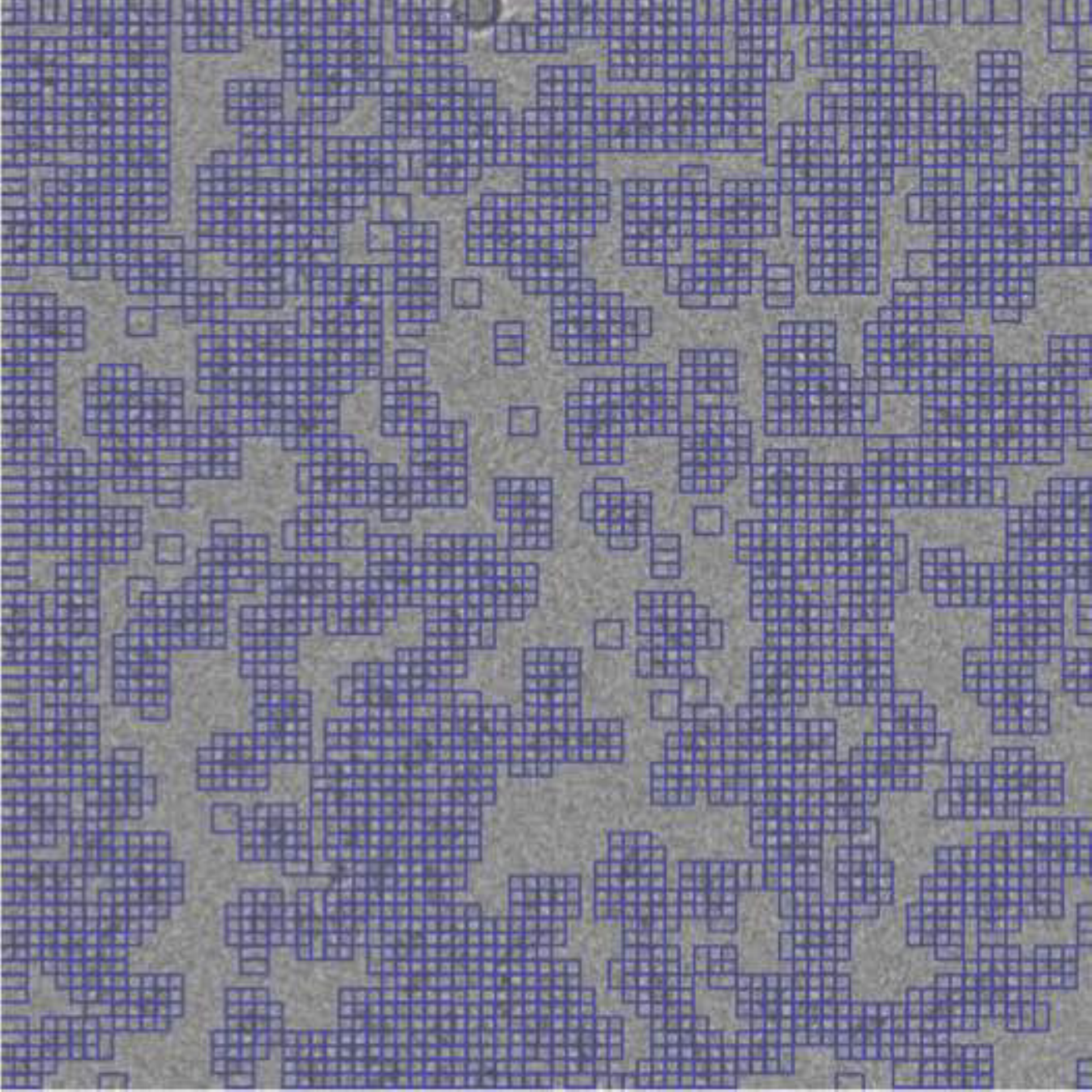}}
\caption{Result of our cross-correlation scheme. (a) Micrograph of $\beta$-Galactosidase. 
(b) The $1000$ regions contained in boxes have high $k \left( \mathbf{s}_g \right)$ and 
thus should contain a particle. 
(c) There are $9000$ regions contained in boxes. These regions have high or intermediate $k \left( \mathbf{s}_g \right)$ 
and thus may contain a particle. Consequently, the   regions not contained in boxes have low $k \left( \mathbf{s}_g \right)$ 
and thus are likely to be pure noise.}
\label{fig:particle_demarc}
\end{figure*}

The set of reference windows must contain both windows with noise and windows with particles. 
It may seem counter-intuitive to include noise windows in a reference set. However, for roughly symmetric particles 
(\textit{i.e.}, particles with similar projections from each angle), 
any query image will have a similar response to every   
reference image which contains a particle. Thus, if noise images were not included in the reference set, the response 
signal $\mathbf{s}_g$ would be uniform regardless of the content of $g$.

\subsection{Generalization to Micrographs}
\label{subsec:generalize}
We extract a set of $M$ query images from the micrograph. 
These images should have some overlap. In addition, their union 
should cover the entire micrograph. 
For example, we can choose windows on a grid with step size  $n / 2$.  
In order to determine the content of each query image $g$, we examine   
the number  of entries  
that are over a certain threshold, \textit{i.e.},   
\begin{equation} \label{equ:smoothness}
k \left( \mathbf{s}_g\right) =  \vert \{i ~ \text{such that} ~ \mathbf{s}_g \left( i \right)> t \} \vert,%
\end{equation}
where  the threshold $t$ is determined according to the set of response signals and is experimentally set to 
\begin{equation} \label{equ:tau}
t= \frac{ \underset{g, i}{\max} \quad \mathbf{s}_{g} \left( i \right) - \underset{g, j}{\min} \quad \mathbf{s}_{g} \left( j \right) }{20} + \underset{g, j}{\min} \quad \mathbf{s}_{g} \left( j \right).
\end{equation}

Any query image $g$ that possesses high $k\left( \mathbf{s}_g\right)$ is known to have had a relatively strong response to 
a large amount of reference windows and is thus expected to  
contain a particle. On the other hand, 
a query image $g$ that possesses  low $k\left( \mathbf{s}_g\right) $ is expected to  
contain noise.  In this manner we may consider $k\left( \mathbf{s}_g\right) $ as a score for $g$. The 
higher this score, the more confident we can be that $g$ contains a particle.

The strength of the response, and thus the score of a query image, is determined by the threshold $t$. 
Instead of checking the uniformity of the response signal for a single query image as was done in Section \ref{subsec:general concept}, we  
use the response signals of the entire set to determine a threshold above which we consider a response to be strong.

For visualization of our suggested framework, we turn to a micrograph of $\beta$-Galactosidase \citep{relion, scheres1, scheres2}. 
We select $B = 324$ reference images in the manner detailed in Section \ref{subsec:windows}, and 
aim to classify $21904$ query images. The query images are selected from locations throughout the micrograph in a way that ensures some overlap 
between images. For each query image we compute the corresponding response signal according to (\ref{equ:vectorS}). 
The threshold $t$ is then computed from all the response signals according to (\ref{equ:tau}). Once 
this is done, the value $k \left( \mathbf{s}_g\right)$ is computed for each query image. 
We present in Figure \ref{fig:particle_demarc} a visualization of the results. Since we expect query images that 
contain particles to be associated with high-valued $k \left( \mathbf{s}_g\right)$, we present the $1000$ query 
images with highest $k \left( \mathbf{s}_g\right)$. Figure \ref{fig:high} shows that, as expected, these regions do contain particles. 
In addition, we present the $9000$ query 
images with highest $k \left( \mathbf{s}_g\right)$. The regions not contained in any of these query images are associated with low-valued $k \left( \mathbf{s}_g\right)$ 
and can be seen in in Figure \ref{fig:low}  to contain no particle.

We note that 
for the sake of reducing the computational complexity of our suggested framework,  
the cross-correlation score is computed using fast Fourier transforms.  
This is a well-established method of reducing complexity \citep{review}. 

\subsection{APPLE Classification}
\label{subsubsec:class}

A particle picking framework should produce a single window containing each picked particle. 
It is possible to use the output of the cross-correlation scheme introduced in  Sections  \ref{subsec:general concept}--\ref{subsec:generalize} 
as the basis of a particle picker. This is done by defining the query set to be the set of  
all possible $n \times n$ windows contained in the micrograph.  The content of each query window is determined according to 
its score. Specifically, if the score is above a threshold we determine that it contains a particle. 
This determination can be applied to the location of the central pixel in that window to  
provide a classification of each pixel in the micrograph (except for  boundary pixels that are not in the center of any possible $n\times n$ window). 
Unfortunately, the cost of such an endeavor, both in runtime and in memory consumption, is prohibitive.  

In order to improve performance, the APPLE picker does not define the set of query 
images as all possible $n \times n$ windows in the micrograph. Instead, the 
set of query images $\{h_m\}_{m=1}^{C}$ is defined as the set of all $n \times n$ windows extracted from the 
micrograph at $n/2$ intervals. However, applying a determination to the center pixel of each query window will no longer  
allow for successful particle picking. Indeed, where two overlapping query windows are 
determined to contain a particle, it is unknown whether they both contain the same particle or whether 
each contains a distinct particle. It is possible that the interval of $n/2$ between the query windows 
caused us to skip over windows that would have been classified as noise.  
In other words, in order to get a good localization of the particle, the content of each possible 
window of the micrograph should be determined. 

To achieve this, the APPLE picker determines the content of all possible windows in the micrograph 
via a  support vector machine (SVM) classifier.  This classifier is based on a few simple and easily calculated features that are known to 
differ between particle regions and noise regions. In this manner we achieve fast and localized particle 
picking. 
The classifier is trained on the images  whose  
classification (as particle or as noise) is given with high confidence by our cross-correlation scheme.

To train the classifier, we need a training set.
This is composed of a set of examples for the particle images, $S_1$, and a set of examples for the noise images, $S_2$.
The complete training set is $S_1 \cup S_2$. 
The choice of $S_1$ and $S_2$ depends on two parameters,  $\tau_1$ and $\tau_2$.  
These parameters correspond to the percentage of training images that we believe do contain a particle ($\tau_1$) and the 
percentage of training images that we believe may contain a particle ($\tau_2$).  

The selection of $\tau_1$ and $\tau_2$ can be made according to the concentration of the particle 
projections in the micrograph. This information can be estimated visually at the time of data collection 
from a set of initial acquired micrographs.

To demonstrate the selection of $\tau_1$ and $\tau_2$, we consider a micrograph with $M = 20000$ query images. 
If it is known that there is a mid to high concentration  of  
projected particles, we can safely assume that, \textit{e.g.}, $1000$ images with highest $k \left(\mathbf{s}_{h_m} \right)$ 
contain a particle. Thus we set $\tau_1 = 5 \%$. 
In addition, it is 
possible that out of $20000$ query images 
 $15000$ may contain some portion of a particle.  We can therefore safely assume that the regions of the micrograph that are not 
 contained in any of the $\tau_2 = 75\%$ images with highest $k\left(\mathbf{s}_{h_m} \right)$ 
will be regions of noise. 

When the concentration of particle projections is unavailable, the selection 
of  $\tau_1$ and $\tau_2$ can be done heuristically. For instance, $\tau_1=5\%$ and $\tau_2=75\%$  
is often a good selection for $\tau_1$ and $\tau_2$. 
We note that when the concentration of macromolecules is not high, the value of 
$\tau_2$ is less important than that of $\tau_1$.

Once $\tau_1$ is selected, the set $S_1$ is determined. Due to the overlapping nature of  query images, 
there is no need to use all $\tau_1$ percent of images with highest $k \left(\mathbf{s}_{h_m} \right)$ for training. 
Instead, we note that these images 
form several connected regions in the micrograph 
(see Figure \ref{fig:particle_demarc}).  The set $S_1$ is made of all non-overlapping windows extracted from these regions.

The $ \tau_2$ percent of query images with highest $k \left(\mathbf{s}_{h_m} \right) $ form  the regions in the micrograph that 
may  contain particles. An example of these regions can be seen in Figure \ref{fig:low}.
The set $S_2$ 
 is made of non-overlapping windows extracted from the complement of these regions.  
The reason for the difference between the determination of $S_1$ and $S_2$ is that the query images 
overlap, and we do not want to train the noise model from any section of the $\tau_2$ 
percent of query images with moderate to high $k \left(\mathbf{s}_g \right)$.

The training set for the classifier  consists of vectors $\mathbf{x}_1, \dots, \mathbf{x}_{\vert S_1 \cup S_2 \vert}  \in \mathbb{R}_{+}^2$
and labels $y_1, \dots, y_{\vert S_1 \cup S_2 \vert} \in \{0, 1\}$.
Each vector $\mathbf{x}_i$ in the training set contains the mean and standard deviation of a window  $h_i \in S_1 \cup S_2$, and is associated with a
label $y_i$, where
\begin{equation*}
y_i =
    \begin{cases}
      1, & \text{if}\ h_i \in S_1. \\
      0, & \text{if}\ h_i \in S_2.
          \end{cases}
\end{equation*}
We note that while training the classifier on mean and variance works sufficiently well, they are not necessarily optimal and 
other features can be added. This is the subject of future work.

\begin{figure*}
\centering
\includegraphics[width=1\linewidth]{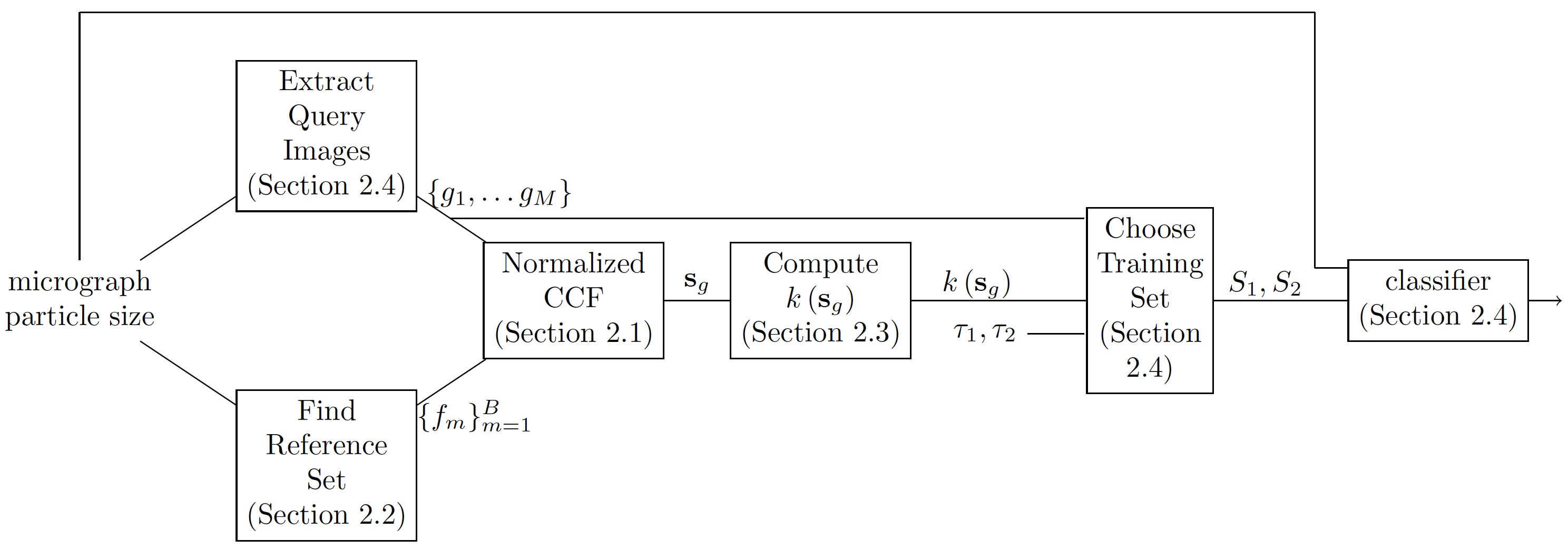}
\caption{Overview of APPLE picker.}
 \label{fig:overall}
\end{figure*}

The training set is used in order to train a support vector machine classifier 
\citep{svm3, svm7}. 
We propose using a Gaussian radial basis function SVM.
Once the classifier is trained, a prediction can be obtained for each 
window in the micrograph. 
This classification is attributed to the central pixel of the window, thus classifying each pixel in the micrograph as either a particle 
or a noise pixel. This  provides us with a segmentation of the micrograph. 
Figure \ref{subfig:particle} presents such a segmentation for the micrograph depicted in \ref{subfig:micro}.  For convenience, we summarize our framework in Figure \ref{fig:overall}.

\subsection{APPLE Picking}
\label{subsubsec:pick}
The output of the classifier is a binary image where each pixel is labeled as either {particle} or as {noise}. Each connected region (cluster) of particle pixels may
contain a particle. On the other hand it may contain some artifact. Thus, we disregard clusters that are too small or too big. 
This is done through examining the total number of pixels in each cluster, and 
discarding any that are  above or below 
a reasonable number of pixels. This number is selected based on the true particle size. 

Alternatively, this can be done through  
use of morphological operations. An erosion \citep{erosion} is a morphological operation preformed on a binary image wherein pixels from each cluster 
are removed. The pixels to be removed are determined by proximity to the cluster boundary. 
In this way, the erosion operation shrinks the clusters of a binary image. This shrinkage can be 
used to determine the clusters that contain artifacts. Large artifacts will remain when shrinking by a factor larger than the particle size. 
Small artifacts will disappear when shrinking by a factor smaller than the particle size. We use this method of artifact removal in Section 
 \ref{sec:klh}. We note that a similar method for contaminant removal was used in AutoPicker \citep{templateDisk}.

Beyond these artifacts, it is possible that two particles are frozen very close together. This will distort the true particle projection and should be disregarded.  
For this reason it is good practice to disregard pairs of clusters of pixels that were classified as particle if they are too close. We do this by 
disregarding clusters whose centers are closer than some distance, for example the particle diameter. 
We then output a box around the center of each remaining cluster of pixels that were classified as particle. 
The size of the box is determined according to the known  particle size. 
The pixel content of each box is a particle picked by our framework. See Figure \ref{subfig:picked}. 

After all particles are picked, it is possible to create templates out of them and use a template matching scheme to 
pick additional particles, as in \citep{azimut, eman, relion}.

\subsection{CTF correction}
\label{subsec:ctf}
In the process of acquiring  the micrograph each particle projection is convolved with a point spread function. This function is the 
inverse  Fourier transform of a function called the Contrast transfer function (CTF), which is defined as follows \citep{ctf_paper}
\begin{equation}
\begin{split}
& CTF \left( g \right) = -\sqrt{1-A^2} \sin \left(\chi \right) - A \cos \left( \chi \right)\\
& \chi = \pi \lambda g^2 \Delta f - \frac{\pi}{2} C_s \lambda^3 g^4,
\end{split}
\end{equation}
where $\Delta f$ is the defocus, $\lambda$ is the wavelength, $g$ is the radial frequency, $C_s$ is the spherical aberration and $A$ is 
the amplitude contrast. 

A well-known effect of the CTF is increasing the support size of the projection image. 
This effect may cause nearby particle projections to become difficult to distinguish.  
Another issue is that the CTF decreases the contrast of the projection images, which makes them harder to find. 
Due to the above, while there is no strict necessity to apply the APPLE picker to  CTF-corrected micrographs, it is good practice to 
do so. The problems of CTF estimation \citep{ctffind} and CTF-correction \citep{downing_ctf, ctf2_paper} are well researched problems.  
We use CTFFIND4 \citep{ctffind} for CTF estimation. 

One method of CTF-correction is phase-flipping, which preserves the statistics of the noise, while effectively preventing the CTF 
from changing sign. While this method does not correct for the amplitude of the CTF, the phase correction already brings 
the support size close to its true value. It also slightly increases the particle contrast.

\begin{figure*}
\centering
\subfigure[]{\includegraphics[width=0.38\linewidth]{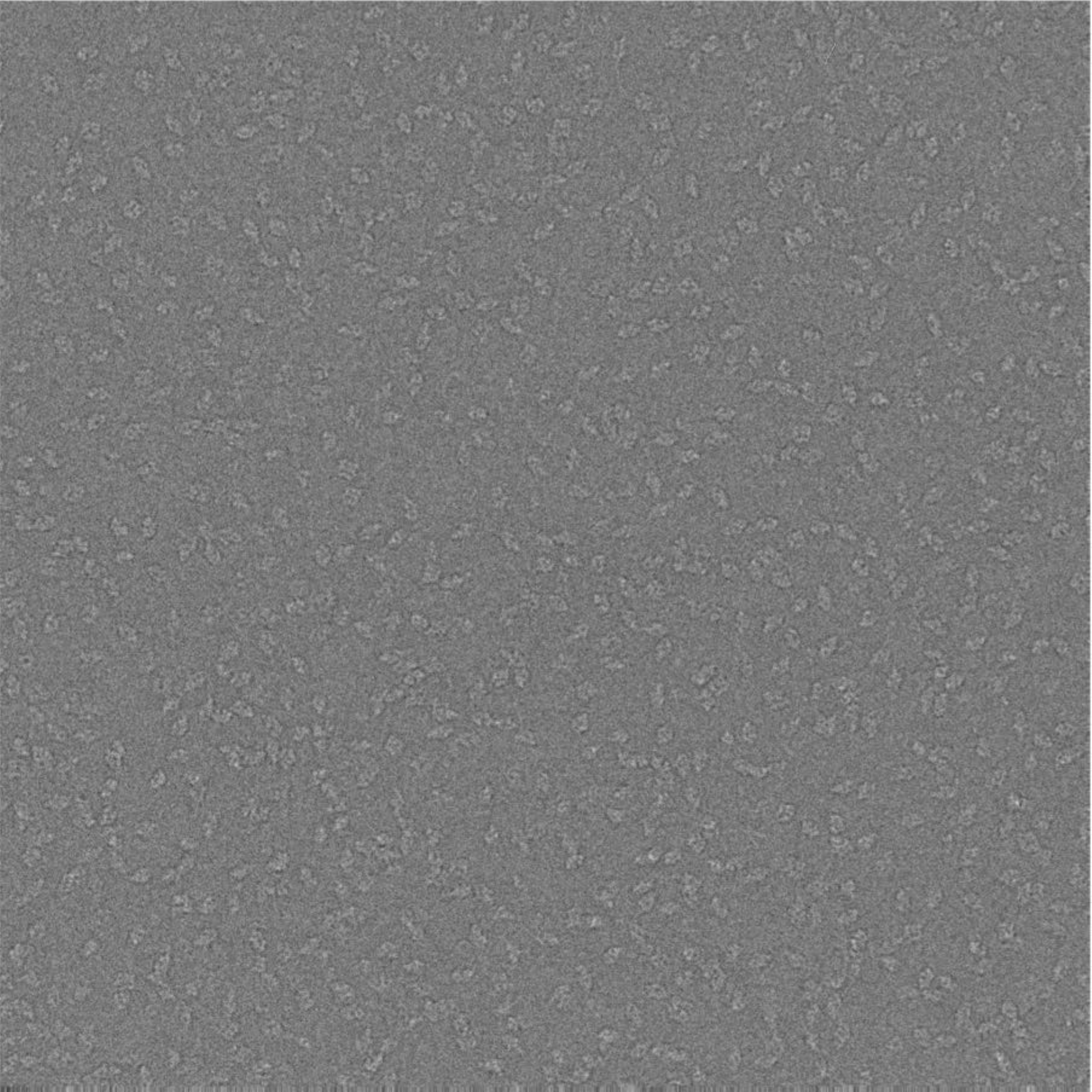}} \hspace{0.25cm}
\subfigure[]{\includegraphics[width=0.38\linewidth]{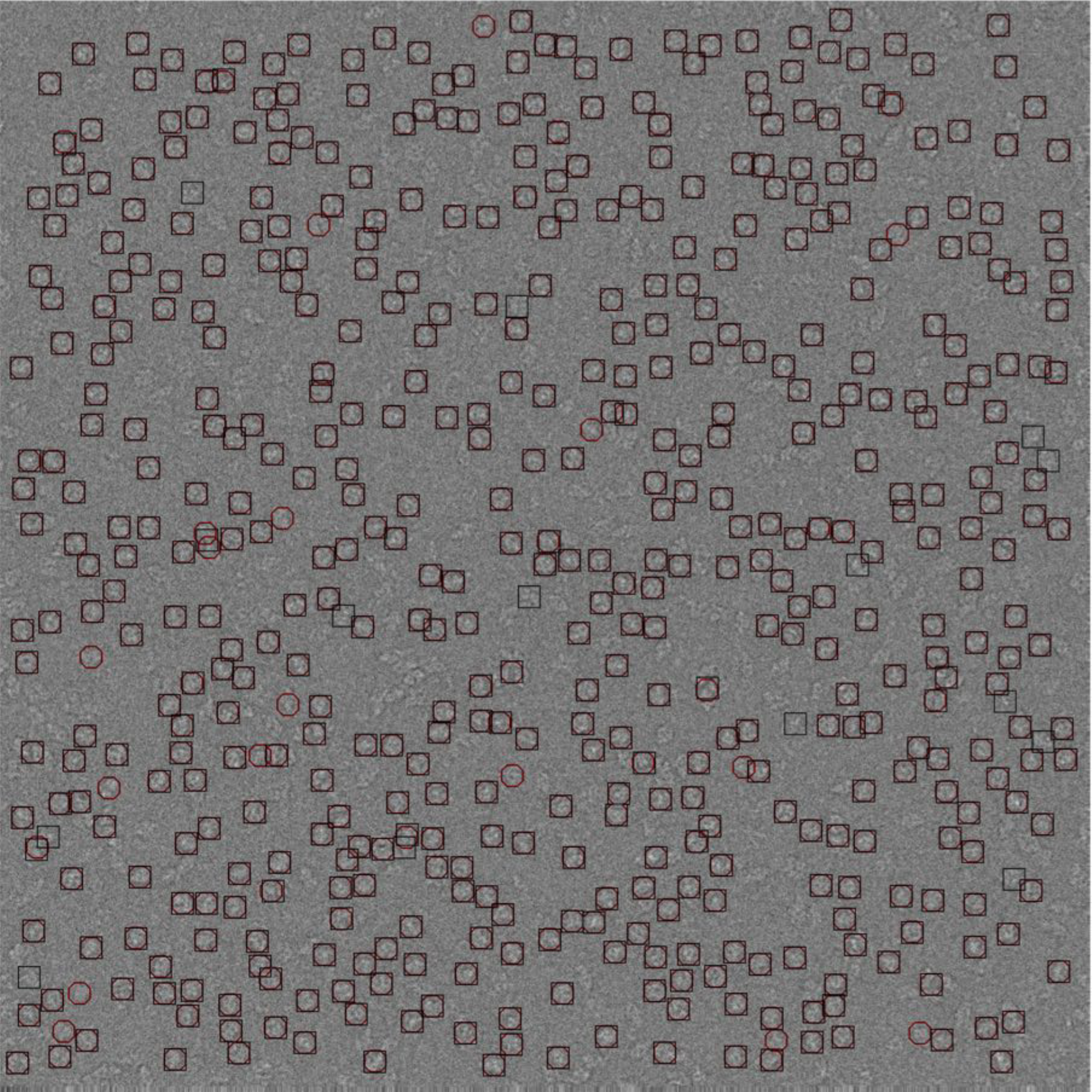}} \hspace{0.25cm}
\caption{{Illustration of picking with and without CTF correction. (a) A CTF-corrected micrograph. CTF estimation was done using CTFFIND4 \citep{ctffind}  
followed by phase-flipping. (b) Particles picked from the original micrograph are surrounded by a black box. 
Particles picked from phase-flipped micrographs are surrounded by a red circle.}}
\label{fig:ctf}
\end{figure*}

Figure \ref{fig:ctf} contains a comparison between our particle picking framework when applied to the micrographs with and without phase-flipping. 
We note that, while most of the picked particles appear in both micrographs, there are slight differences around some of the near-by particles. Thus, 
we recommend this method when applying CTF correction to micrographs before particle picking.

\section{Experimental Results}
\label{sec:experimental}

We present experimental results for the framework presented in this paper. We apply our framework to datasets of  $\beta$-Galactosidase, 
T20S proteasome, 70S ribosome and keyhole limpet hemocyanin (KLH) particles. 

The $\beta$-Galactosidase dataset we use is publicly available from
 EMPIAR (the Electron Microscopy Public Image Archive) \citep{empire} as EMPIAR-10017.\footnote{This dataset was obtained by the  
 FALCON \Romannum{2} direct detector. 
 Another $\beta$-Galactosidase dataset is 
 EMPIAR-10061 \citep{betaGal2}, which was obtained using the 
 K$2$ direct detector. We note the APPLE picker is effective for this dataset as well. For a comparison between 
 FALCON \Romannum{2} and K$2$ direct detectors see \citep{detectors}.} It 
consists of 84 micrographs of $\beta$-Galactosidase.  
The T20S proteasome dataset is publicly available as EMPIAR-10057\footnote{This dataset was obtained using the 
 K$2$ direct detector.} \citep{e10057}. It 
contains $158$ micrographs. The 70S ribosome dataset is available as EMPIAR-10077 \citep{n10077} and contains thousands of micrographs. 
 The KLH dataset we use  \citep{bakeoff, KLH} contains 82 micrographs. 

The experiments are run on a $2.6~\mathrm{GHz}$ Intel Core i7 CPU with four cores and $16~\mathrm{GB}$ of memory.
Our method has also been implemented on a GPU.
It is evaluated using an Nvidia Tesla P100 GPU.

\subsection{$\beta$-Galactosidase}
\label{sec:betaGal}

We ran the suggested framework on a  
 $\beta$-Galactosidase dataset \citep{relion}. 
 We compare the performance of the APPLE picker to the semi-automated particle picker included in RELION. 
 For this comparison, 
 we input the locations of our picked particles into RELION and  
  obtain a 3D reconstruction. 
 We then compare this to the reconstruction obtained by the full RELION pipeline in \citep{relion}.

The $\beta$-Galactosidase micrographs are obtained using a FALCON II detector. Thus, each micrograph is 
of size  $4096 \times 4096$ pixels. 
The outermost pixels in these micrographs do not contain 
important information. In light of this, when running the APPLE picker on these micrographs, we discard the $100$ outermost pixels.  
In addition, for  runtime reduction, 
each dimension of the micrograph is reduced to half its original size, bringing the micrograph in total to a 
quarter of its original size. This is done by averaging adjacent pixels, also known as binning.  

\begin{figure*}
\begin{center}
\includegraphics[width=0.31\linewidth]{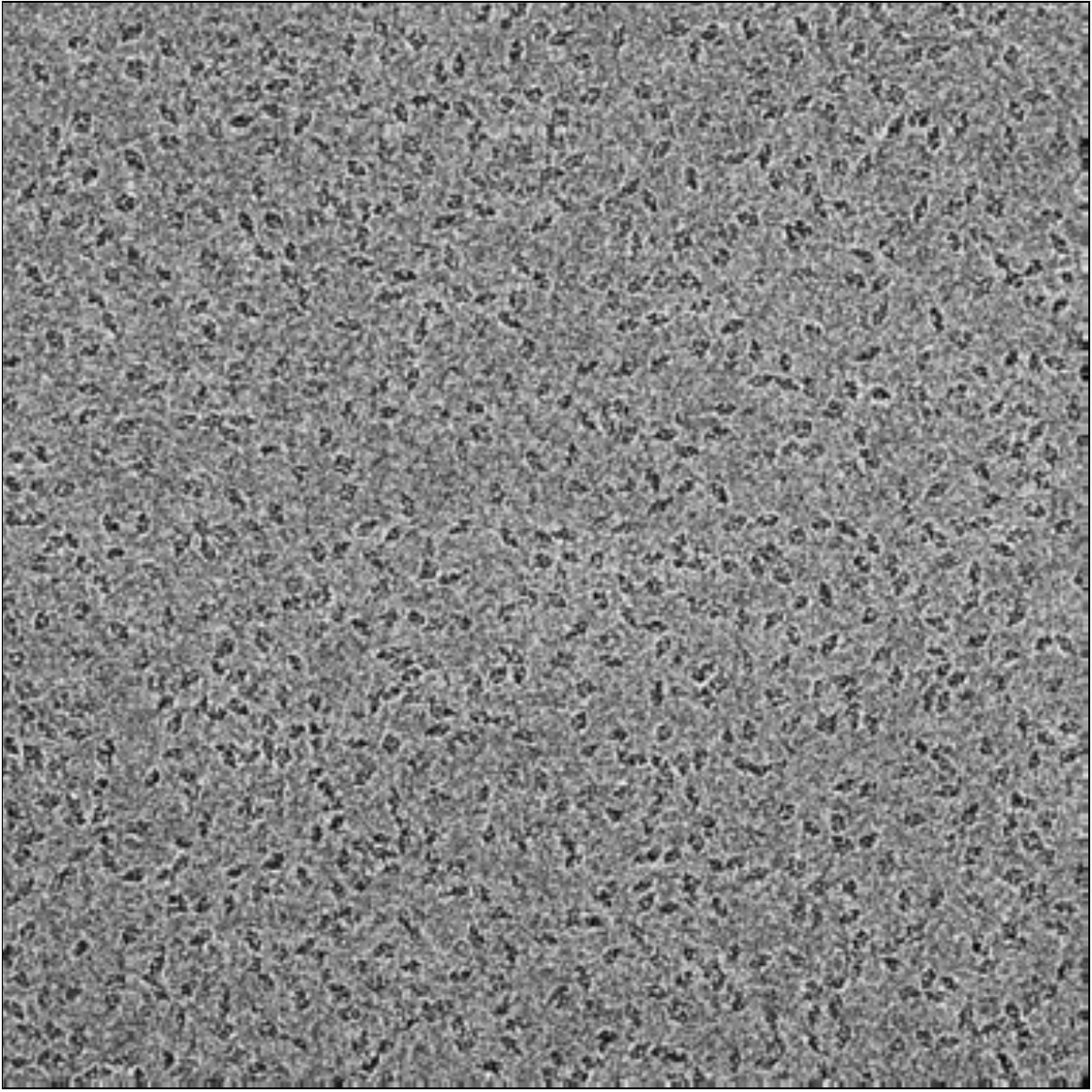}
\includegraphics[width=0.31\linewidth]{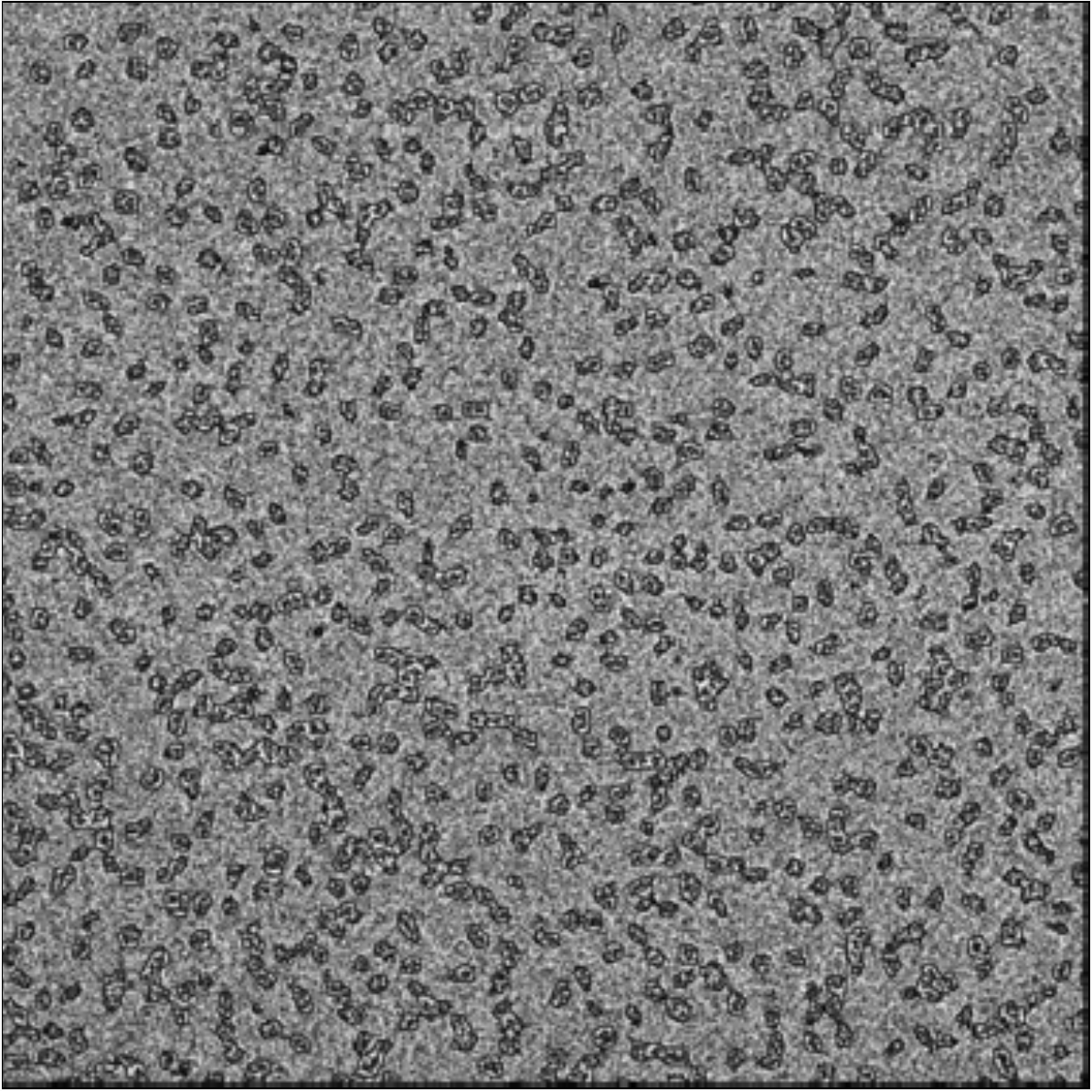}
\includegraphics[width=0.31\linewidth]{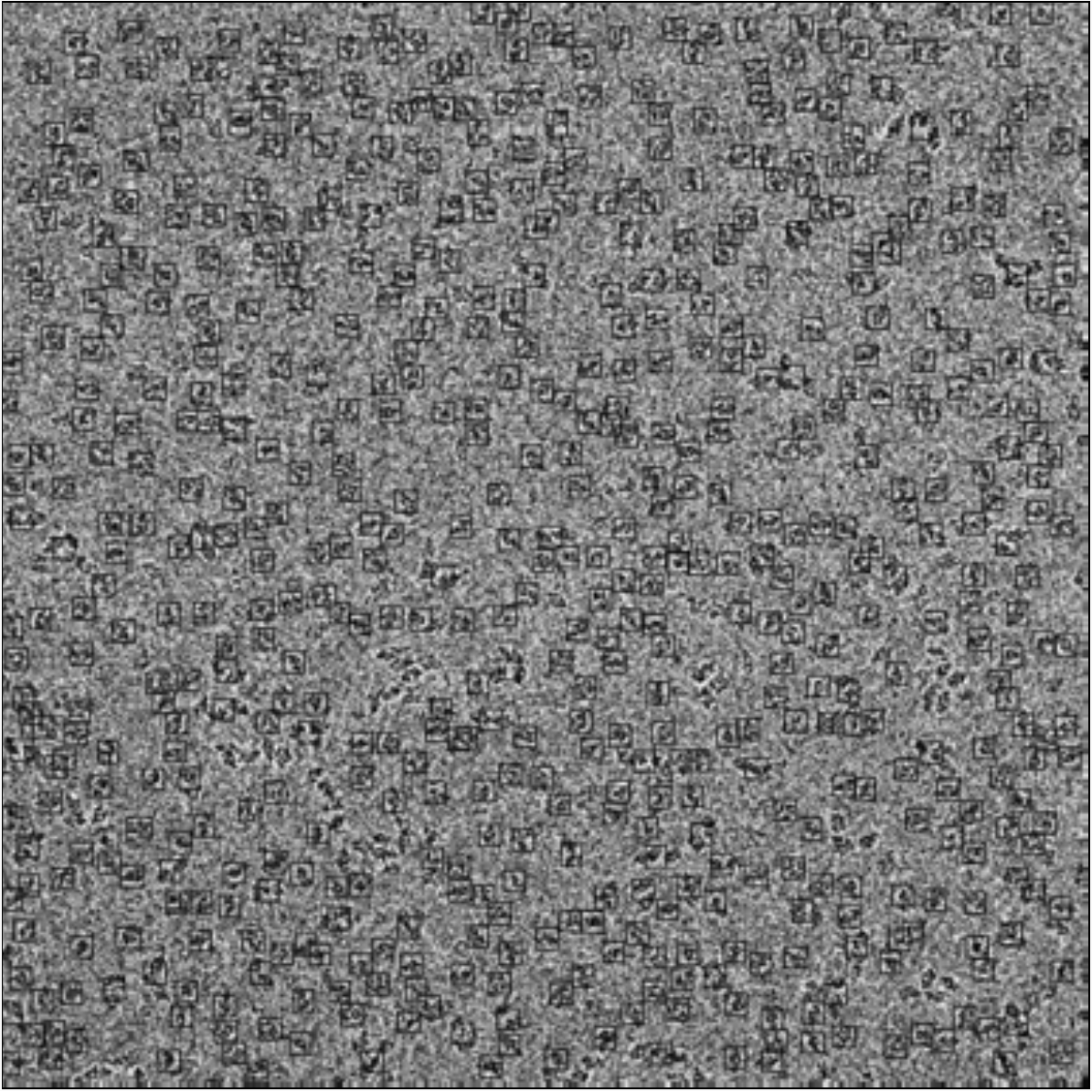}
\includegraphics[width=0.31\linewidth]{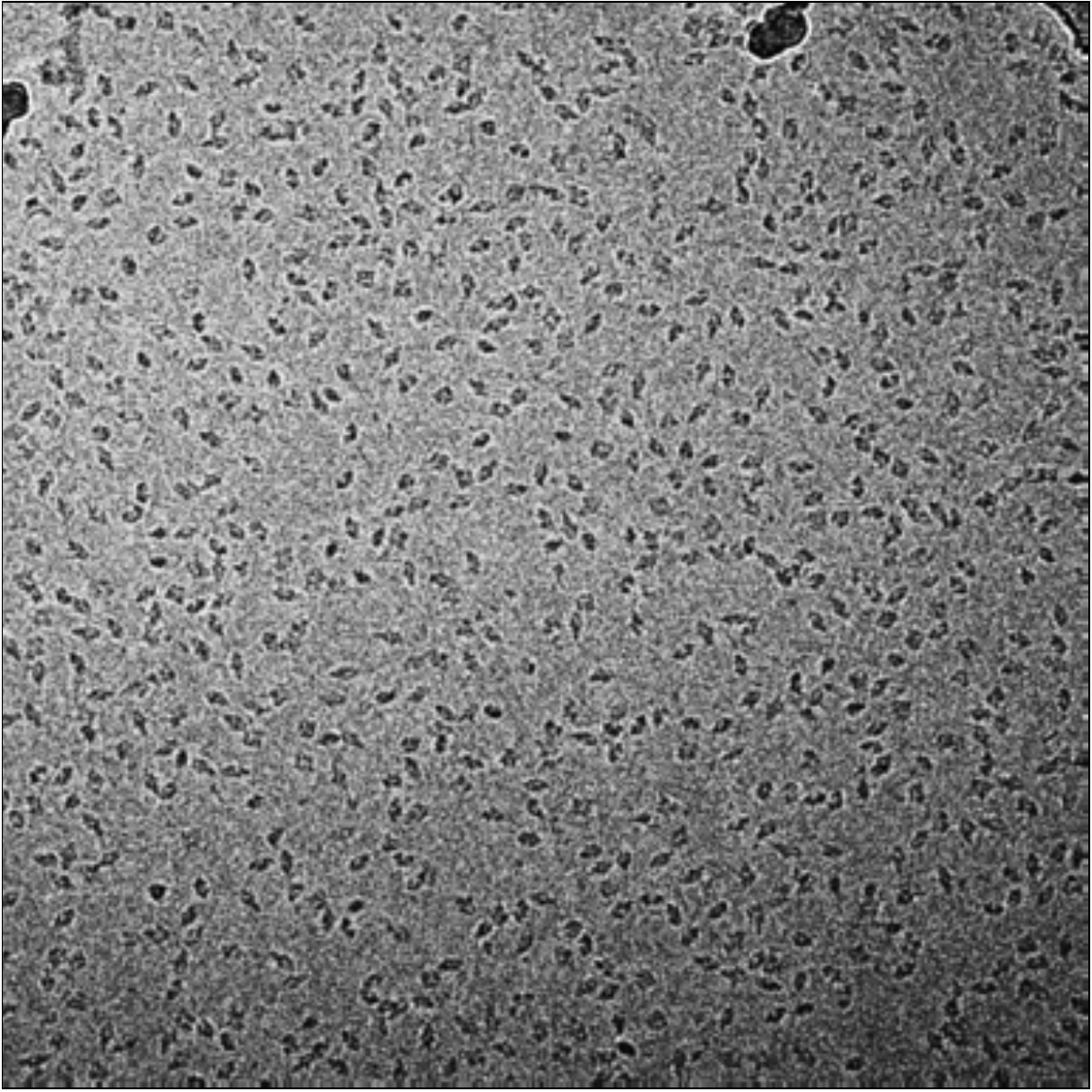}
\includegraphics[width=0.31\linewidth]{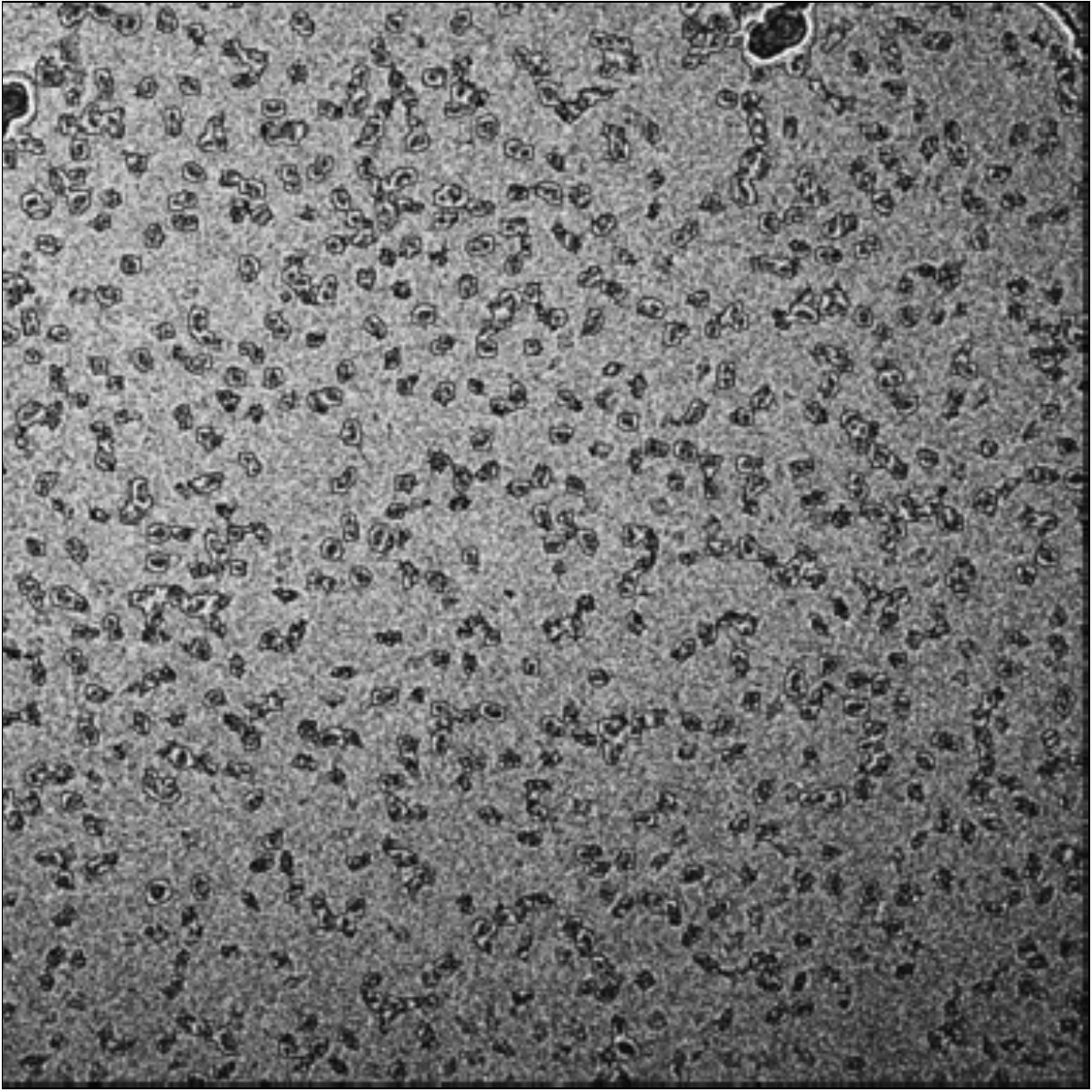}
\includegraphics[width=0.31\linewidth]{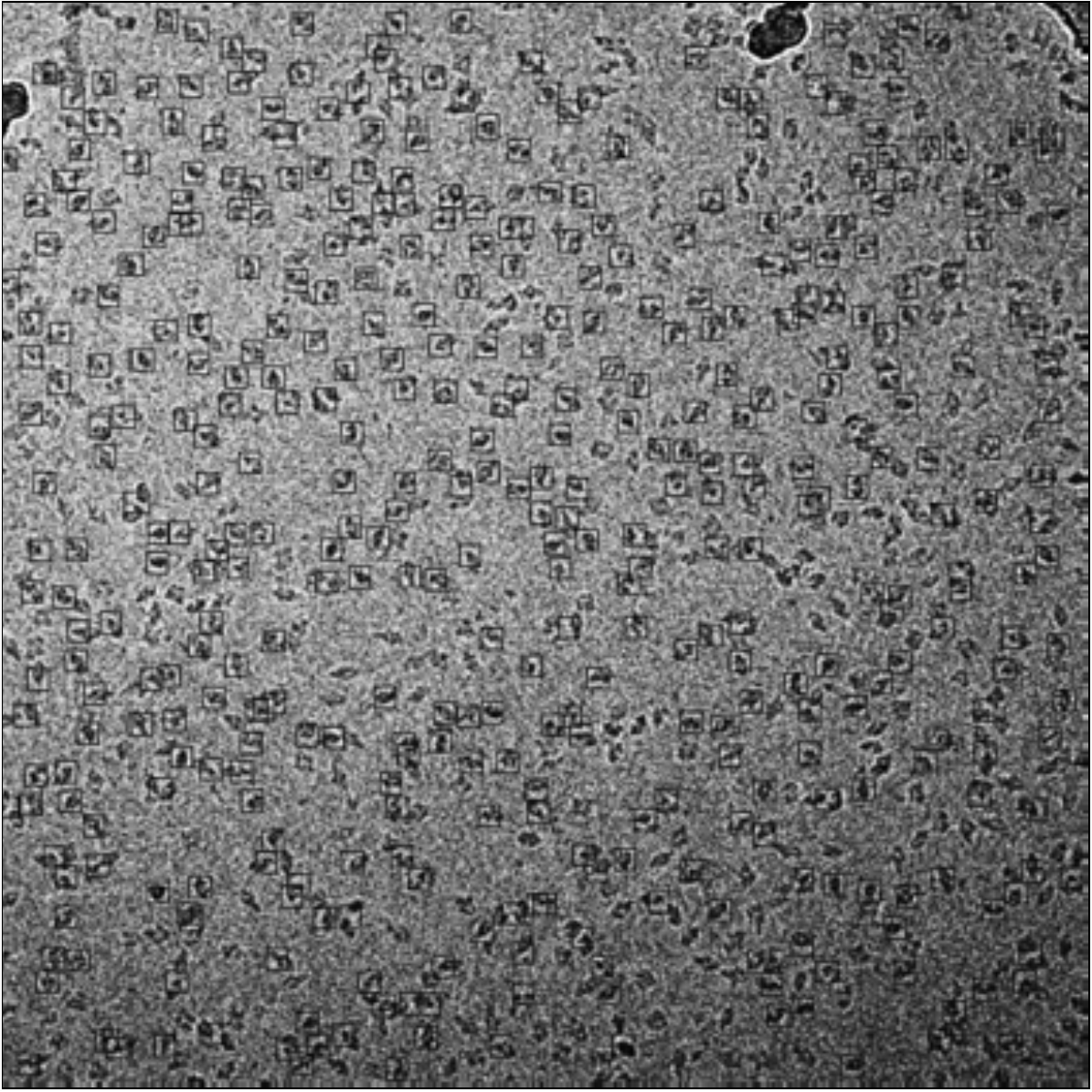}
\includegraphics[width=0.31\linewidth]{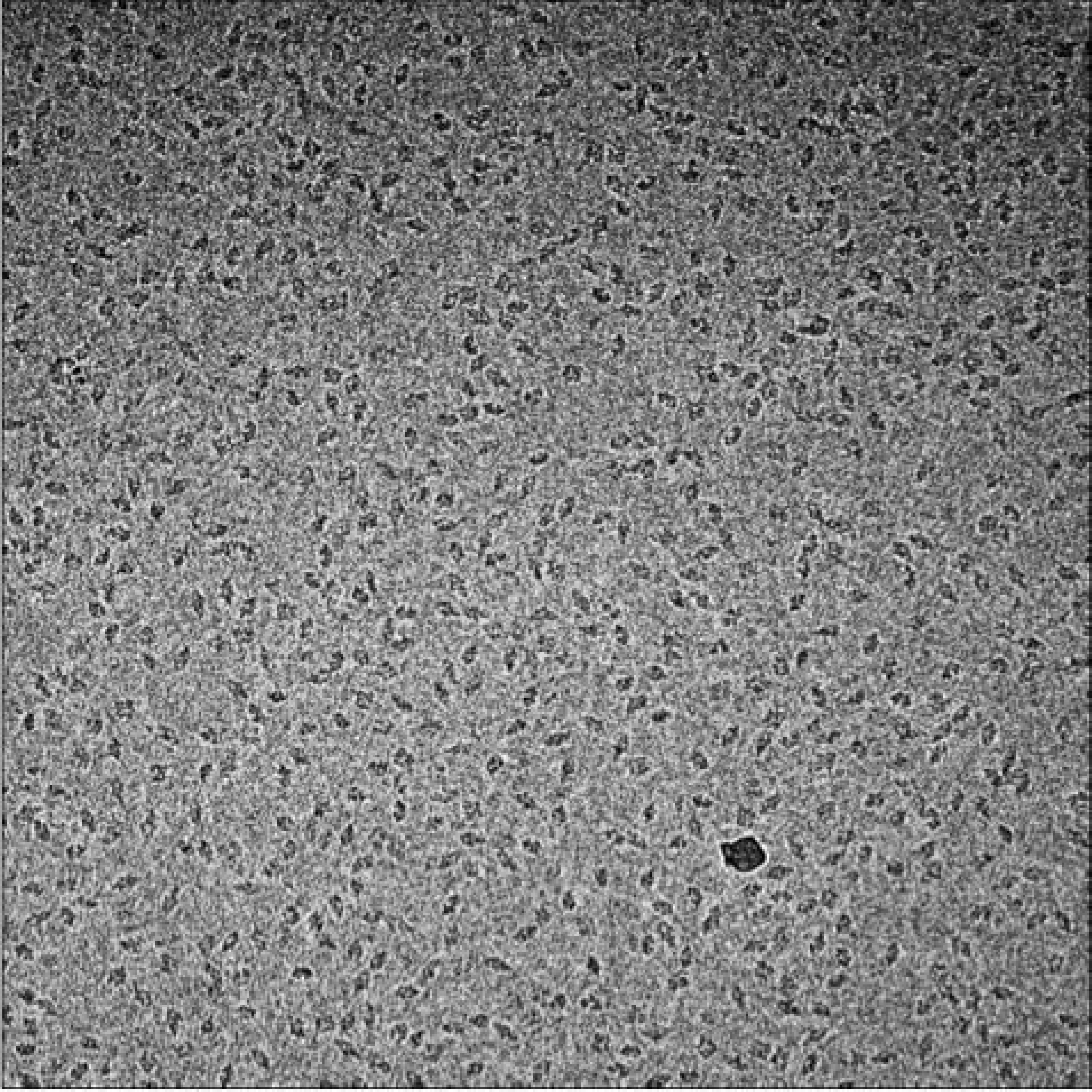}
\includegraphics[width=0.31\linewidth]{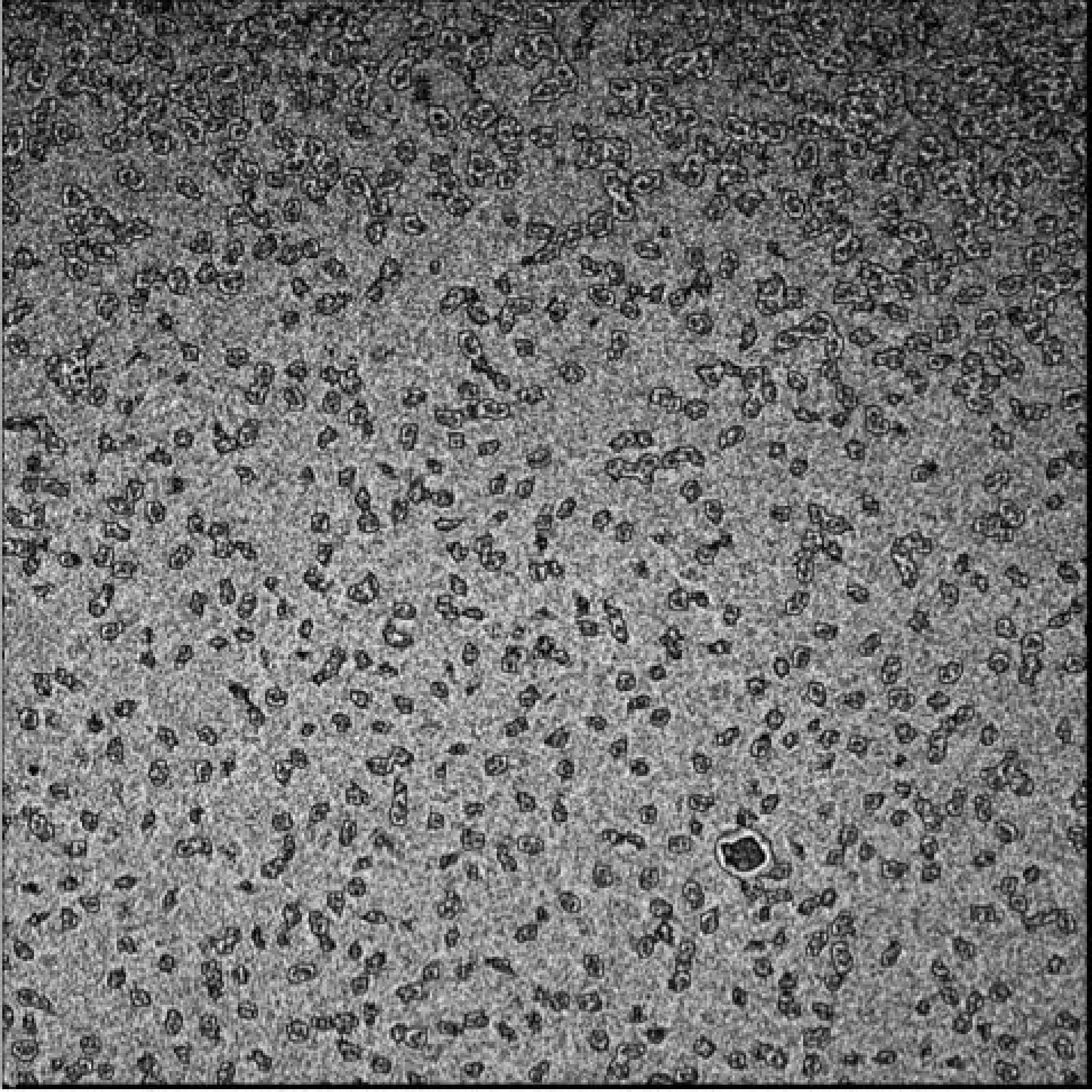}
\includegraphics[width=0.31\linewidth]{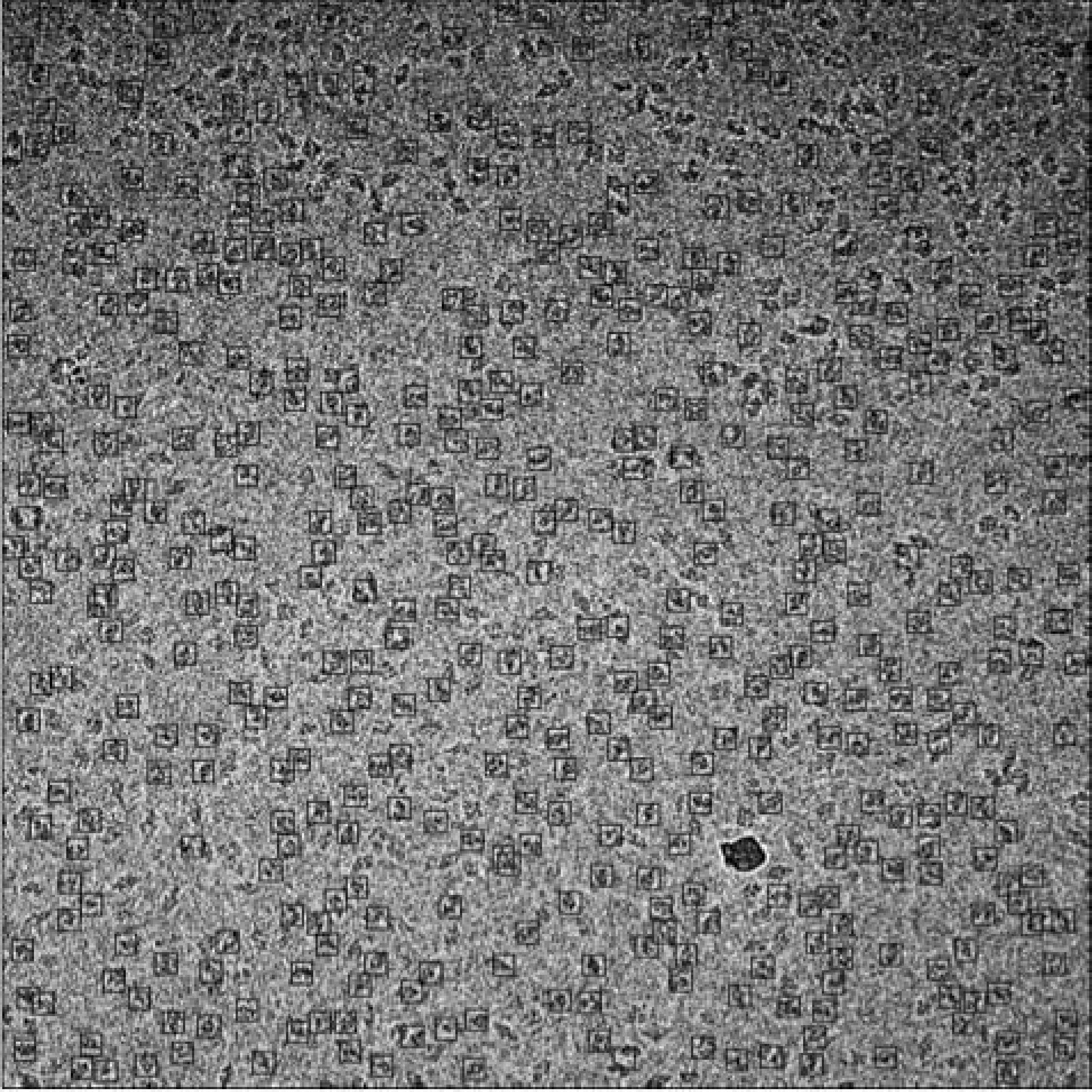}
\caption{Picked particles of sample $\beta$-Galactosidase micrographs. The micrographs are presented in the 
left column. Classification results are presented in the center. The picked particles are on the right.}
\label{fig:10017}
\end{center}
\end{figure*}

Each query and reference image extracted from the reduced micrograph is of size $26 \times 26$ and each container is of size $225 \times 225$. 
For classifier training we suggest to use $\tau_1 =  3\%$  and $\tau_2 = 55\%$  to determine the training set.  
We set the bandwidth of the kernel function for the SVM classifier and its slack parameter both to $1$. 
Examples of results for the APPLE picker are presented in Figure 
\ref{fig:10017}.

For the purpose of evaluating our framework, 
we perform a 3D reconstruction of the particle and compare to the reconstruction of \citep{relion} where 
the particle picking was done based on 
$2555$ manually selected particles. From these particles, $25$ class averages were computed and $10$ were manually chosen. The RELION 
particle picker then picked $52495$ particles. Of these, $4185$ particles were discarded according to Z-scores. After the class averaging step 
$42755$ particles were selected. The reported resolution in  \citep{relion} is $4.2$ \r{A}.

In contrast, we use $32997$ particles selected by the APPLE picker. We enter them into the RELION pipeline and 
begin the reconstruction from our particles. After the 2D class averaging step $15198$ particles 
were selected.  The 3D reconstruction using RELION (including CTF correction using the wrapper for CTFFIND4 \citep{ctffind}) 
 reached a  gold-standard FSC resolution of 
$4.5$ \r{A}.\footnote{{We repeated this experiment for CTF-corrected micrographs and achieved the same resolution. 
Another experiment we performed was 3D reconstruction from the manually selected particles available with the $\beta$-Galactosidase dataset. 
While the accuracy of this was reported in \citep{relion} to be $4.2$ \r{A}, we achieve an improvement of $0.05$ \r{A} resolution over the 3D 
reconstruction from the APPLE picked particles.}} 

\begin{figure*}
\begin{center}
\includegraphics[width=0.32\linewidth]{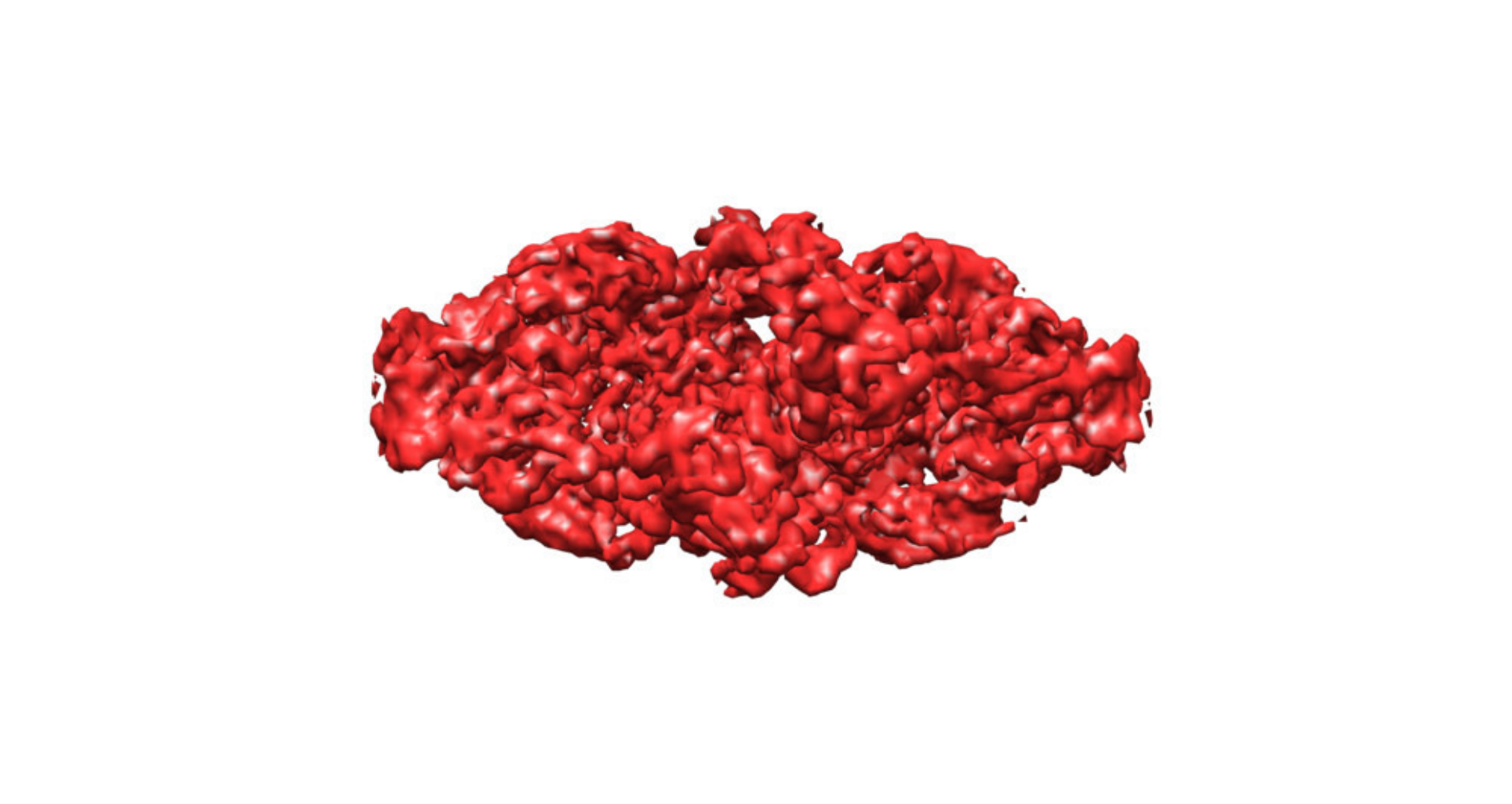}
\includegraphics[width=0.32\linewidth]{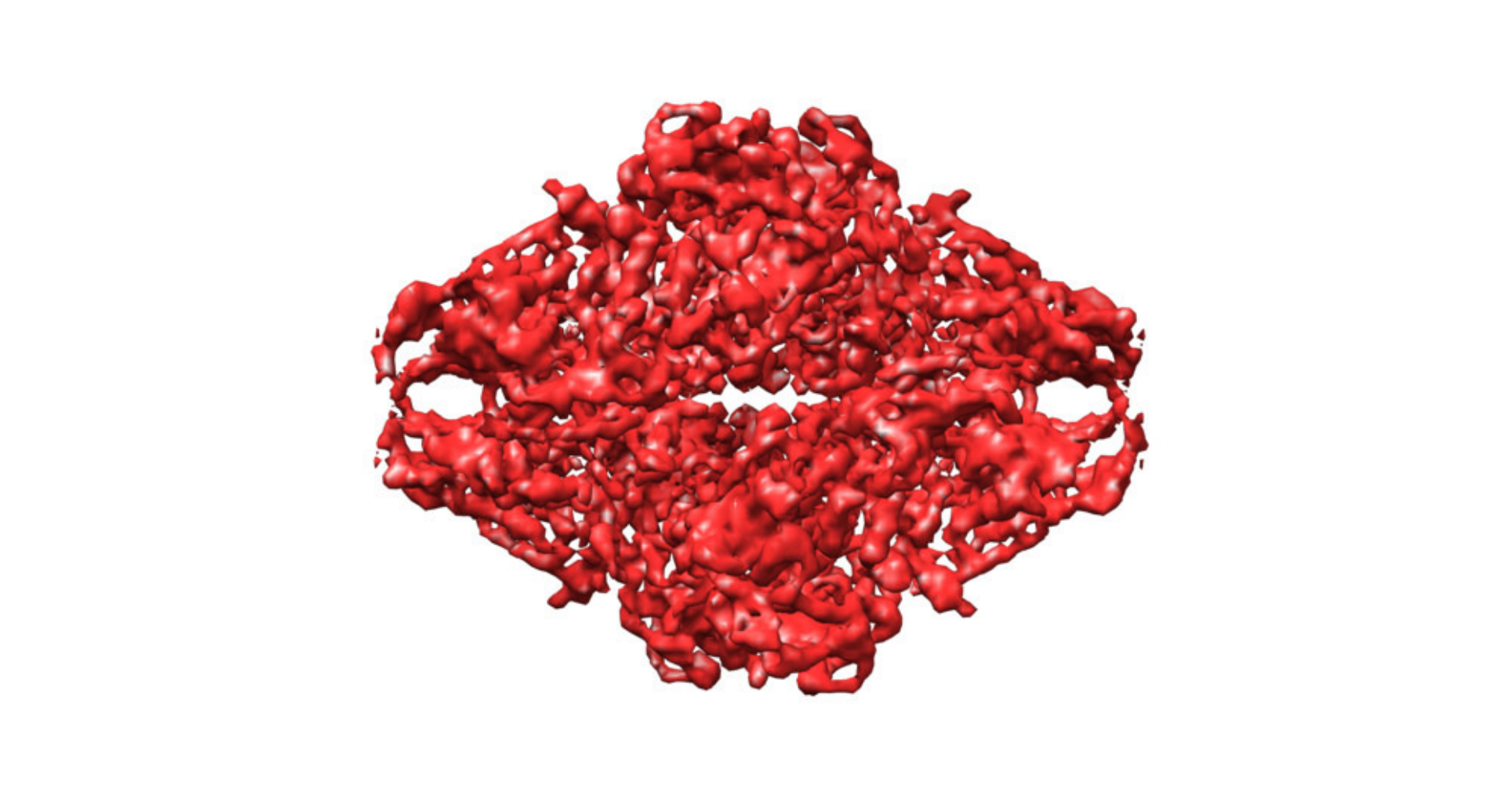}
\includegraphics[width=0.32\linewidth]{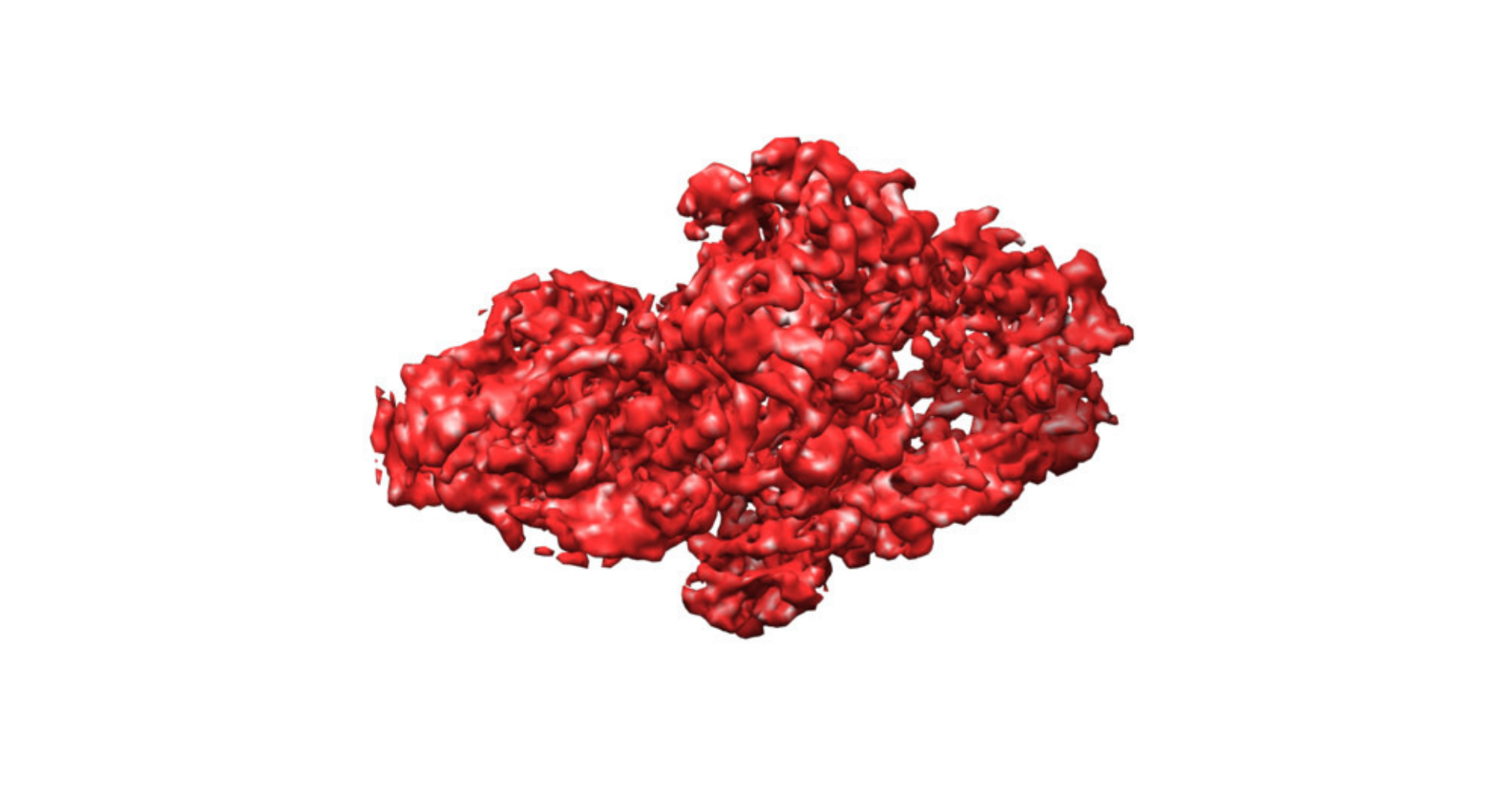}\\
\includegraphics[width=0.32\linewidth]{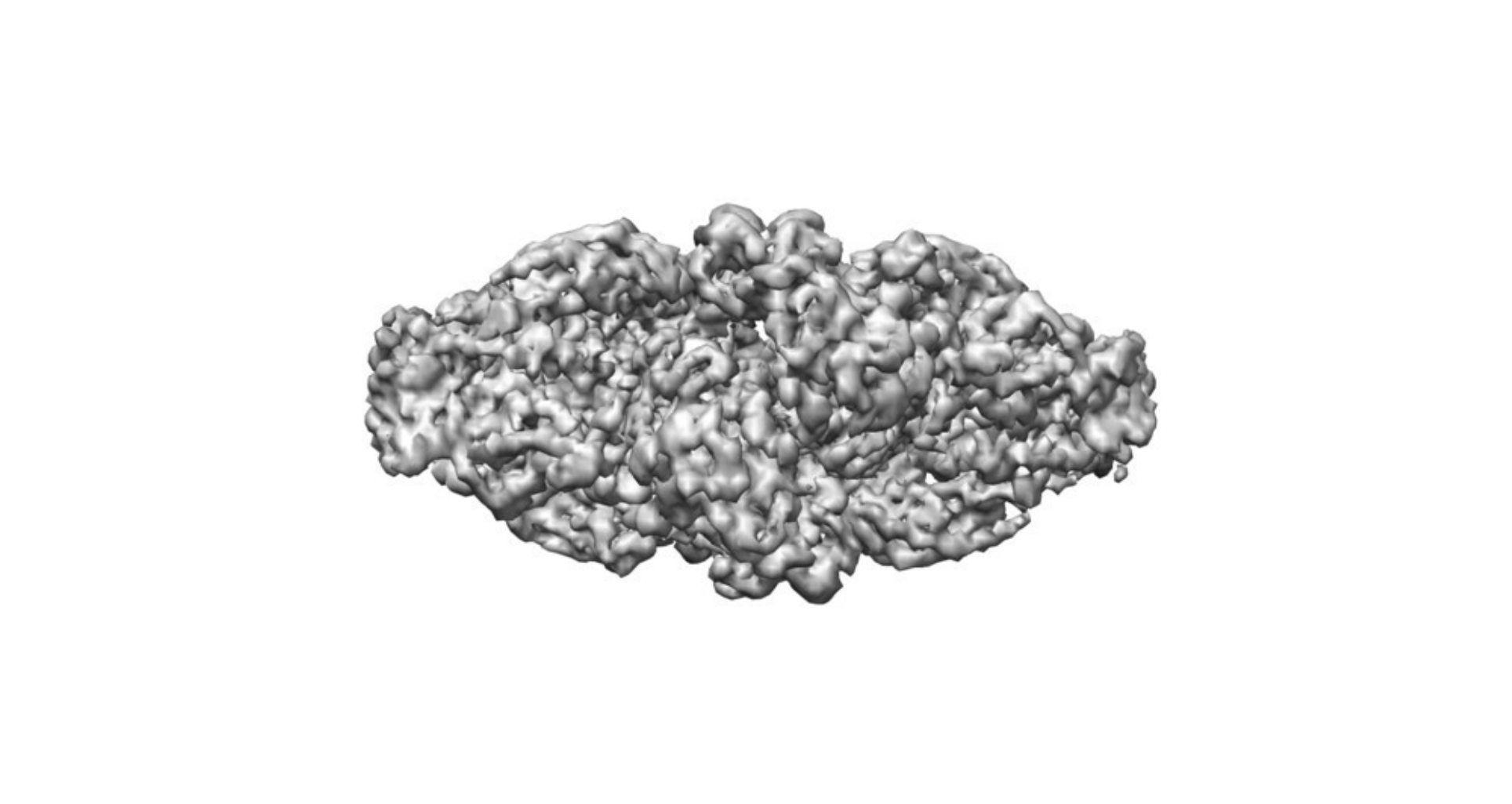}
\includegraphics[width=0.32\linewidth]{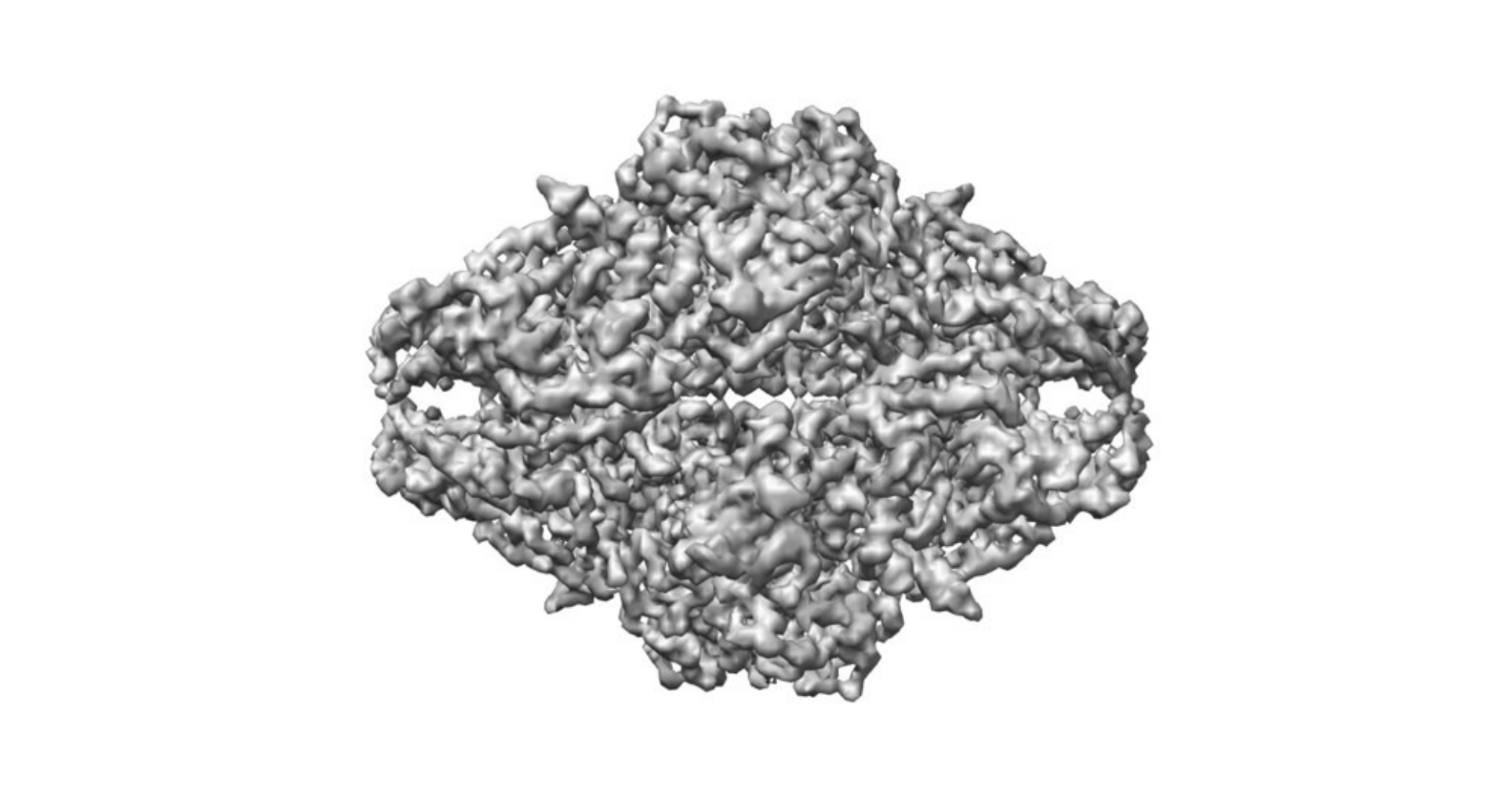}
\includegraphics[width=0.32\linewidth]{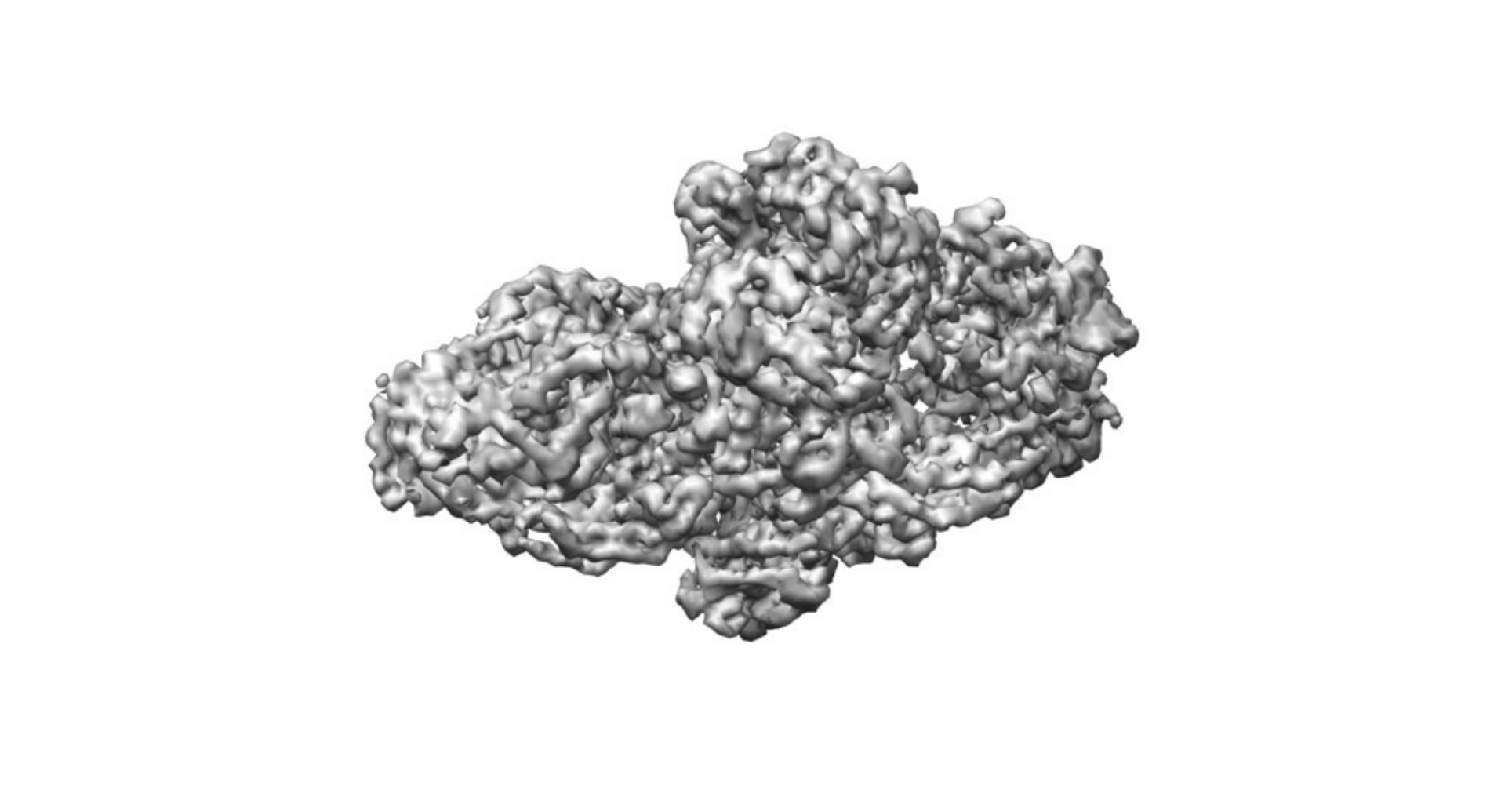}
\caption{Comparison between the APPLE picker and the RELION semi-automatic particle picker. 
On the top are {surface views} of the 3D reconstruction created in RELION from the APPLE picks and 
obtained in UCSF Chimera.
On the bottom are {surface views} of the 3D views detailed in \citep{relion}. We use the reference volume published on EMDB (EMD-2824) and obtain 
the views in UCSF Chimera.}
\label{fig:compare}
\end{center}
\end{figure*}

We present a comparison of {surface views from}  the model reconstructed from particles selected by the APPLE-Picker (in red) 
and the reconstructed model by \citep{relion} in 
Figure \ref{fig:compare}. These renderings were done in UCSF Chimera\footnote{http://www.rbvi.ucsf.edu/chimera.} \citep{chimera1}.

Runtime for a single micrograph is approximately two minutes when running on the CPU. Thus, the entire dataset can be processed in under $3$ hours.
The GPU implementation, on the other hand, takes approximately $8$ seconds. In other words, the APPLE picker 
processes all $84$ micrograph in under $15$ minutes. This is significantly faster than manual picking.

\subsection{T20S proteasome}

The T20S proteasome \citep{e10057} dataset is publicly available as EMPIAR-10057. Its micrographs were acquired using a 
K$2$ direct detector. Thus, they are sized $3838 \times 3710$ pixels.  Unlike the dataset presented in Section \ref{sec:betaGal}, 
this dataset contain elongated particles. 

\begin{figure*}
\begin{center}
\includegraphics[width=0.31\linewidth]{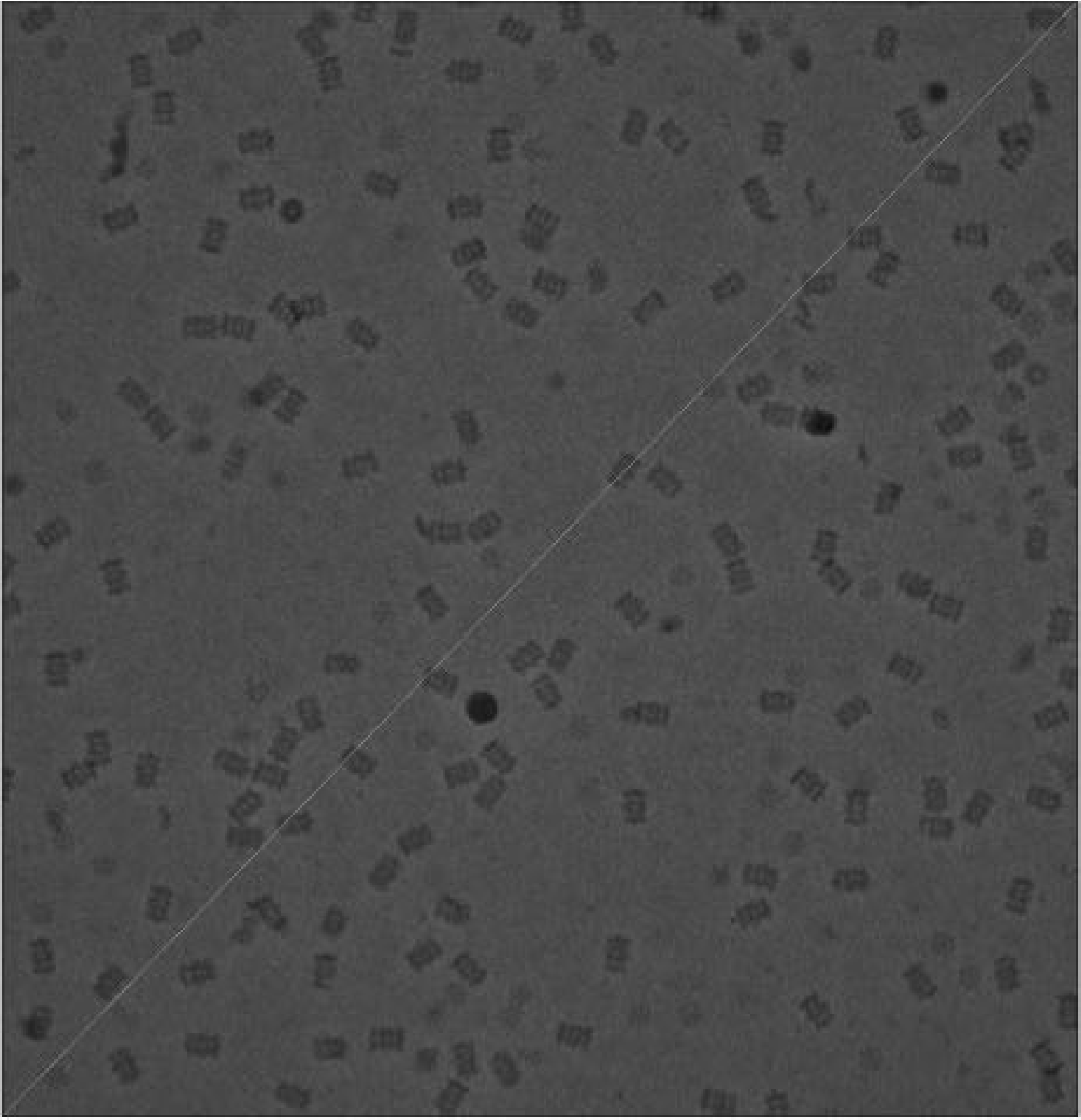}
\includegraphics[width=0.31\linewidth]{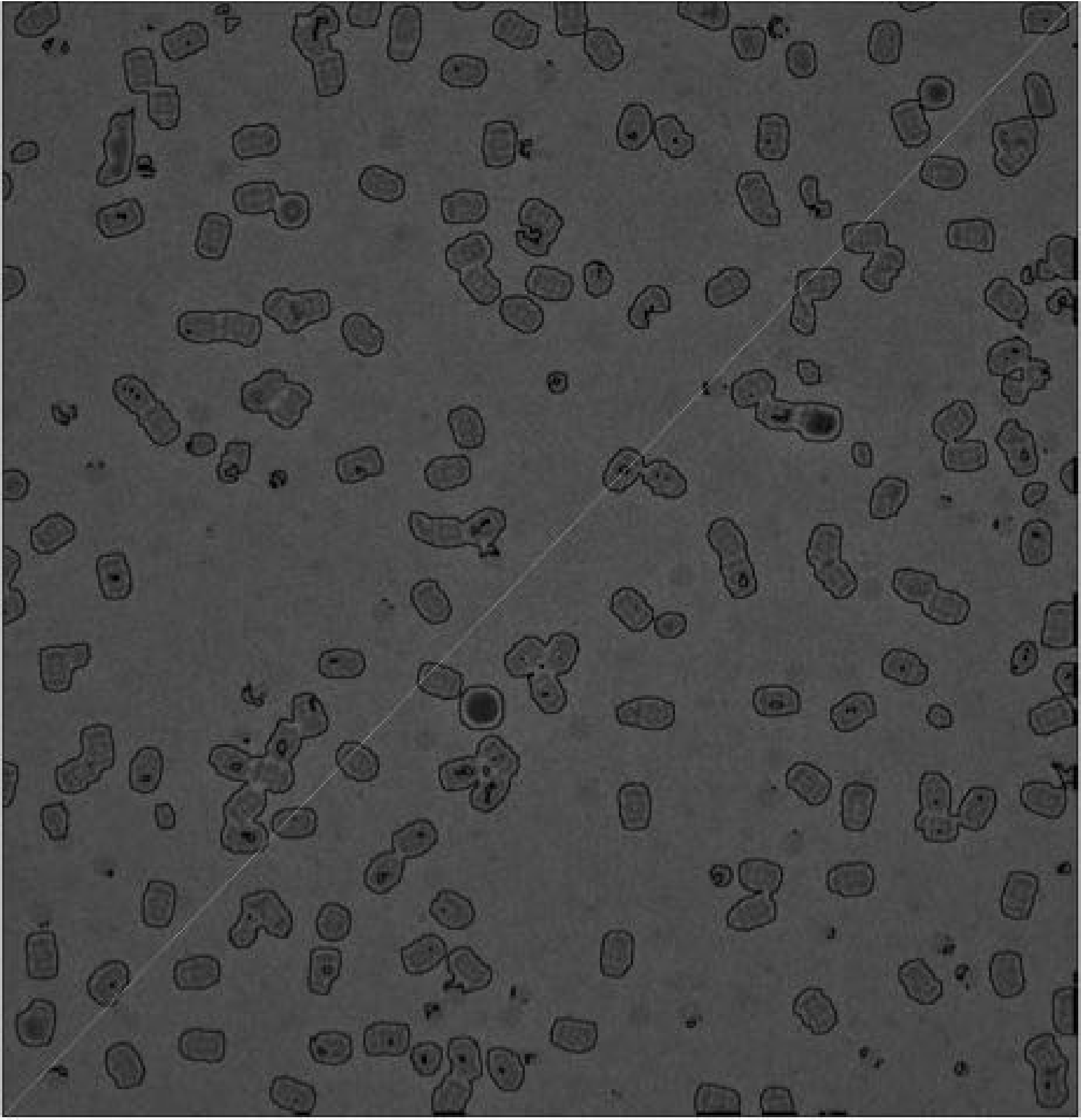}
\includegraphics[width=0.31\linewidth]{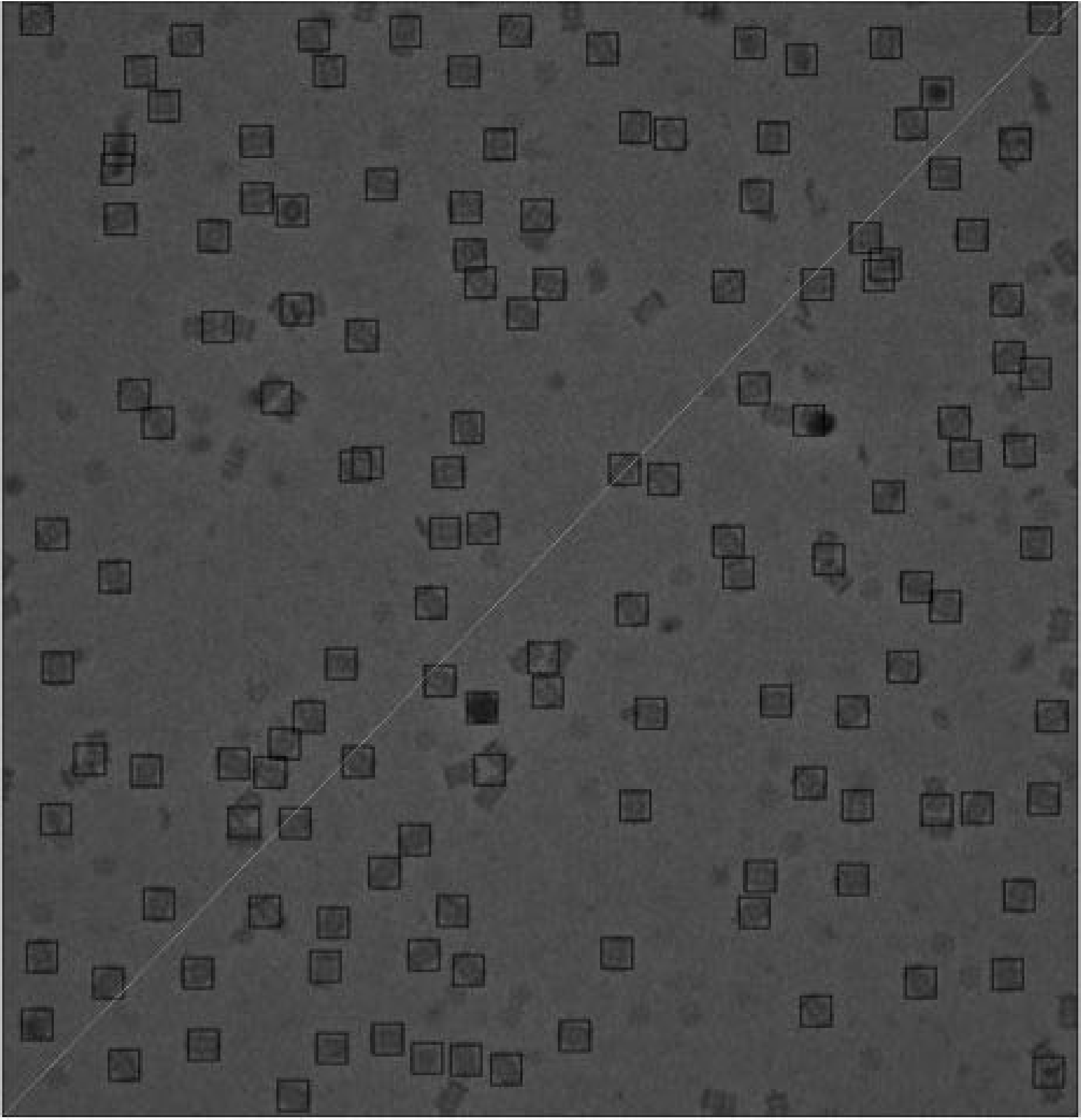}
\includegraphics[width=0.31\linewidth]{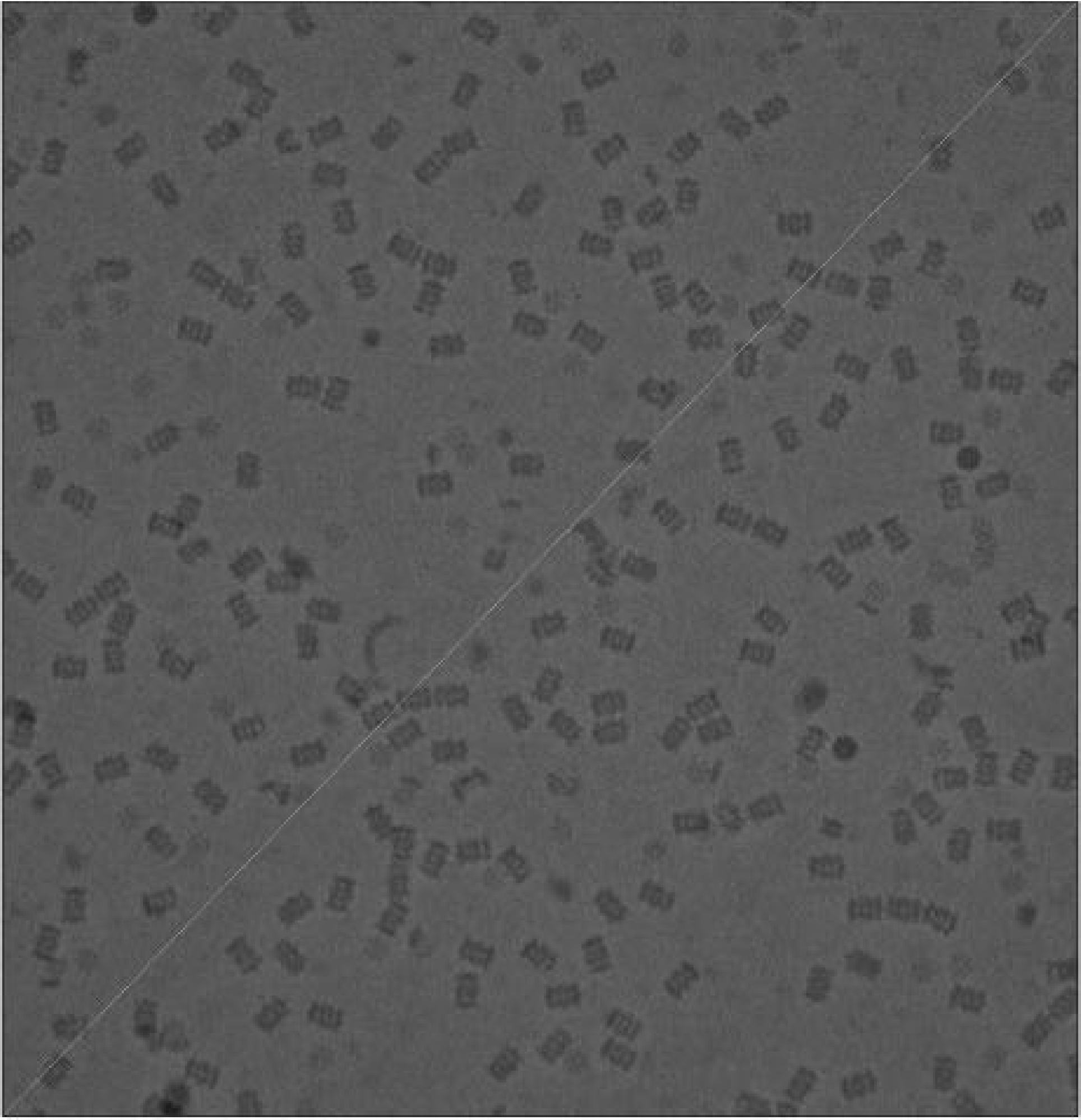}
\includegraphics[width=0.31\linewidth]{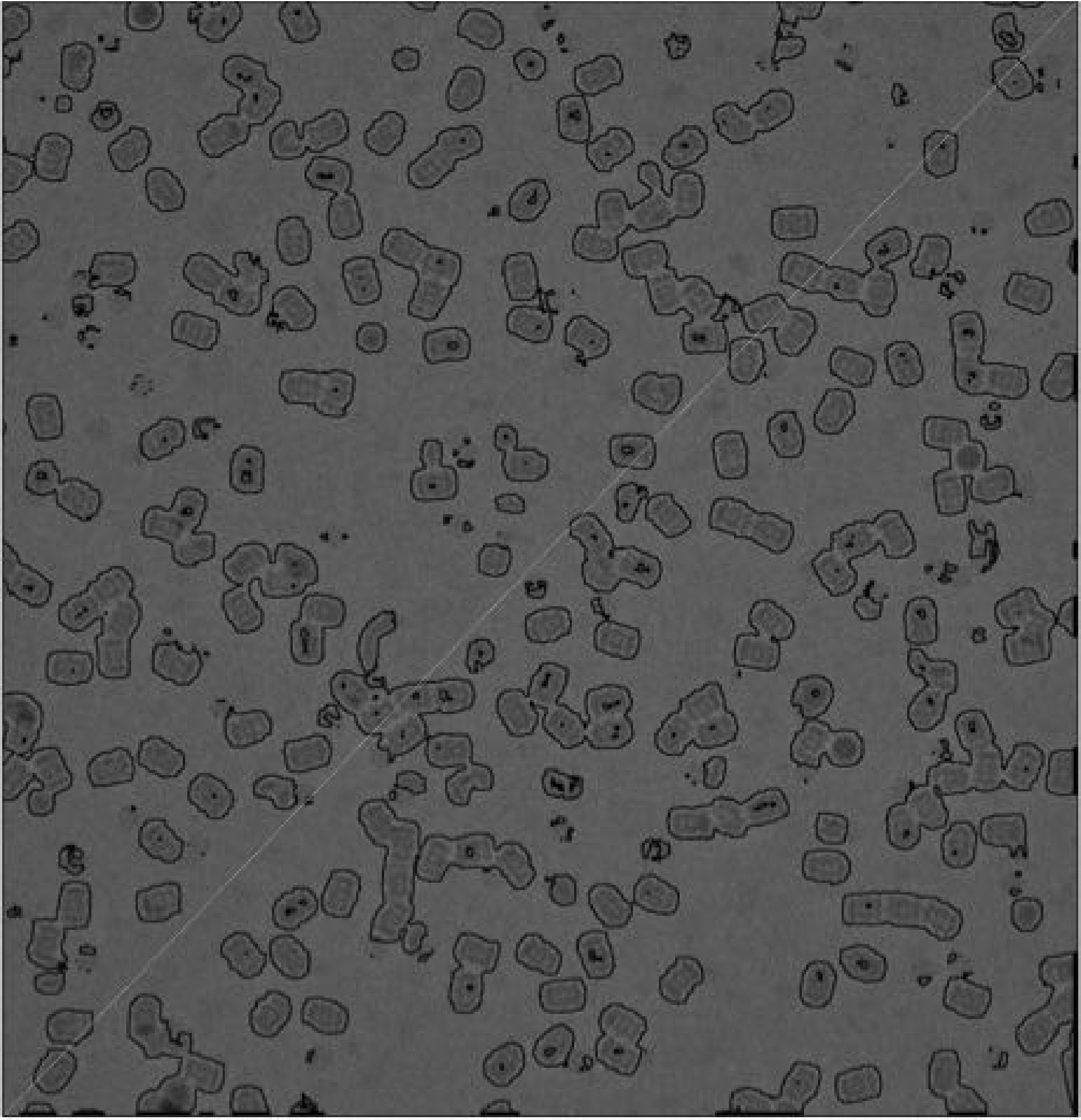}
\includegraphics[width=0.31\linewidth]{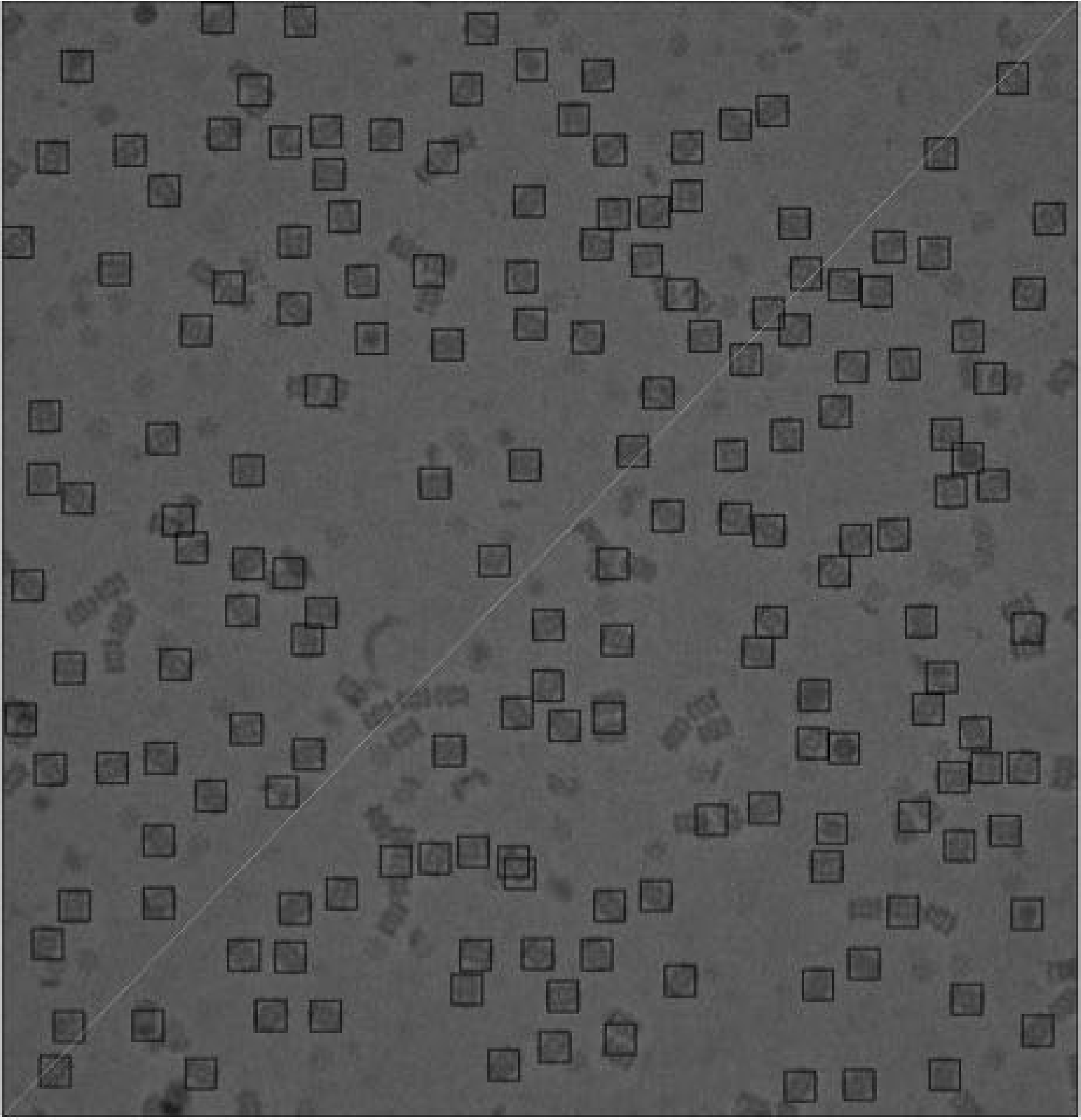}
\includegraphics[width=0.31\linewidth]{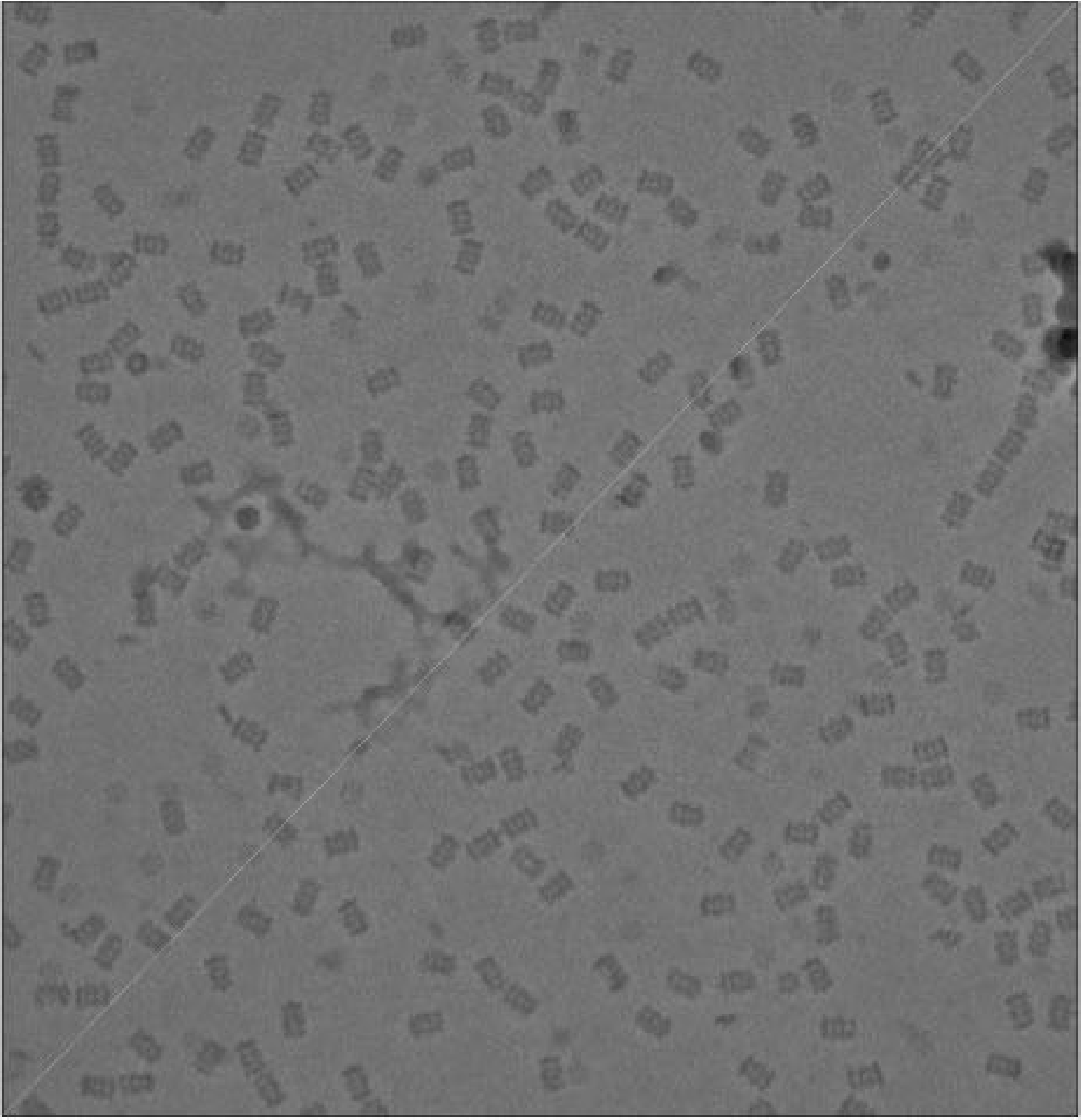}
\includegraphics[width=0.31\linewidth]{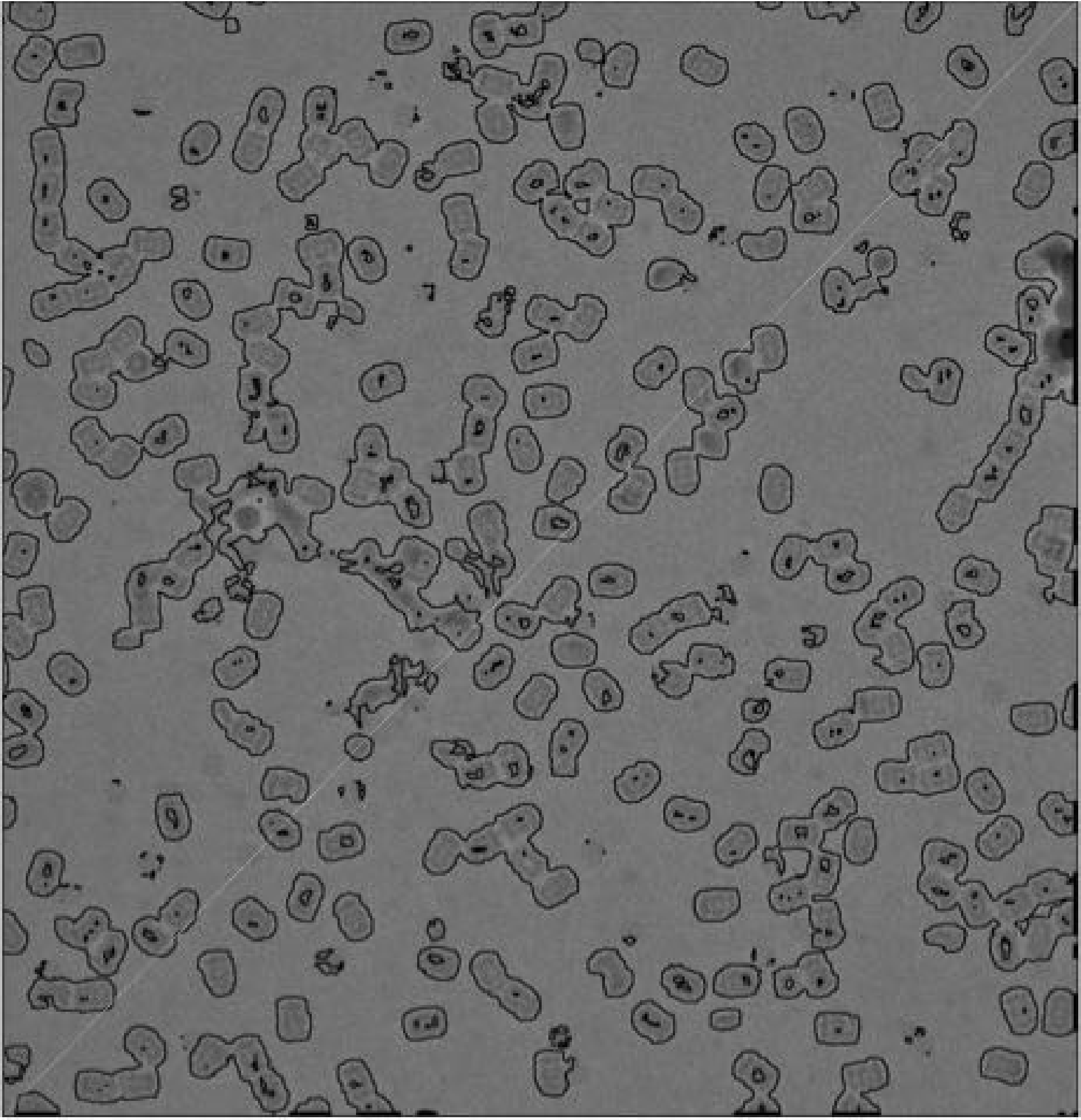}
\includegraphics[width=0.31\linewidth]{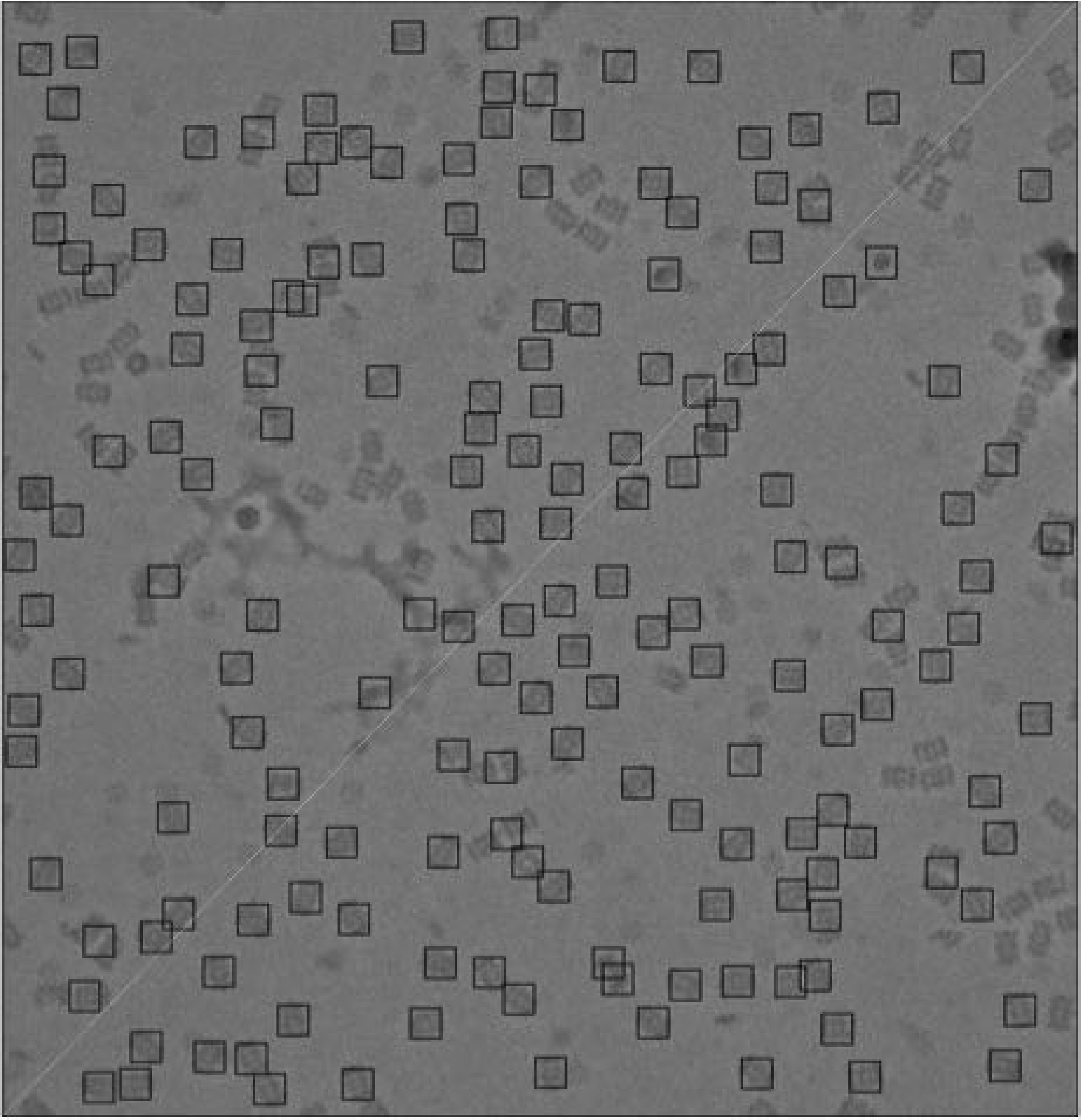}
\caption{Picked particles of sample T20S proteasome micrographs. The micrographs are presented in the 
left column. Classification results are presented in the center. The picked particles are on the right.}
\label{fig:10057}
\end{center}
\end{figure*}

Once again, we use binning to reduce the size of the micrographs. Each query and reference image extracted 
from the reduced micrograph is of size $24 \times 24$. We use the same container size,  $\tau_1$, $\tau_2$ and 
SVM classifier parameters as reported in Section \ref{sec:betaGal}. 
Examples of results for the APPLE picker are presented in Figure 
\ref{fig:10057}. 

{We first corrected for motion using unblur \citep{unblur}. We applied the APPLE picker to the motion-corrected micrographs and 
extracted $21791$ particles. 
These particles were  entered into the RELION pipeline. After the class averaging step $15252$ particles 
were selected.  The 3D reconstruction of RELION reached a  gold-standard FSC resolution of $3.4$ \r{A}.}

{We present a comparison of surface views from  the model reconstructed from particles selected by the APPLE picker (in red) 
and the reconstructed model by \citep{e10057} in 
Figure \ref{fig:compare}. These renderings were done in UCSF Chimera \citep{chimera1}.}

Run time is approximately $90$ seconds per micrograph when running on a CPU, or $7$ seconds per micrograph when running on the GPU.

\begin{figure*}
\begin{center}
\includegraphics[width=0.32\linewidth]{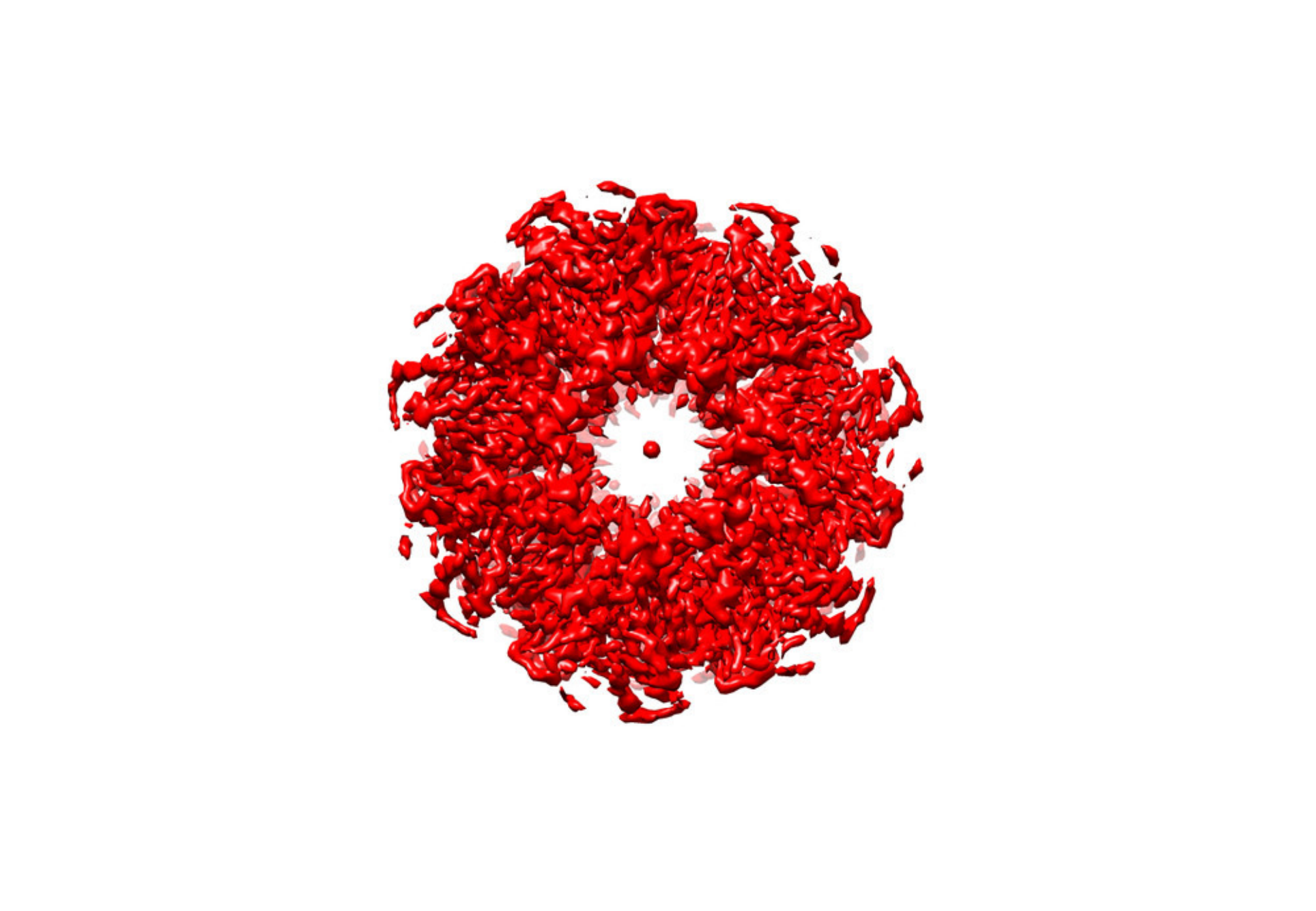}
\includegraphics[width=0.32\linewidth]{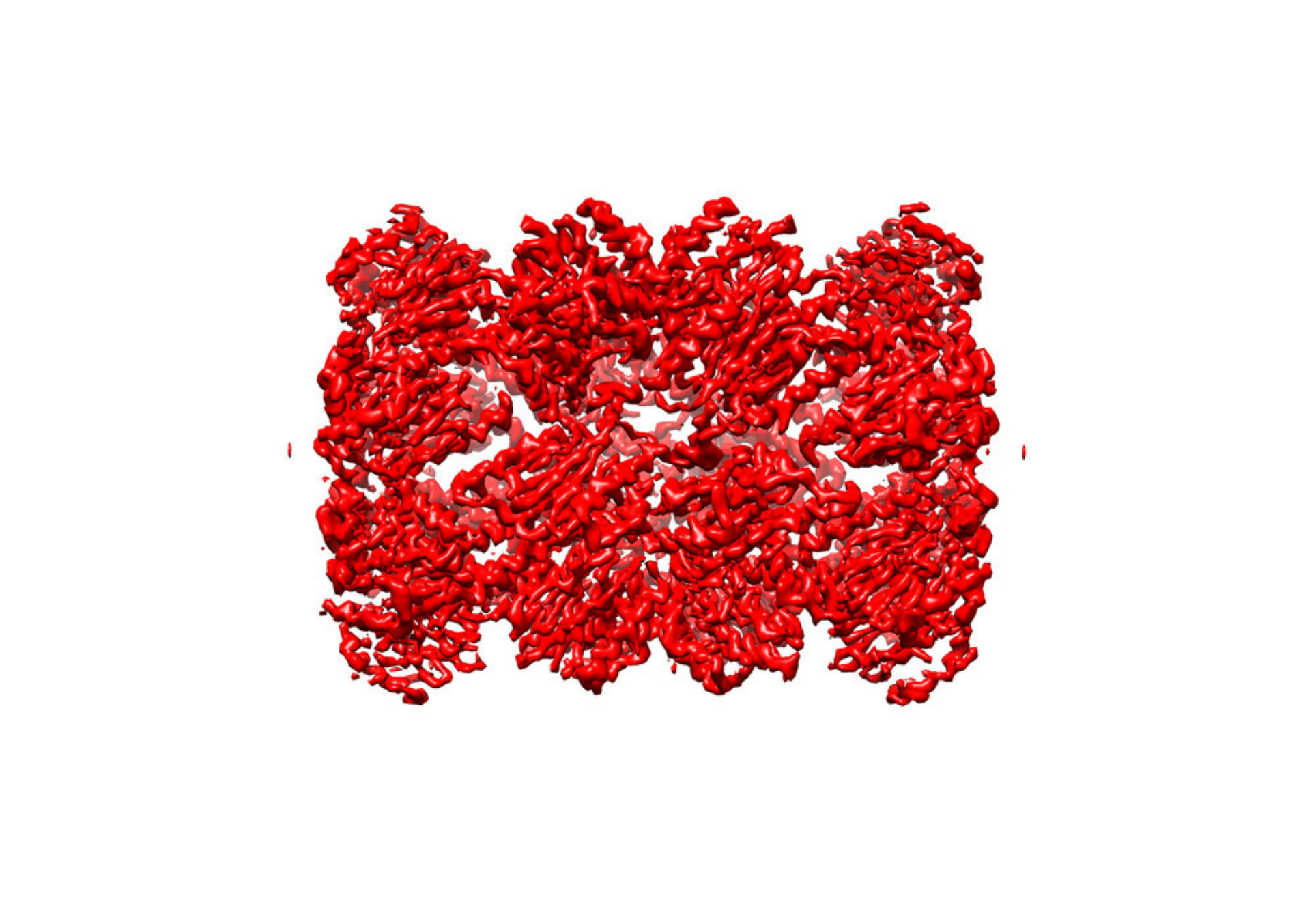}
\includegraphics[width=0.32\linewidth]{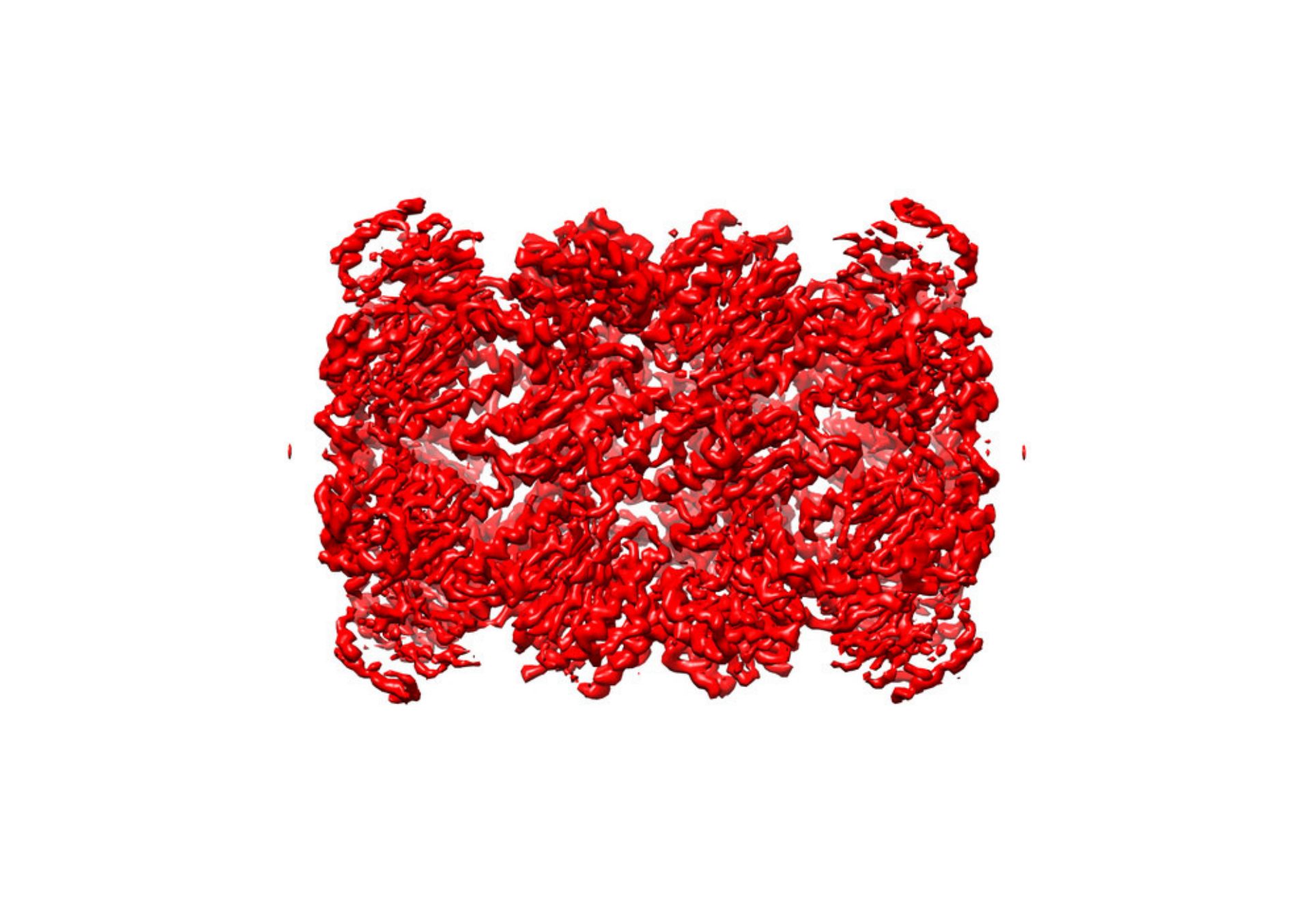}\\
\includegraphics[width=0.32\linewidth]{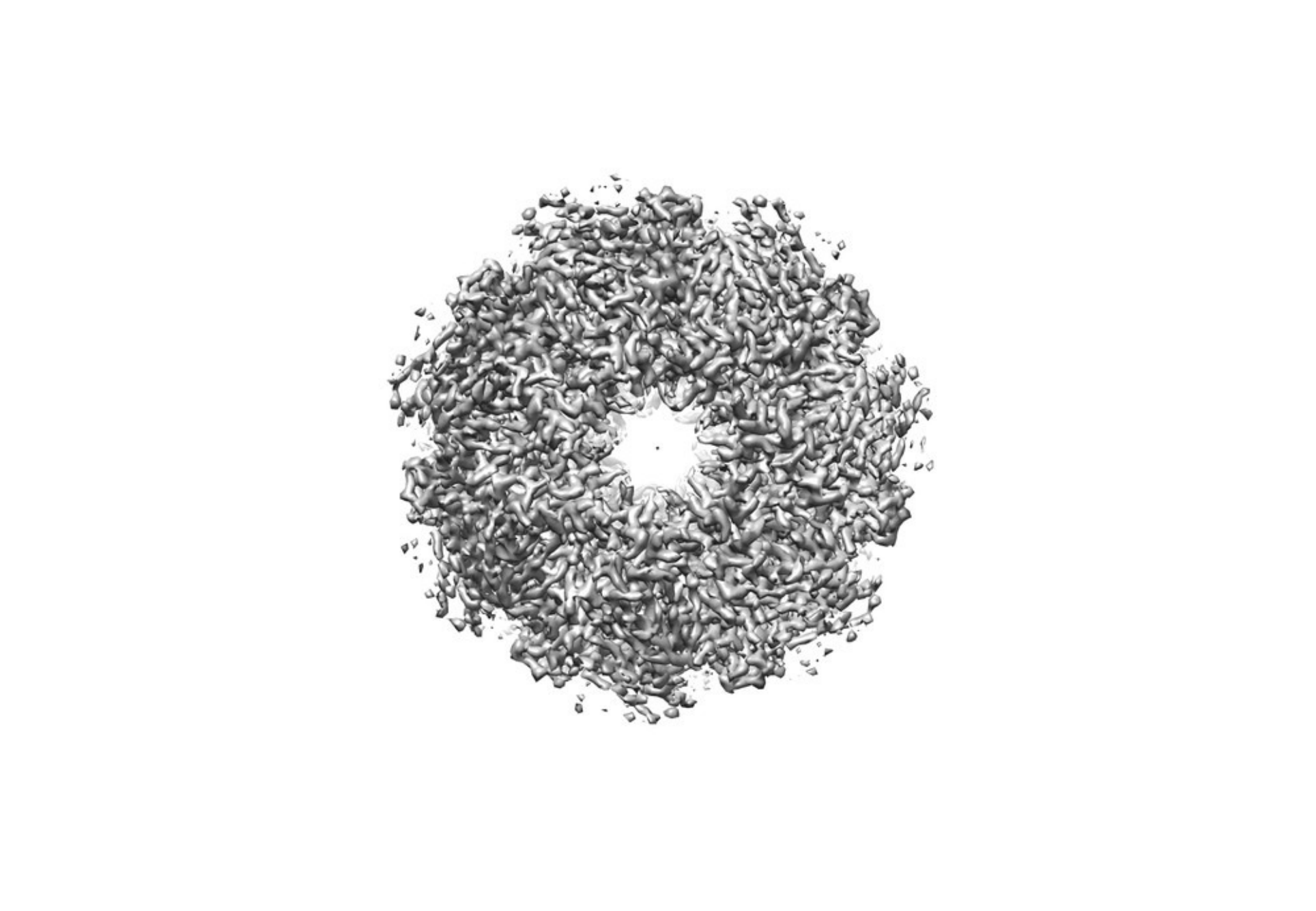}
\includegraphics[width=0.32\linewidth]{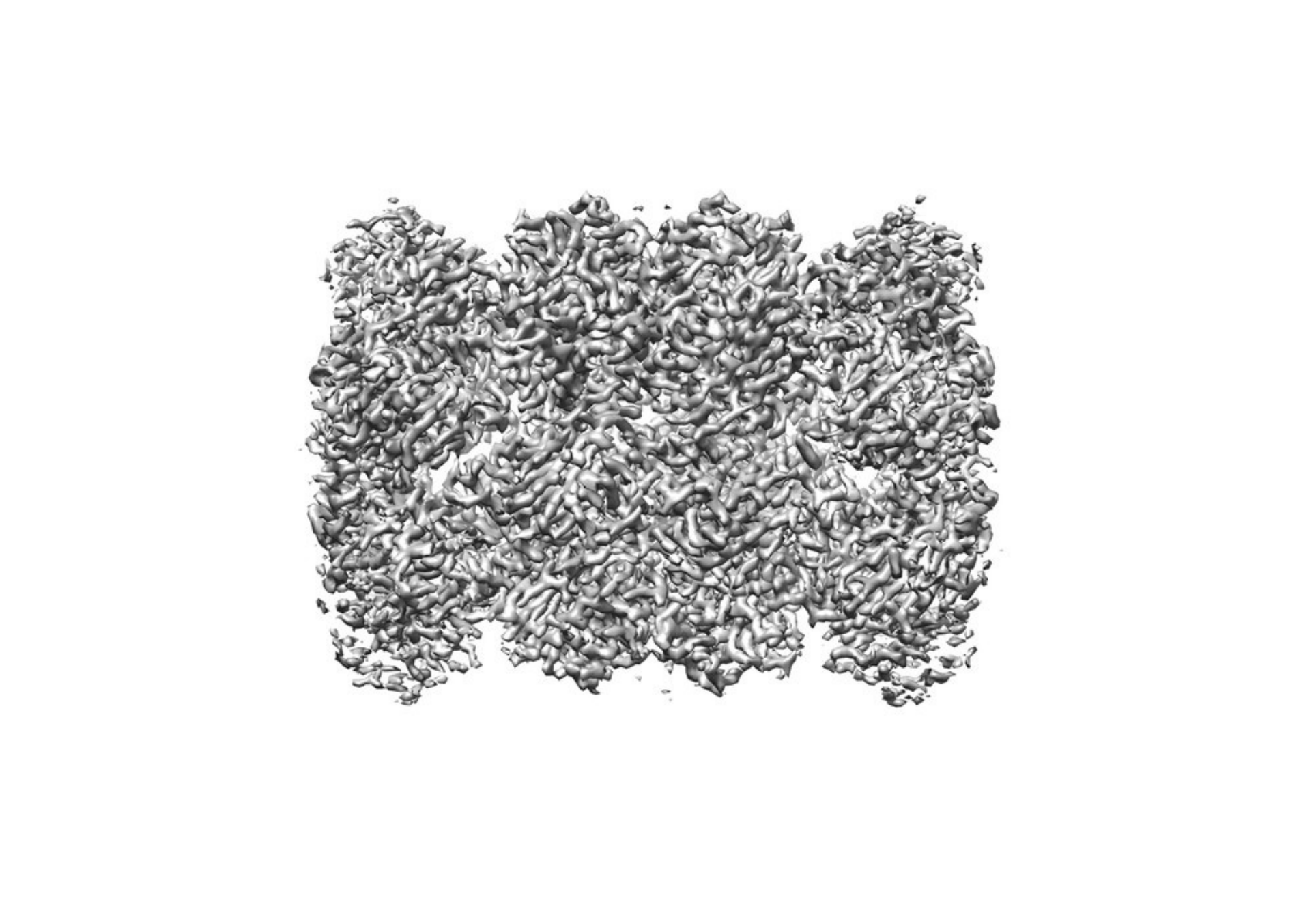}
\includegraphics[width=0.32\linewidth]{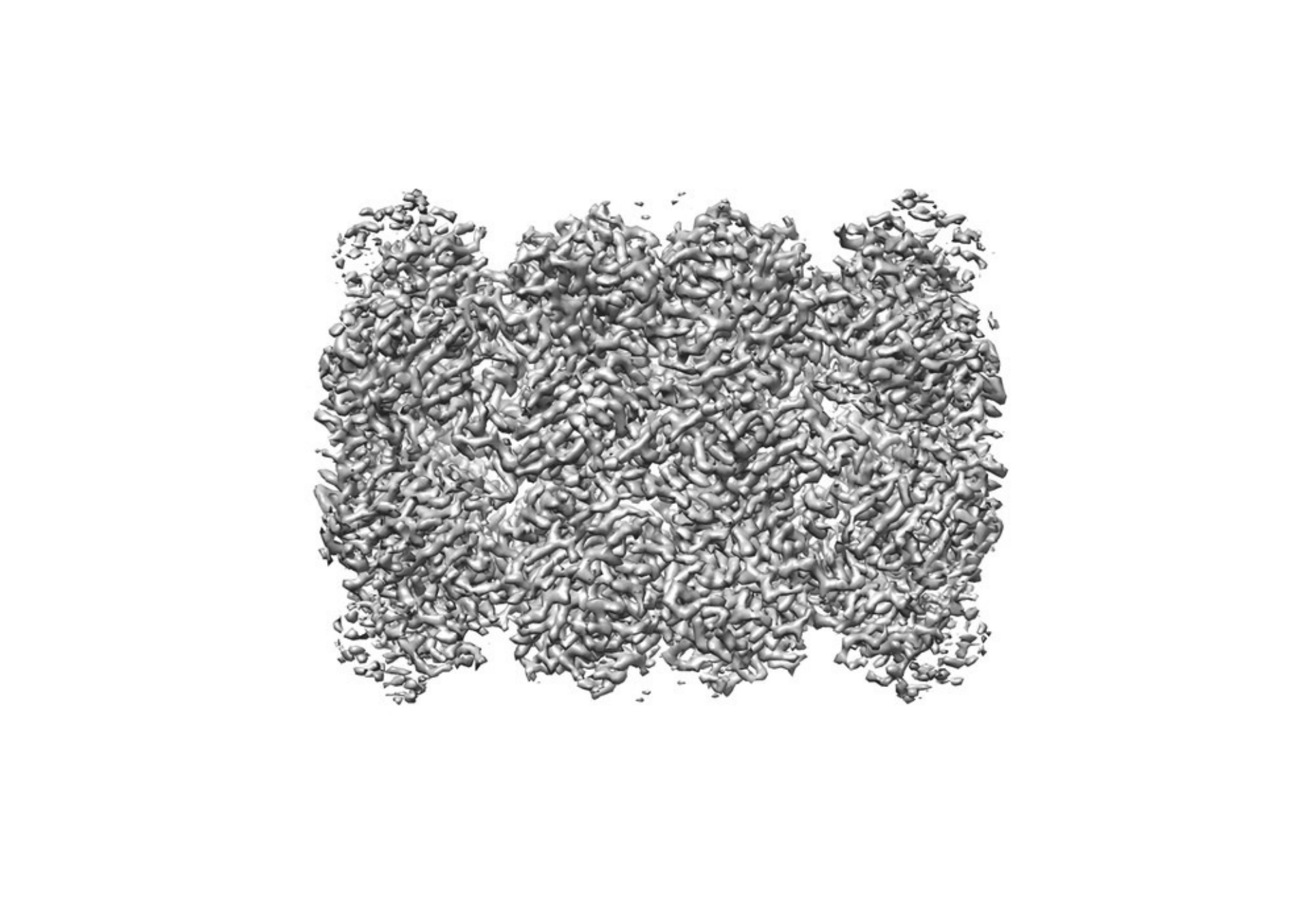}
\caption{Comparison between the APPLE picker and the particles picked in \citep{e10057}. 
On the top are views of the 3D reconstruction created in RELION from the APPLE picks and 
obtained in UCSF Chimera.
On the bottom are views of the 3D reconstruction detailed in \citep{e10057}. We use the reference volume published on EMDB (EMD-3347) and obtain 
the views in UCSF Chimera.}
\label{fig:compare2}
\end{center}
\end{figure*}

\subsection{70S ribosome}
We examine the EMPIAR-10077 \citep{n10077} dataset. The micrograph are of size $4096 \times 4096$ and contain large particles. 
Each query and reference box is  of size $40 \times 40$ pixels in the reduced micrograph. For this reason the container size we use is $500 \times 500$. 
This reduces the number of containers and thus causes the number of reference windows to be smaller.

\begin{figure*}
\begin{center}
\includegraphics[width=0.31\linewidth]{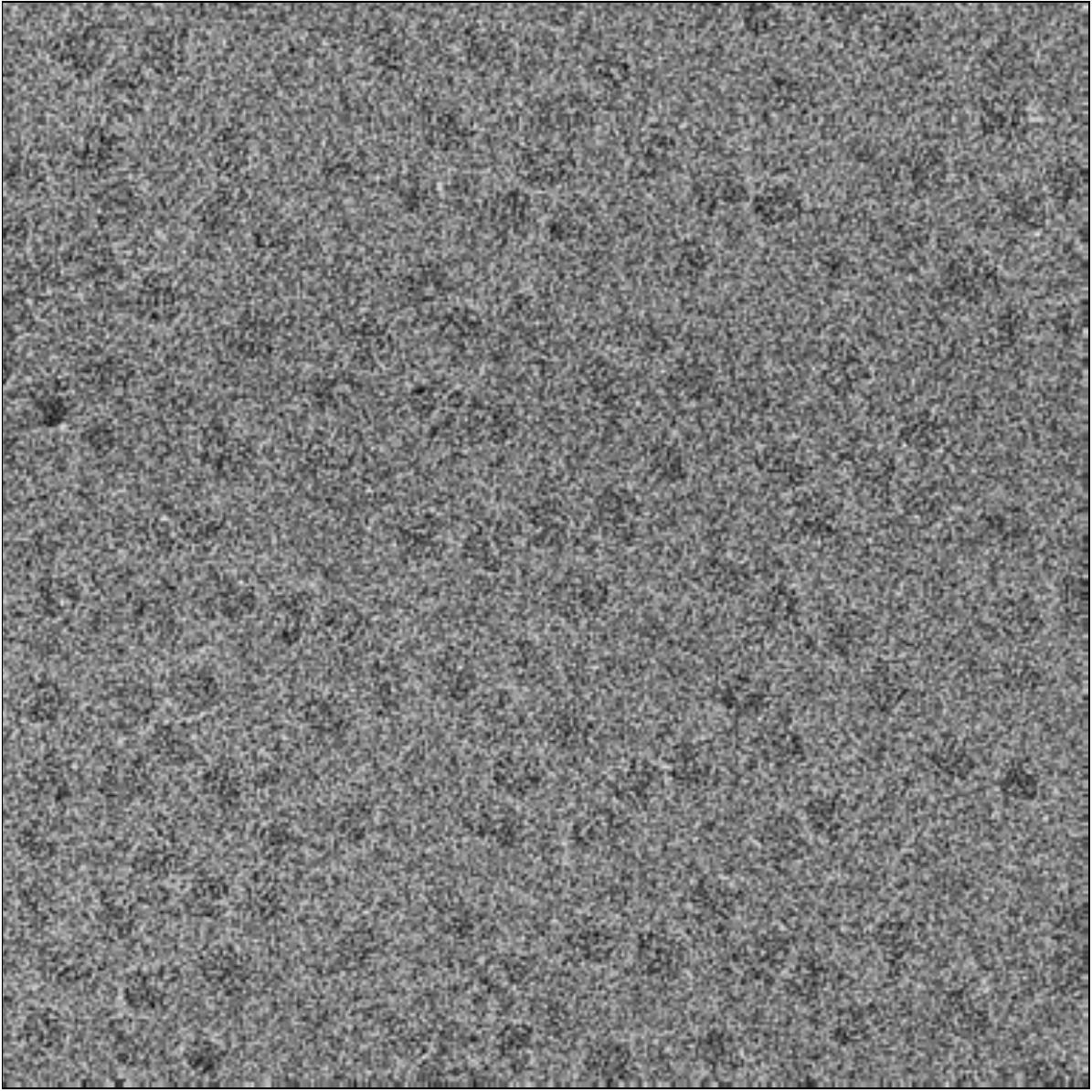}
\includegraphics[width=0.31\linewidth]{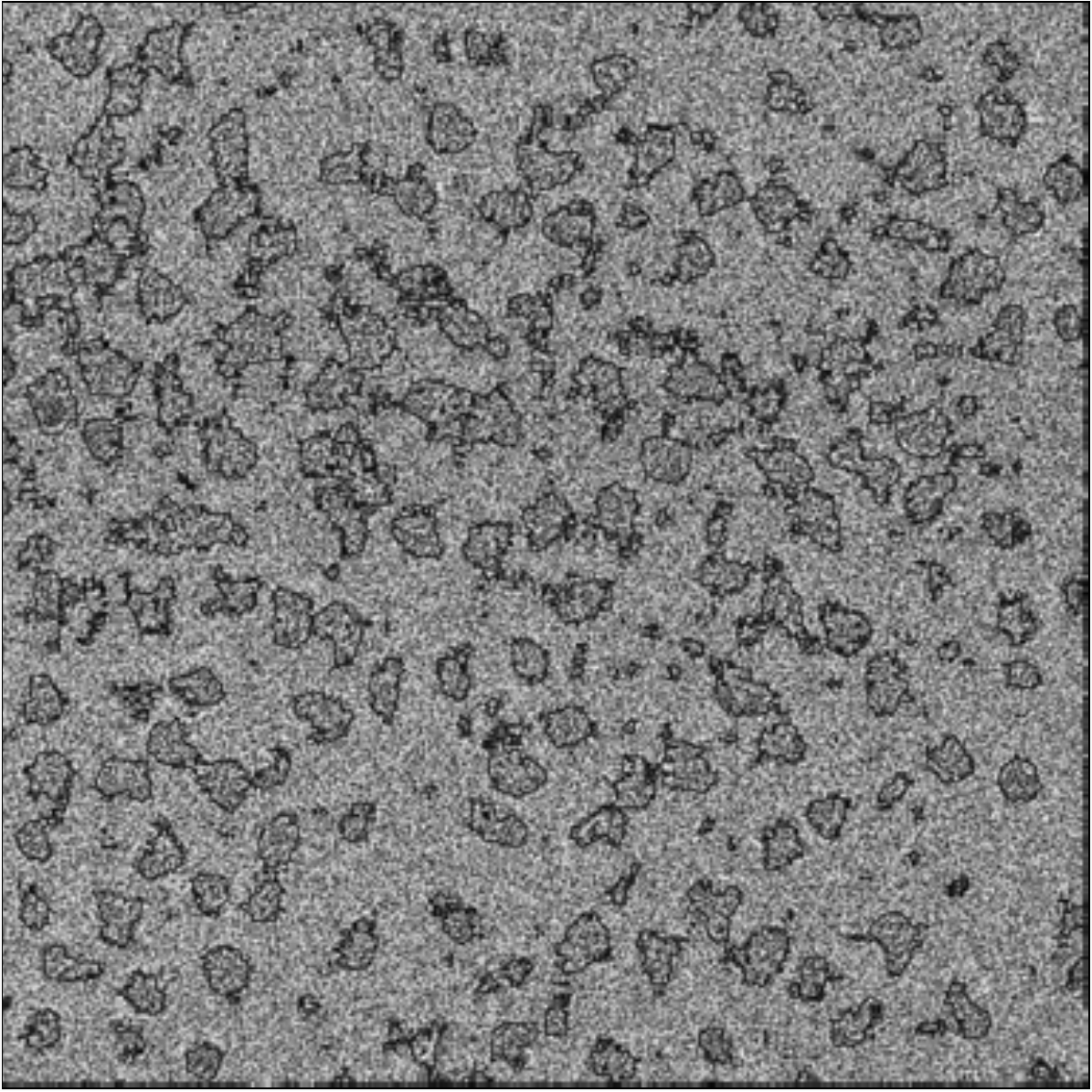}
\includegraphics[width=0.31\linewidth]{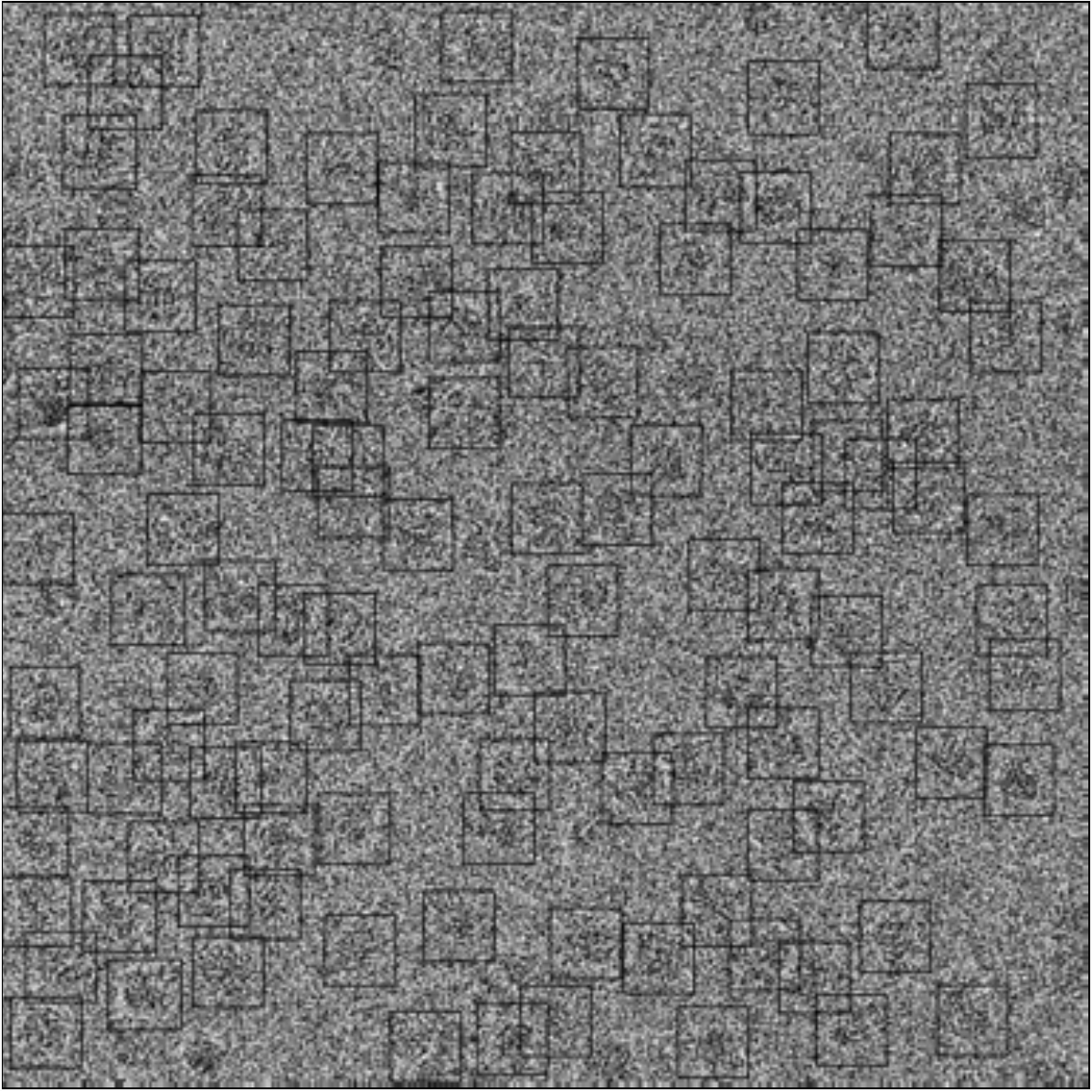}
\includegraphics[width=0.31\linewidth]{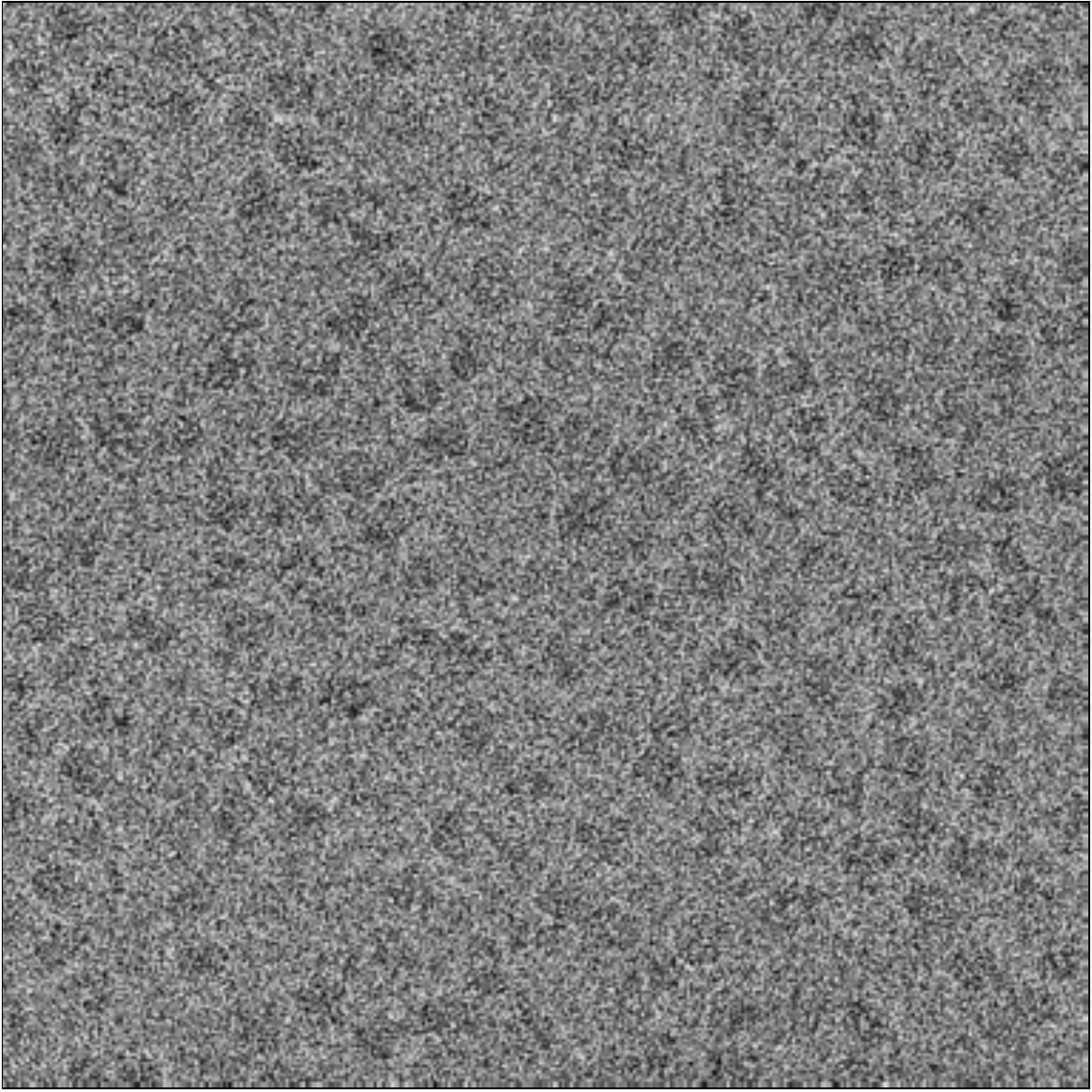}
\includegraphics[width=0.31\linewidth]{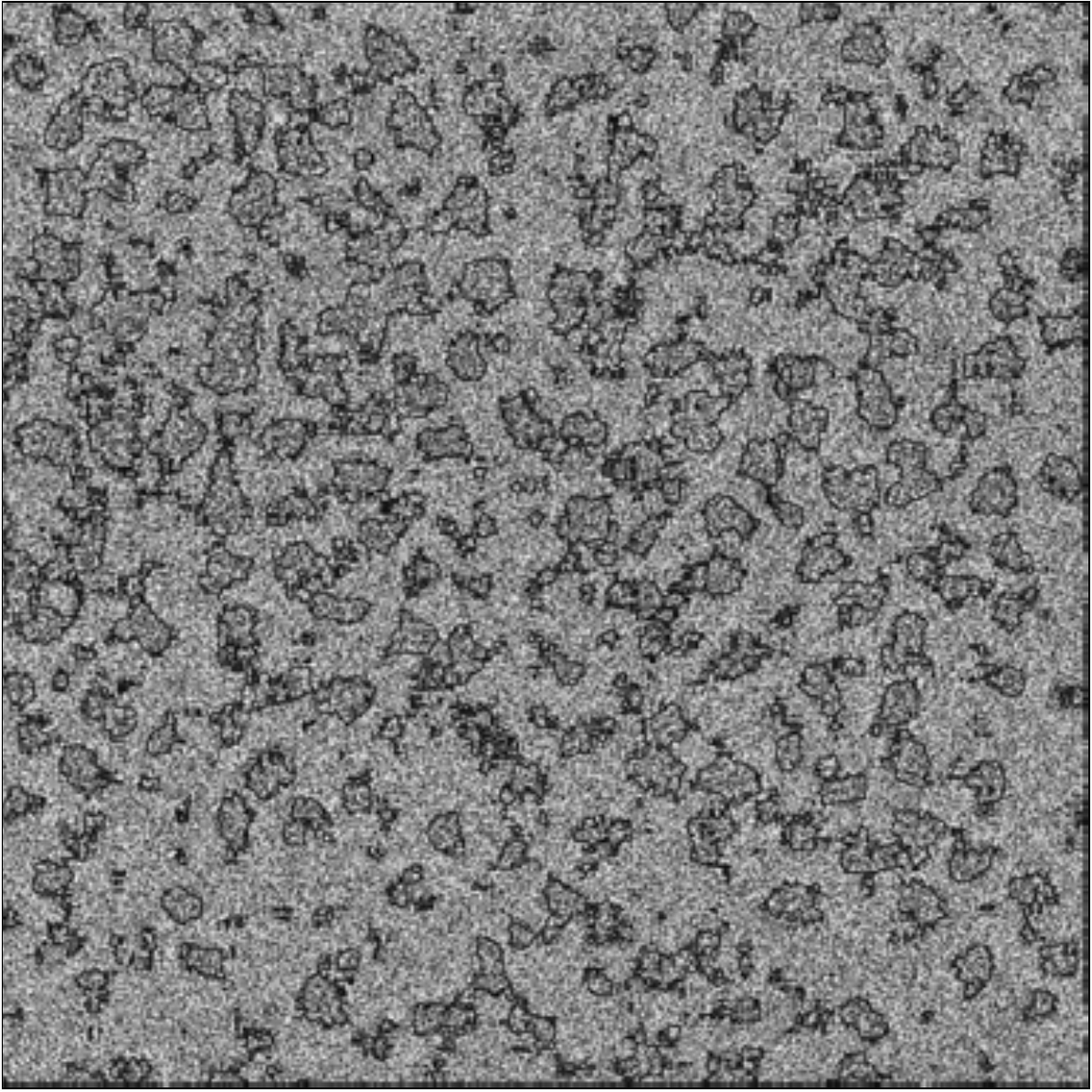}
\includegraphics[width=0.31\linewidth]{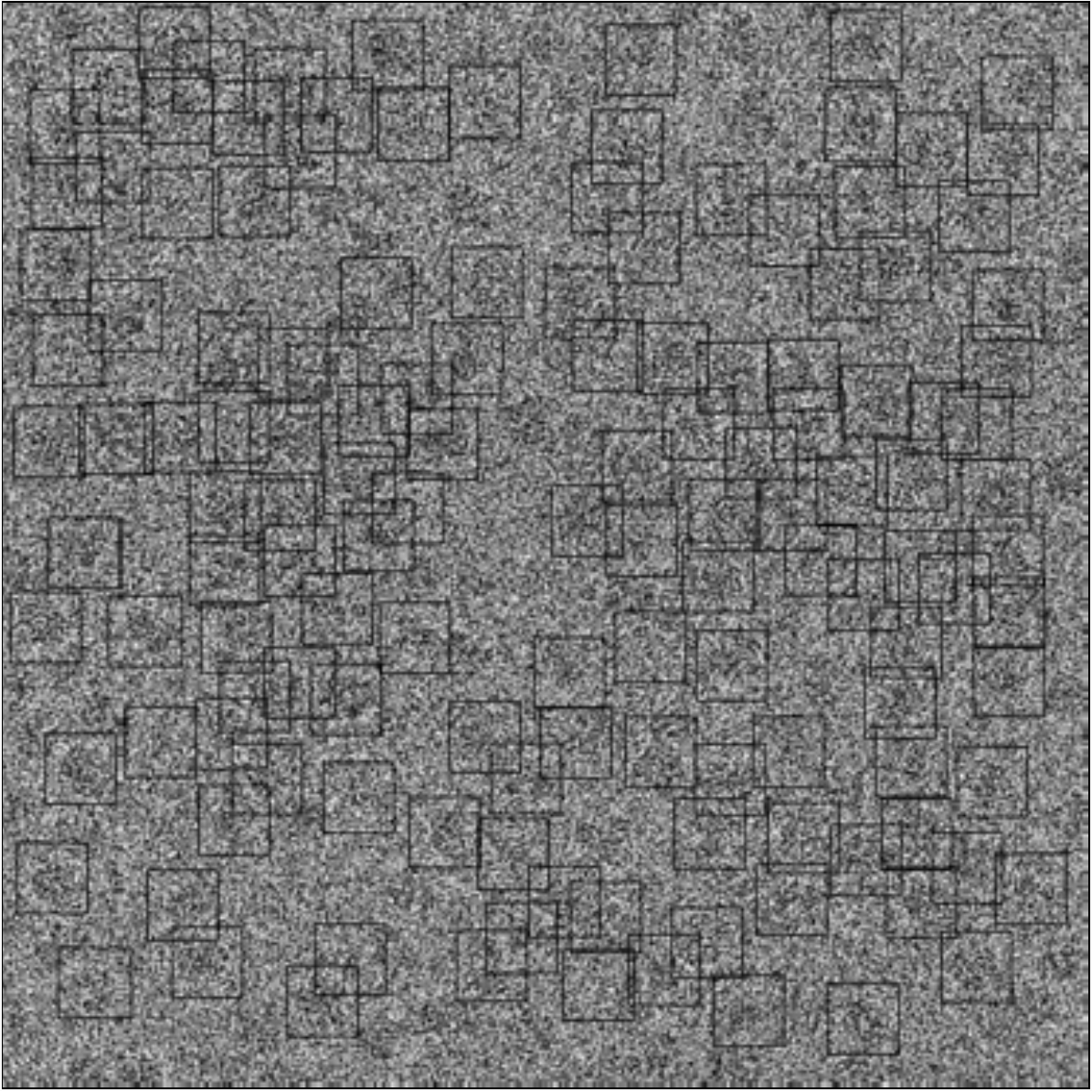}
\includegraphics[width=0.31\linewidth]{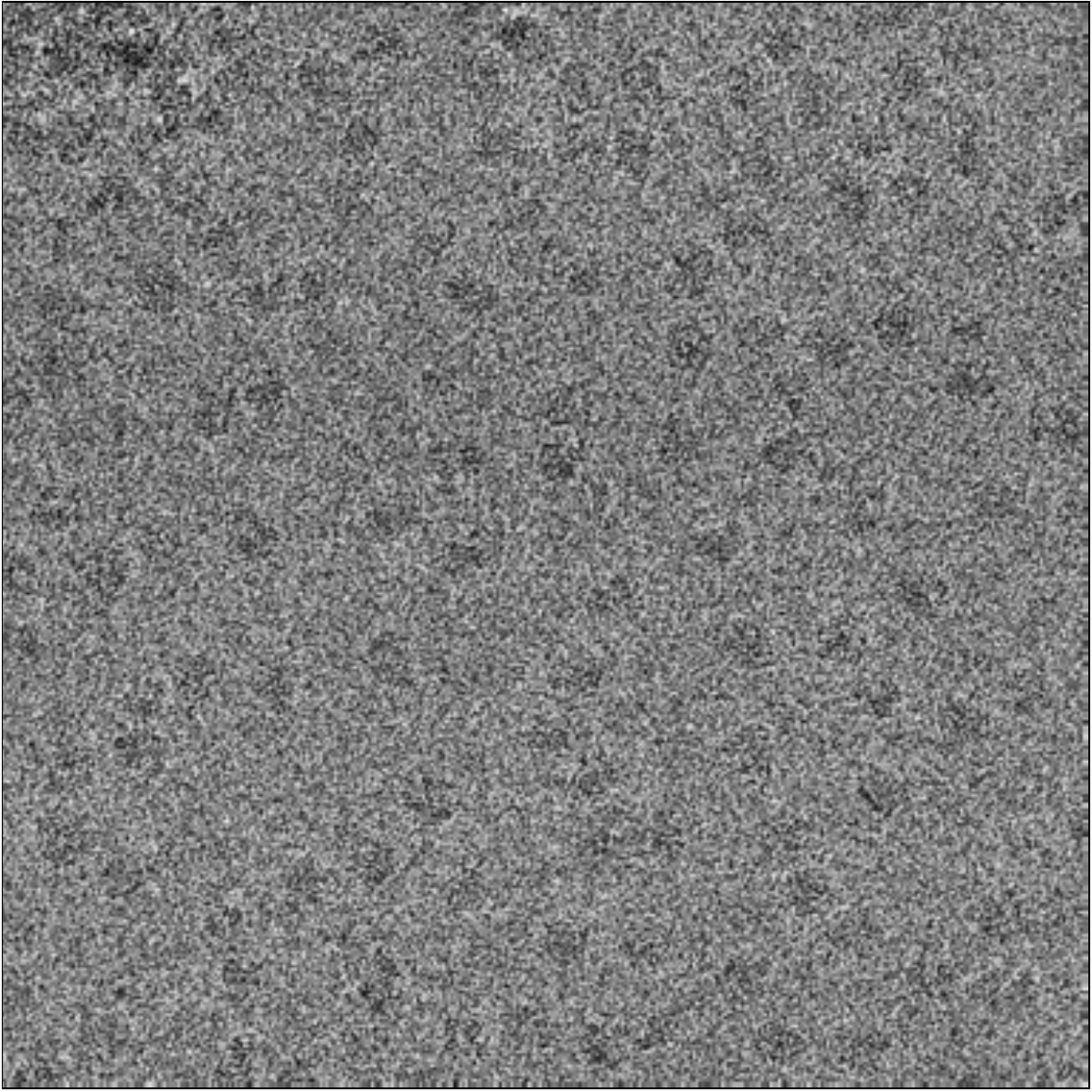}
\includegraphics[width=0.31\linewidth]{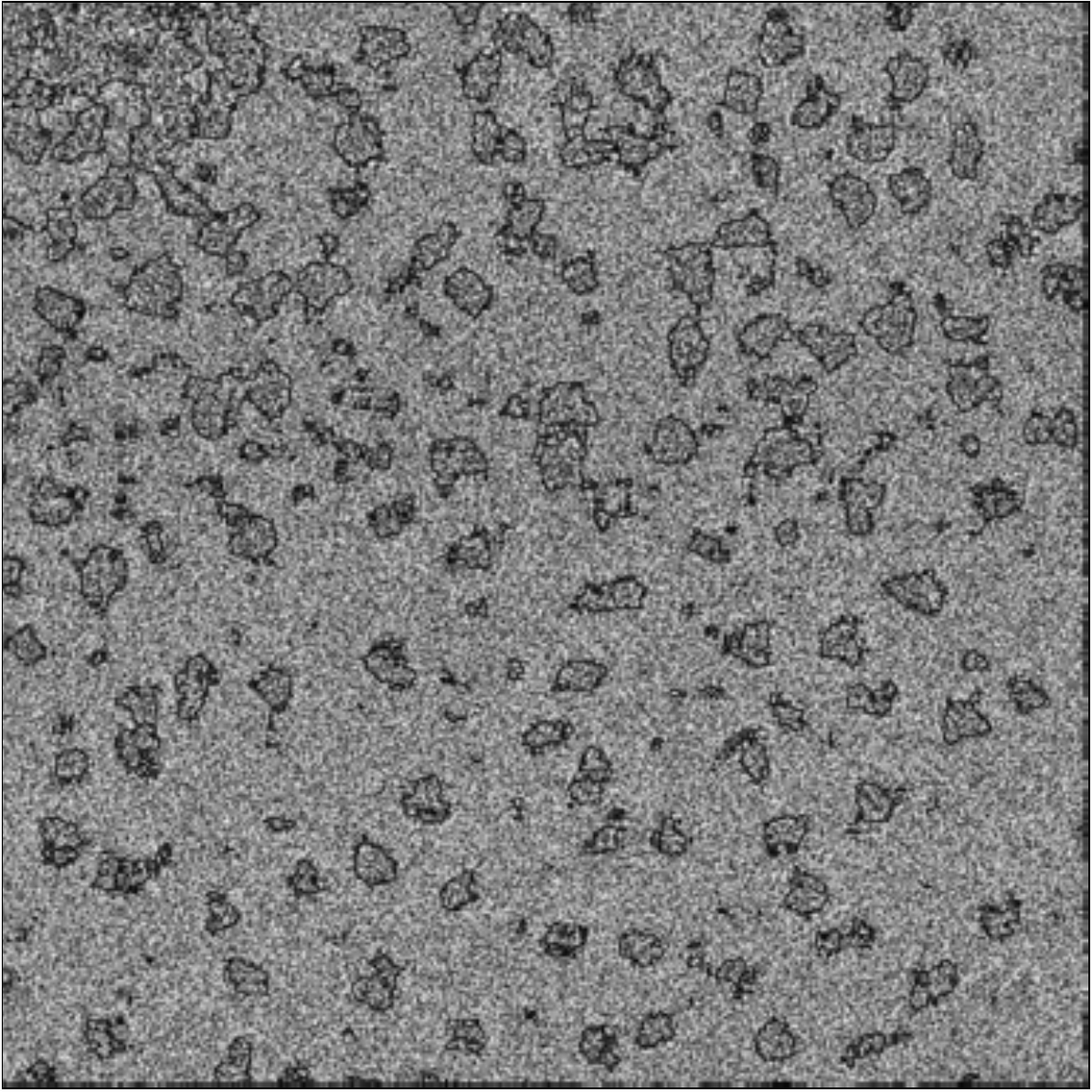}
\includegraphics[width=0.31\linewidth]{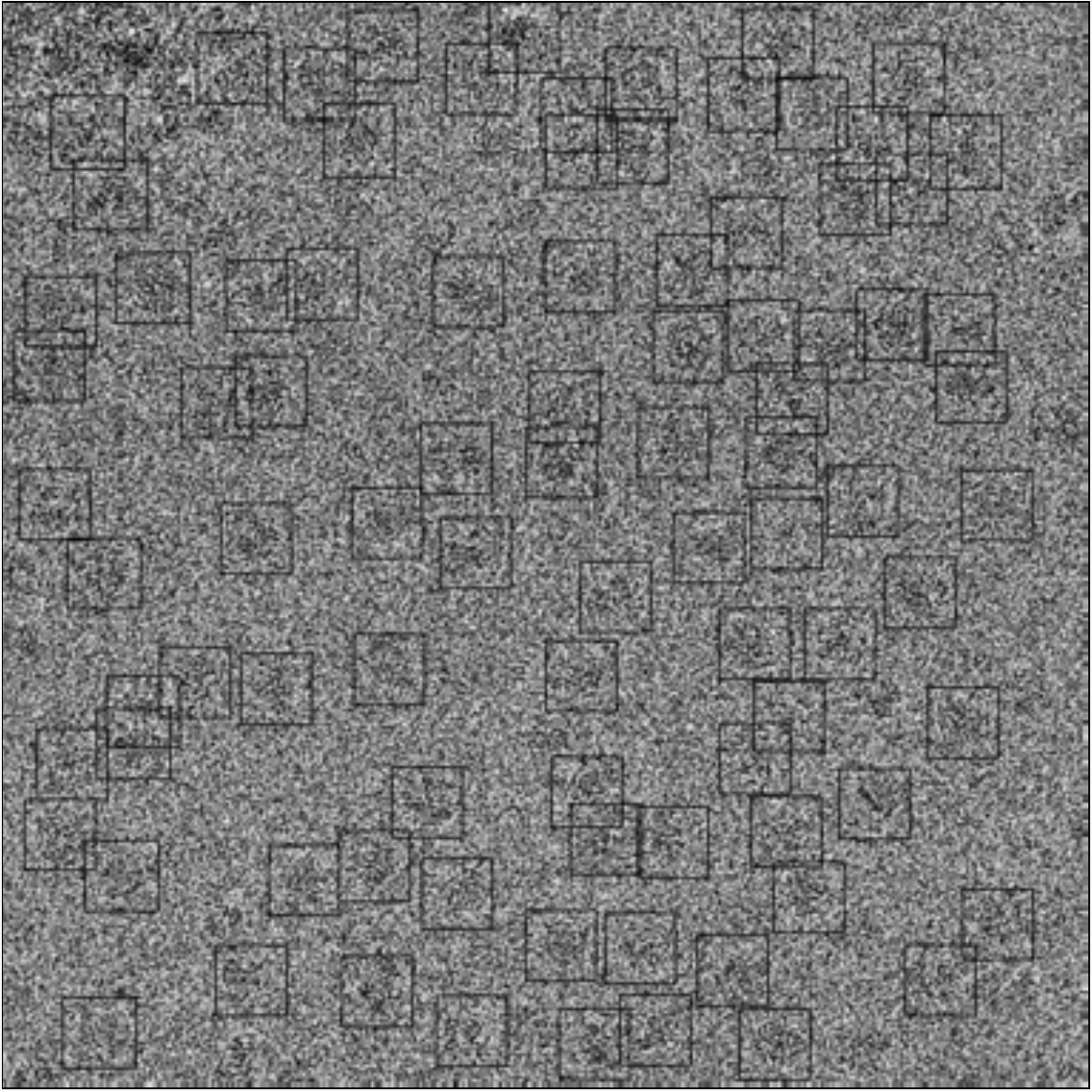}
\caption{Picked particles of sample 70S ribosome micrographs. The micrographs are presented in the 
left column. Classification results are presented in the center. The picked particles are on the right.}
\label{fig:10077}
\end{center}
\end{figure*}

For classifier training we suggest to use $\tau_1 =  7\%$  and $\tau_2 = 7\%$  to determine the training set  {(see Section 4.2 for a discussion about the choice of parameters.)}
We set the bandwidth of the kernel function for the SVM classifier and its slack parameter both to $1$. 
Examples of results for the APPLE picker are presented in Figure 
\ref{fig:10077}.  Run time is approximately 2 minutes per micrograph on the CPU, or approximately $14$ seconds per micrograph on the GPU.

\subsection{KLH}
\label{sec:klh}

The micrographs in the KLH dataset  \citep{bakeoff, KLH} are of size $2048 \times 2048$. 
To lower runtime we once again perform binning.  
Following this reduction in size, we use query and reference images of  size $30 \times 30$ and  containers of size $115 \times 115$. 
The training set for the SVM classifier is determined using the thresholds $\tau_1 =  16\%$ and $\tau_2 = 70\%$.
We use the same configuration of the classifier (bandwidth and slack parameter) as in the previous experiments.

\begin{figure*}
\centering
{\includegraphics[width=0.3\linewidth]{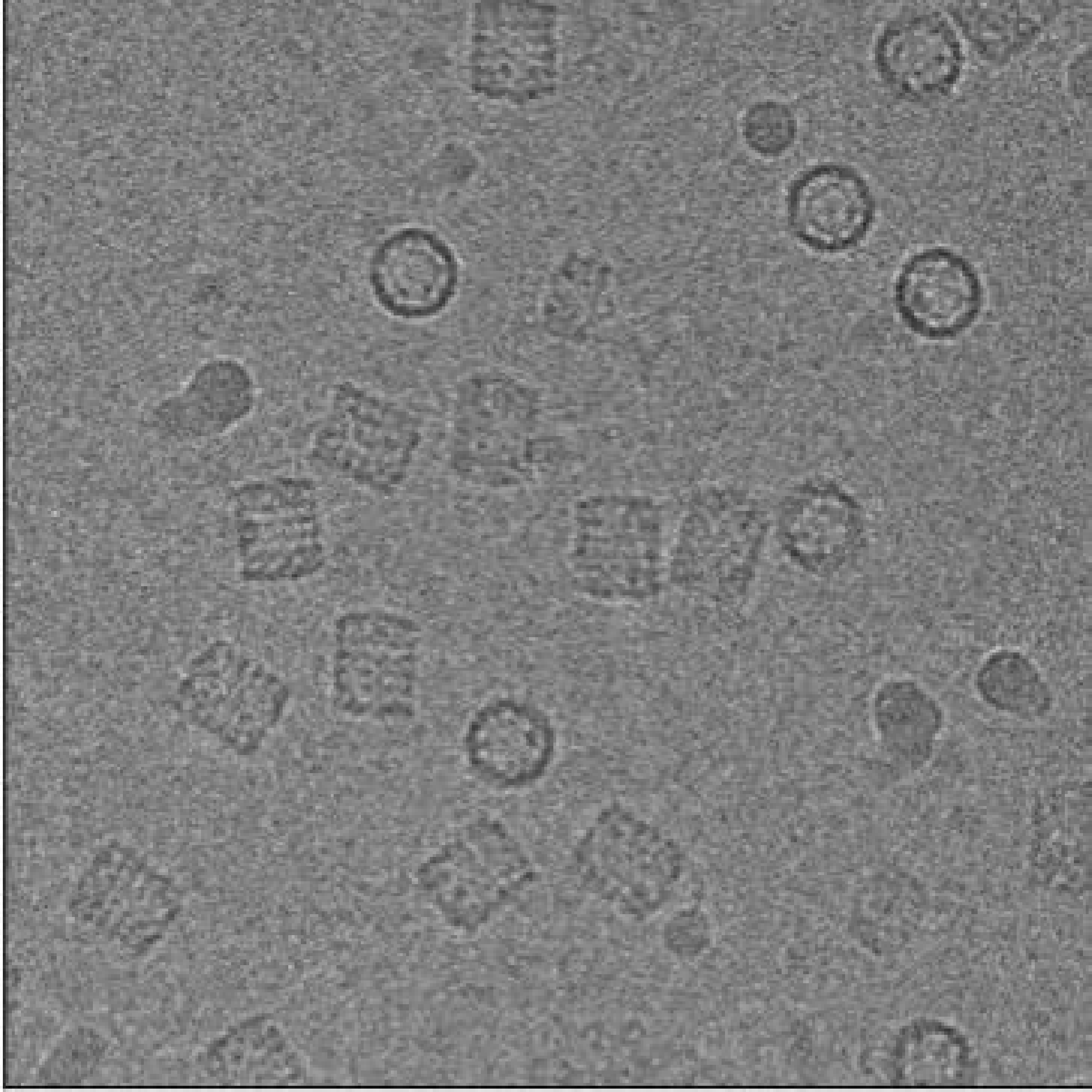}}
{\includegraphics[width=0.3\linewidth]{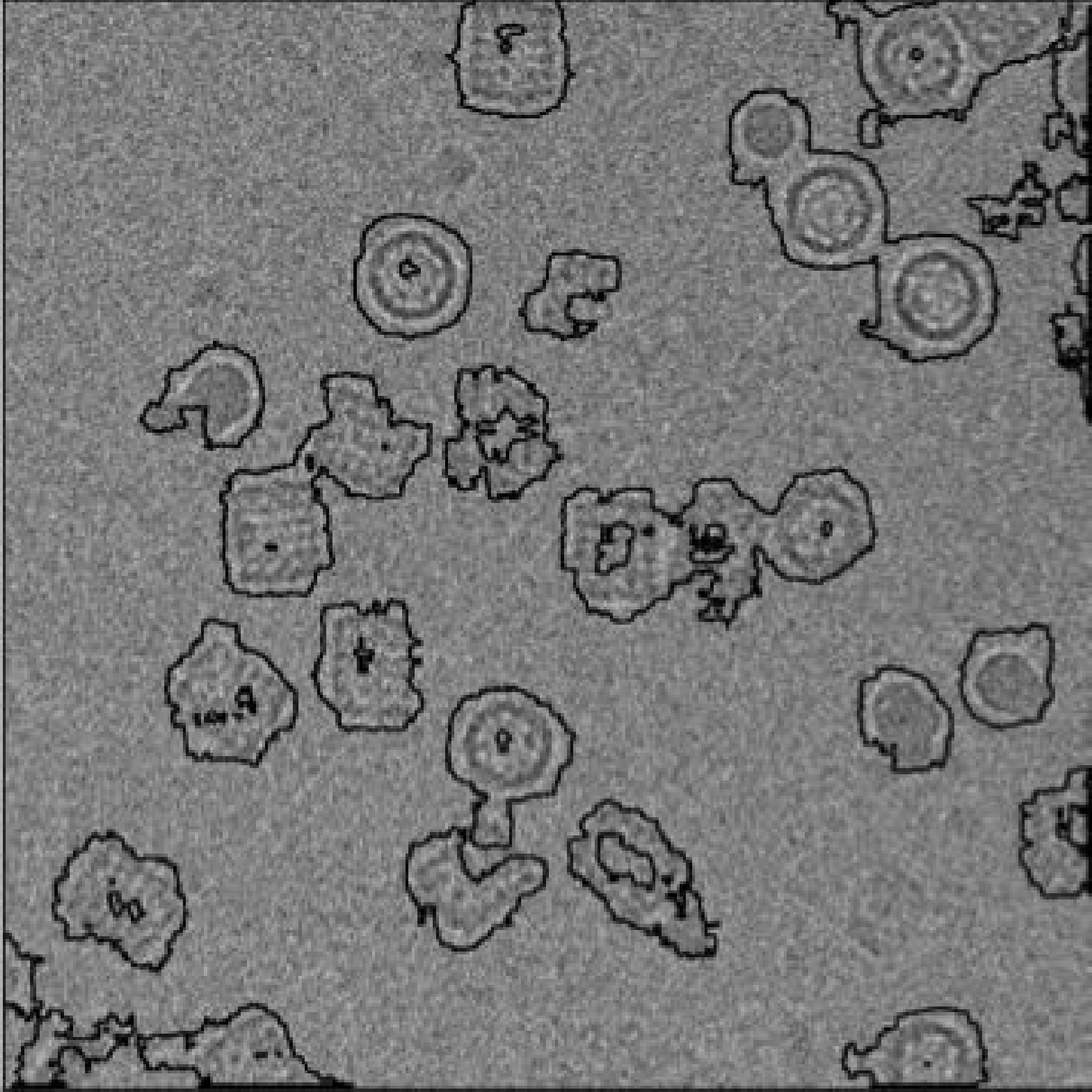}}
{\includegraphics[width=0.3\linewidth]{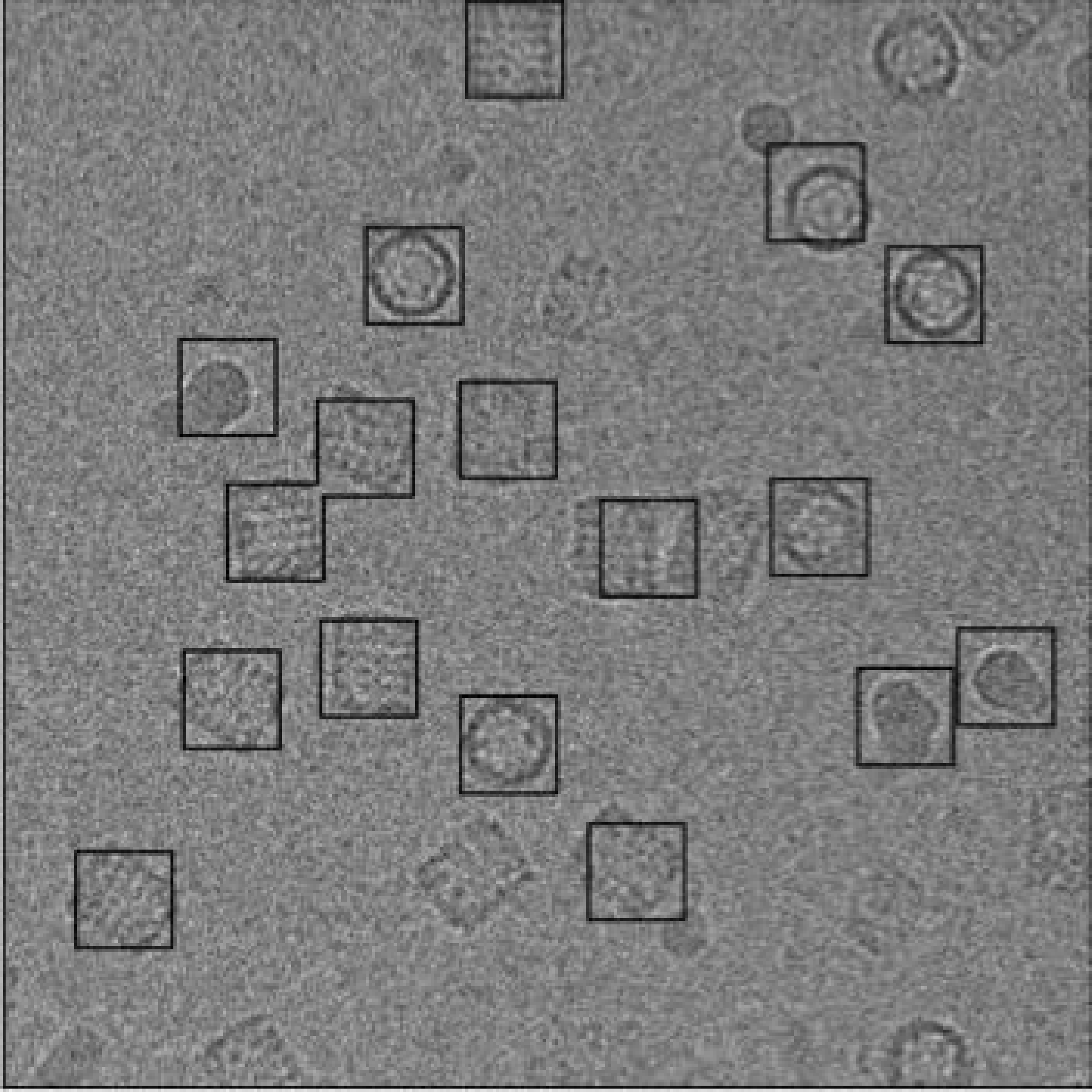}}
{\includegraphics[width=0.3\linewidth]{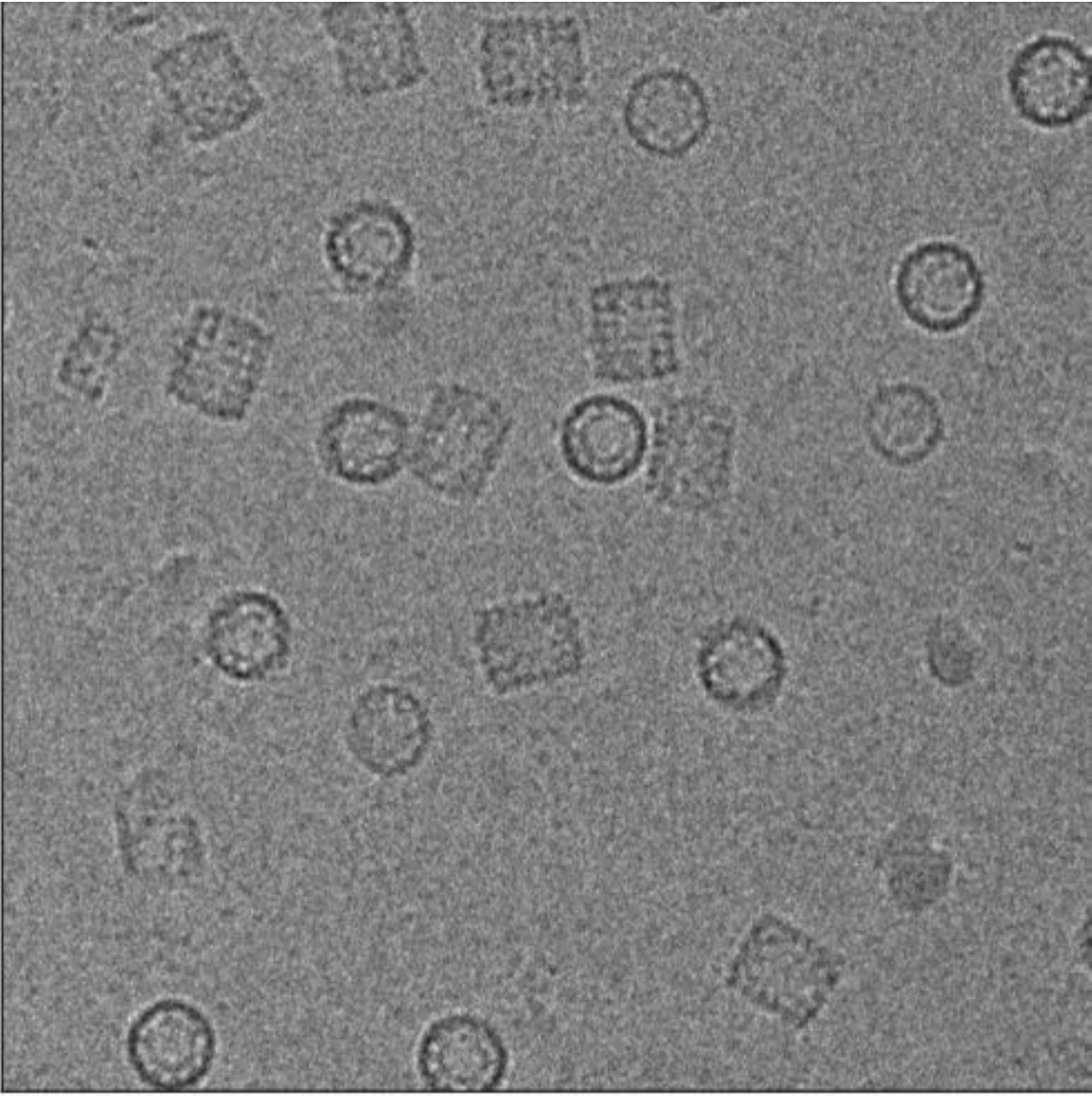}}
{\includegraphics[width=0.3\linewidth]{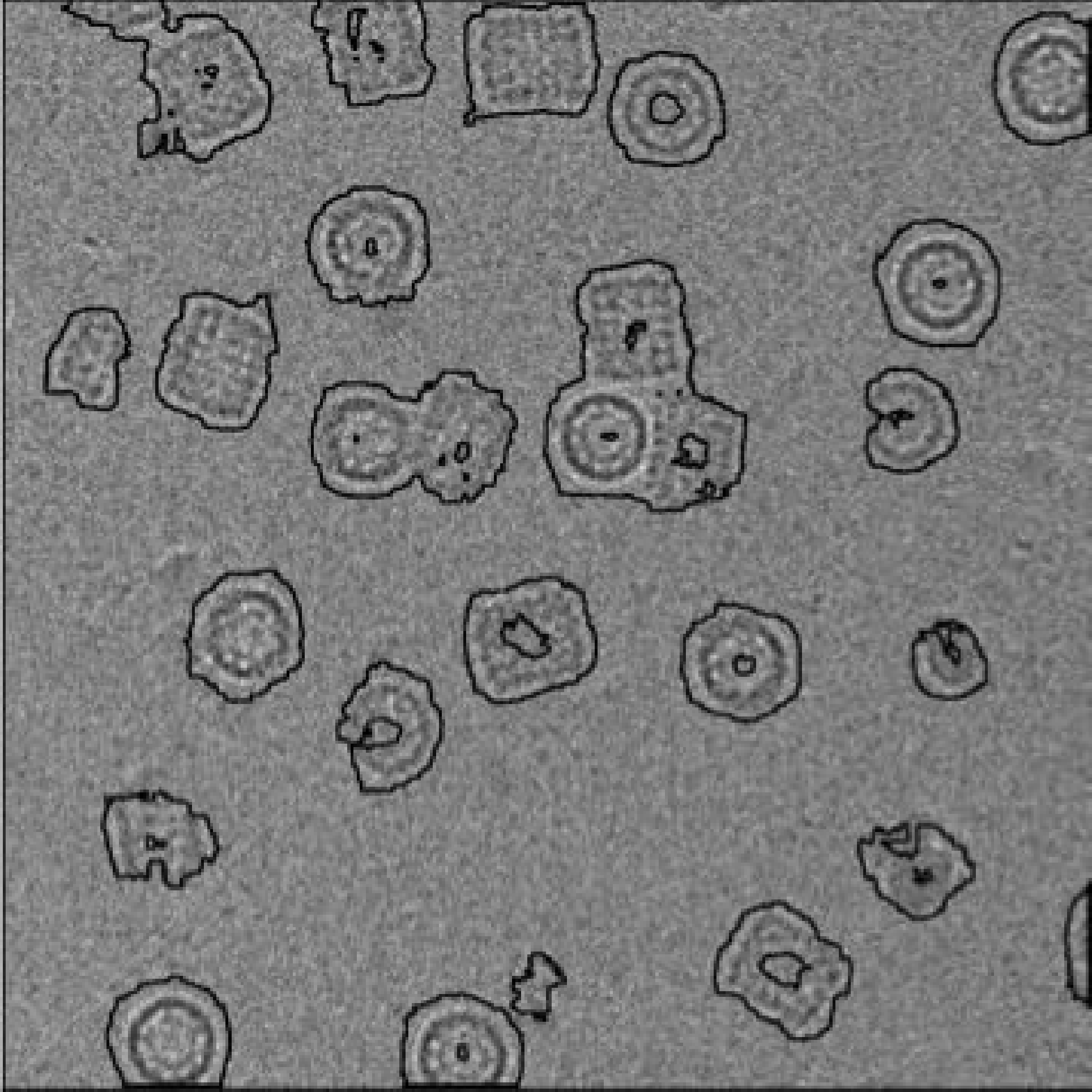}}
{\includegraphics[width=0.3\linewidth]{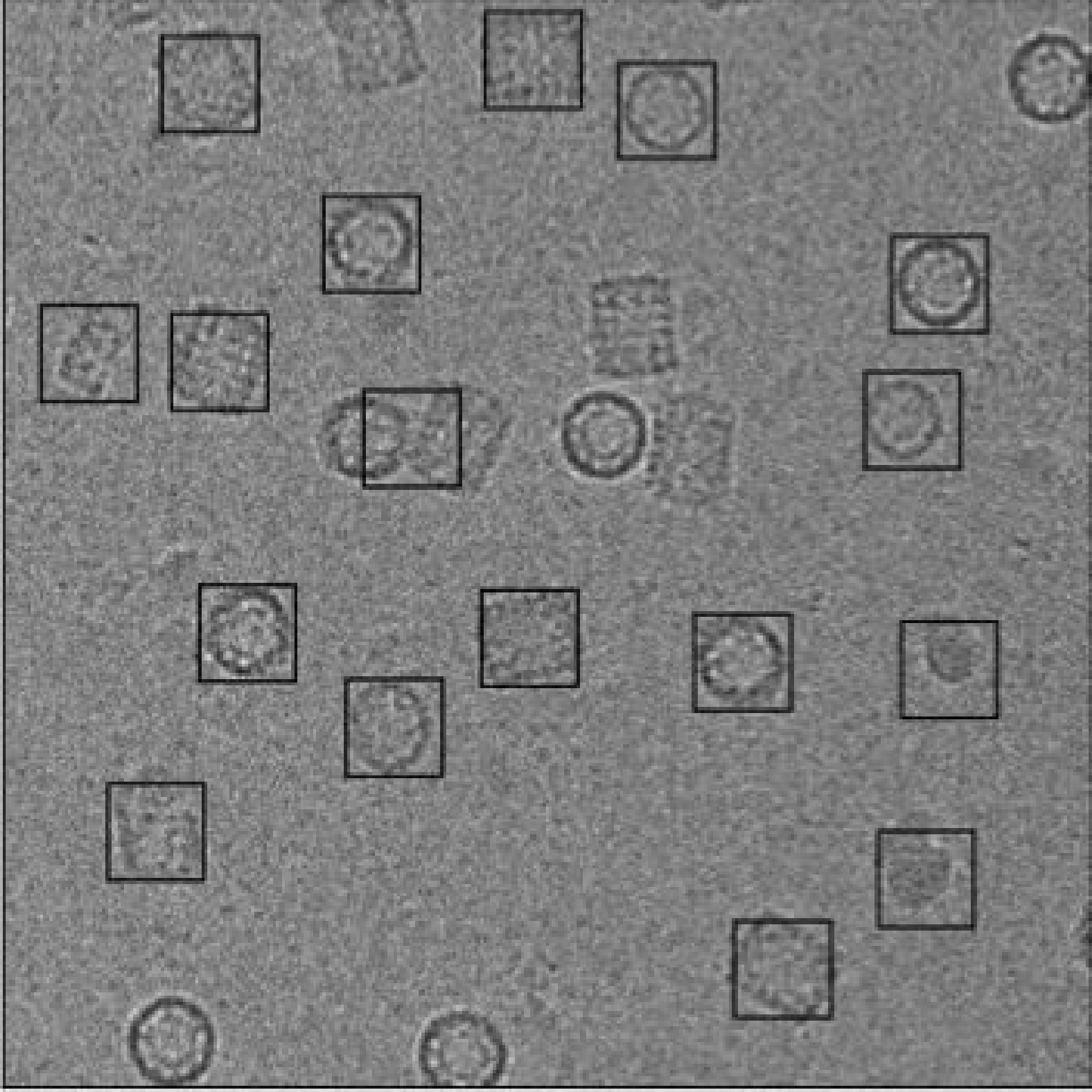}}
{\includegraphics[width=0.3\linewidth]{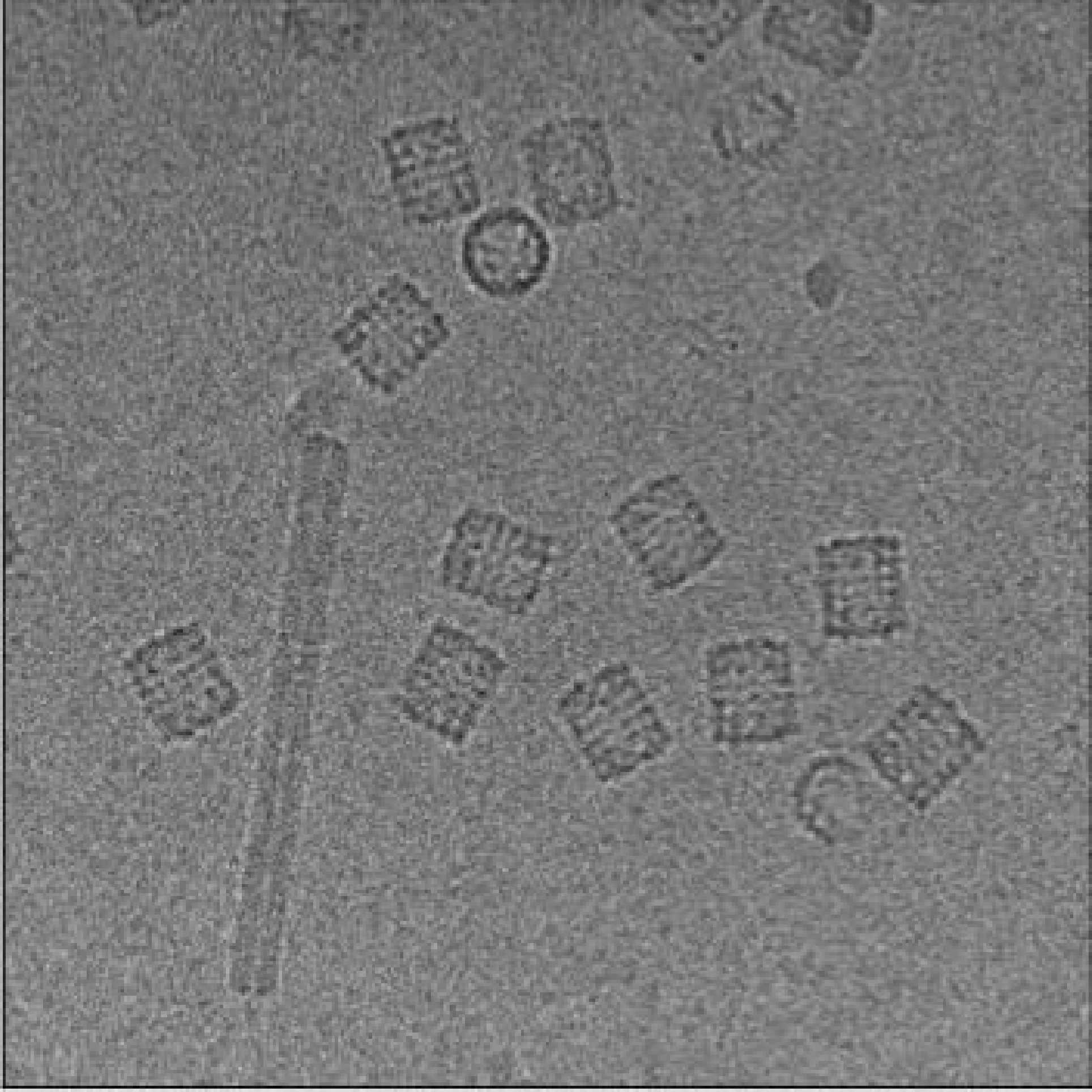}}
{\includegraphics[width=0.3\linewidth]{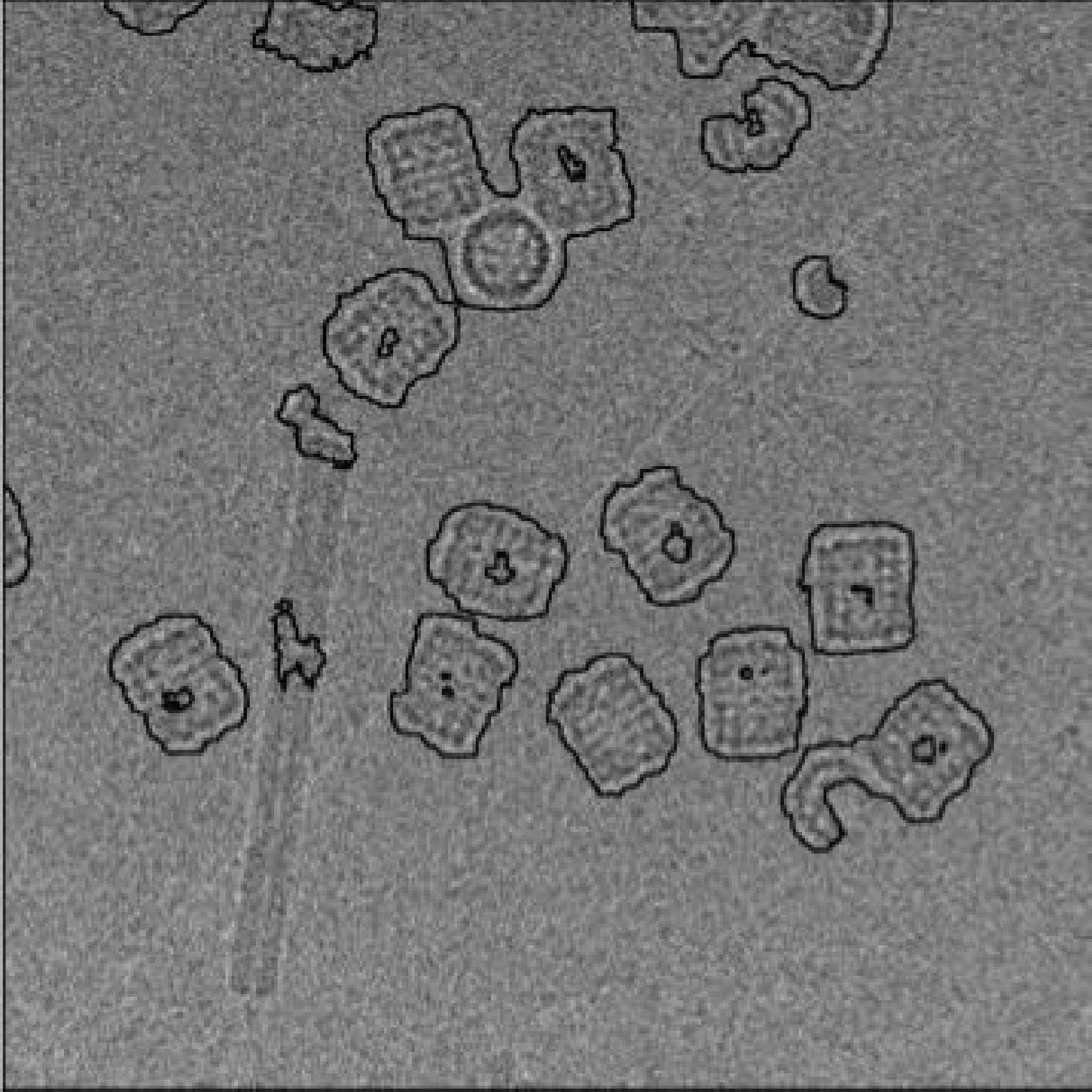}}
{\includegraphics[width=0.3\linewidth]{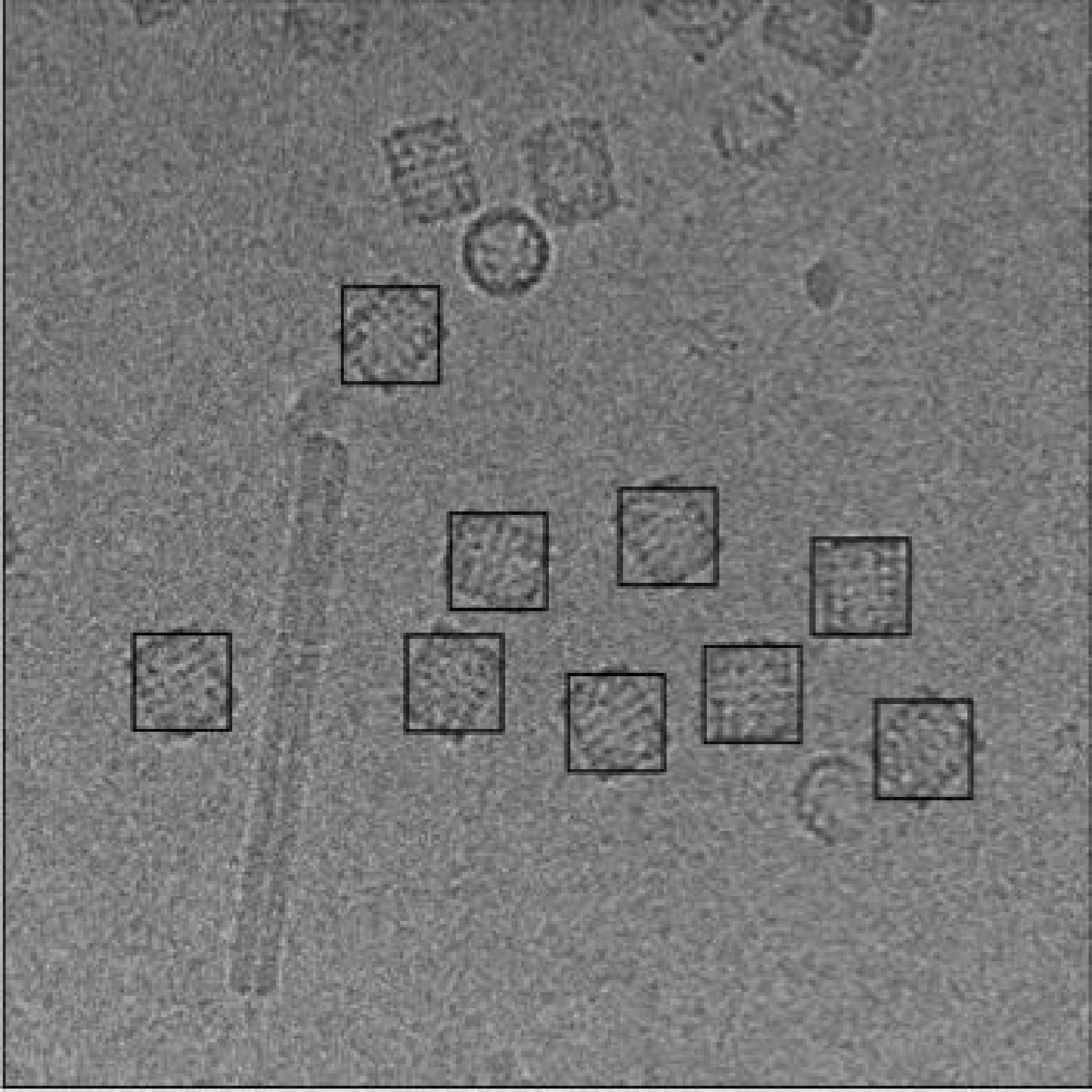}}
\caption{Result on KLH dataset. The left column contains micrographs. The middle column contains the windows classified as
particle by our classifier. The right column contains the picked particles.}
 \label{fig:klh1}
\end{figure*}

\begin{figure*}
\centering
{\includegraphics[width=0.3\linewidth]{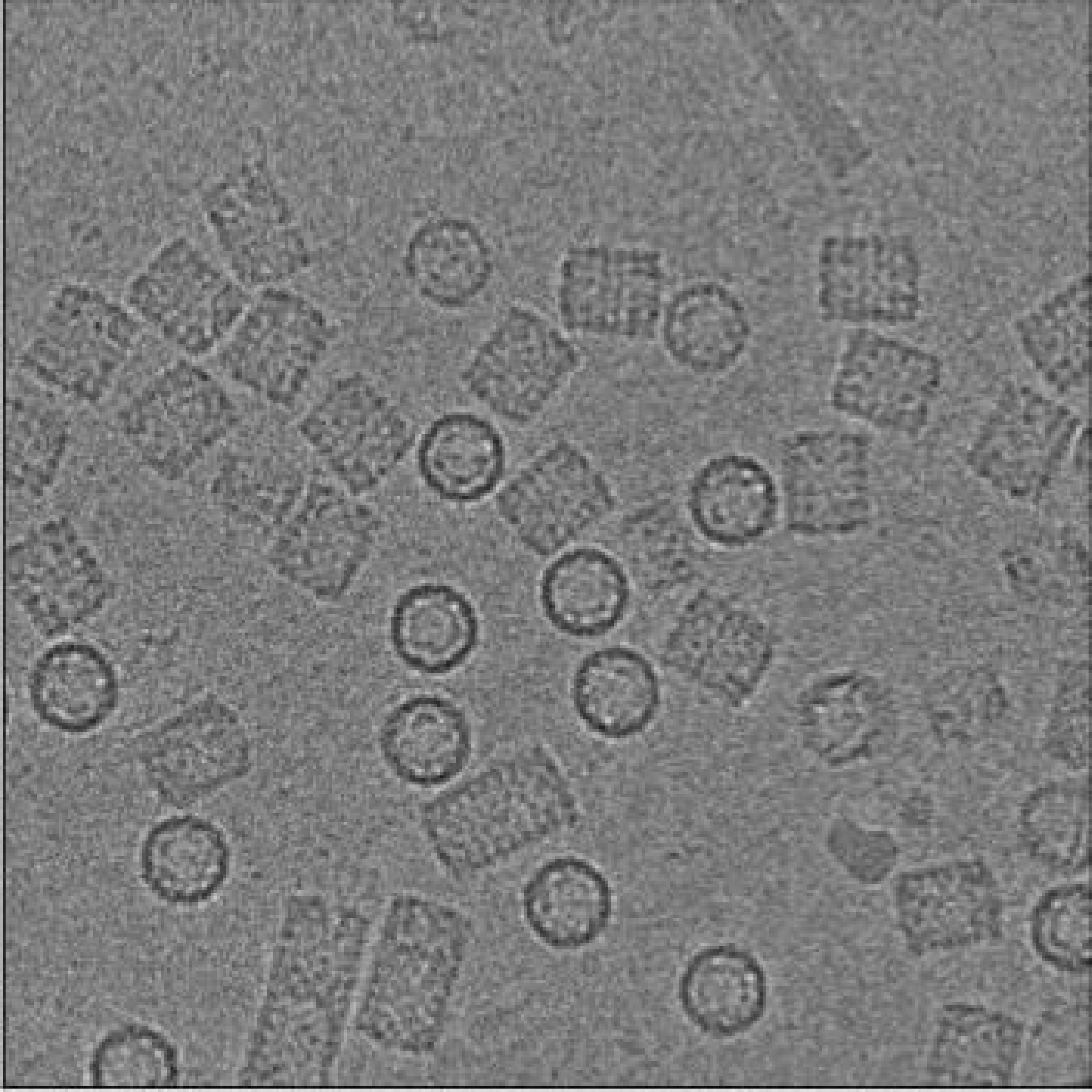}}
{\includegraphics[width=0.3\linewidth]{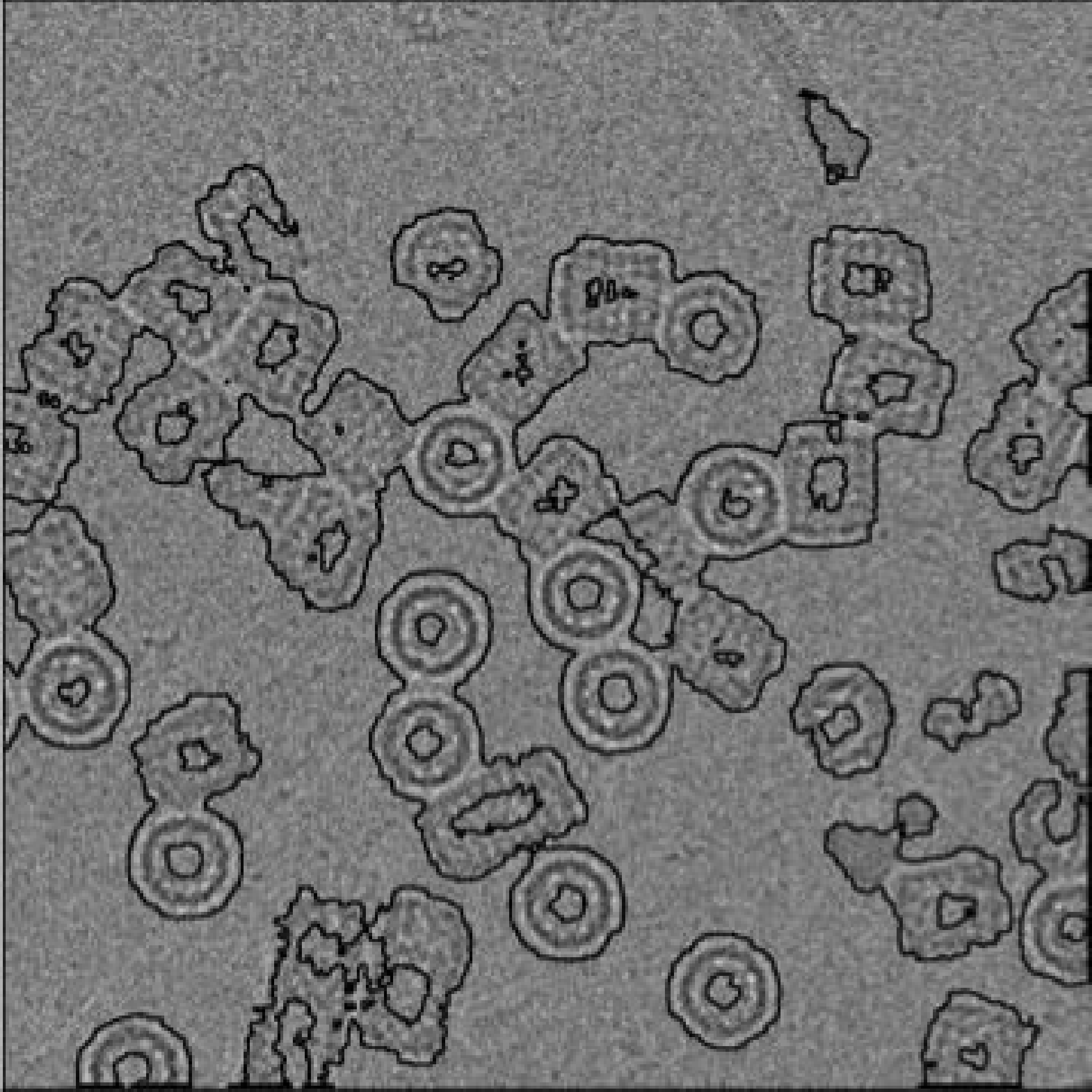}}
{\includegraphics[width=0.3\linewidth]{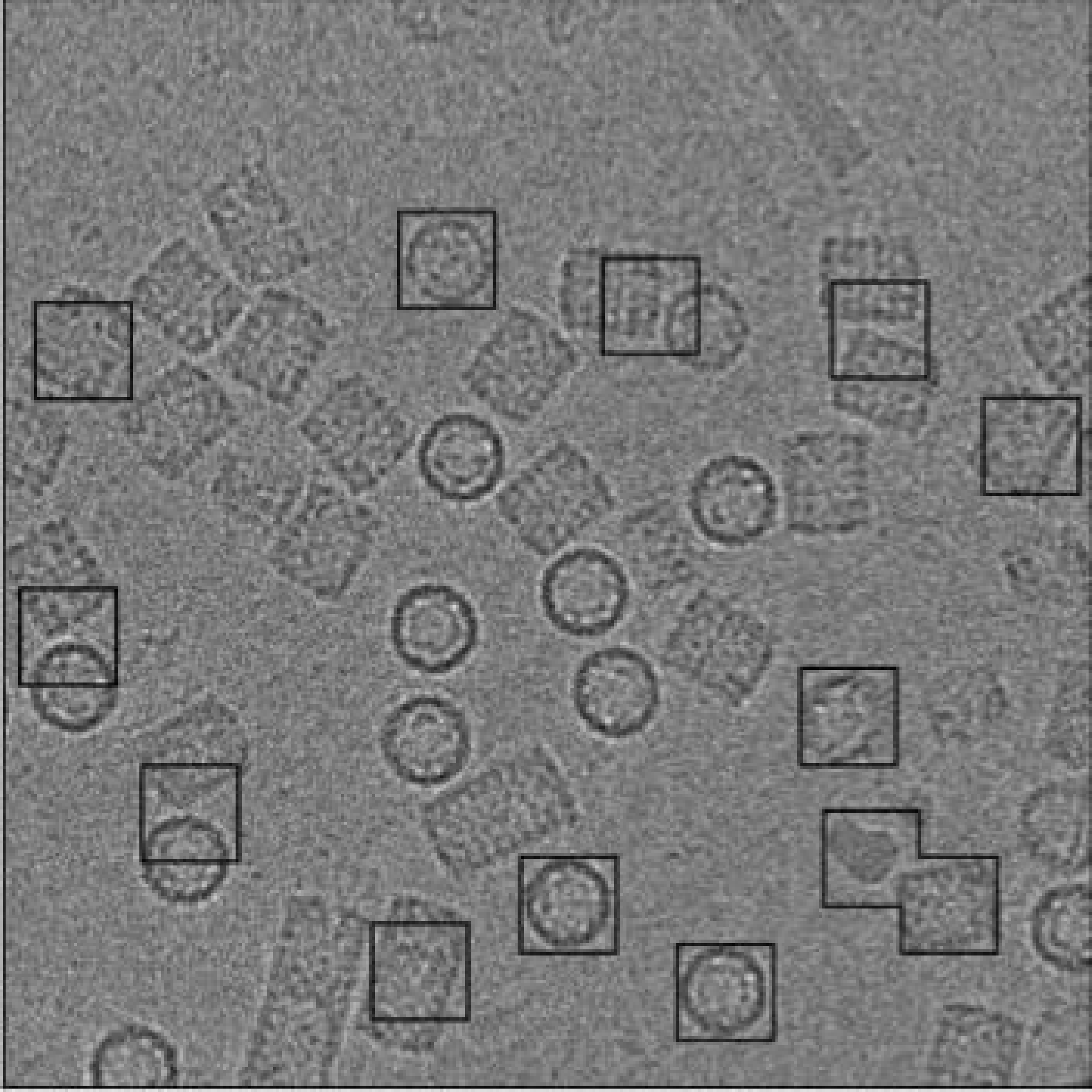}}
{\includegraphics[width=0.3\linewidth]{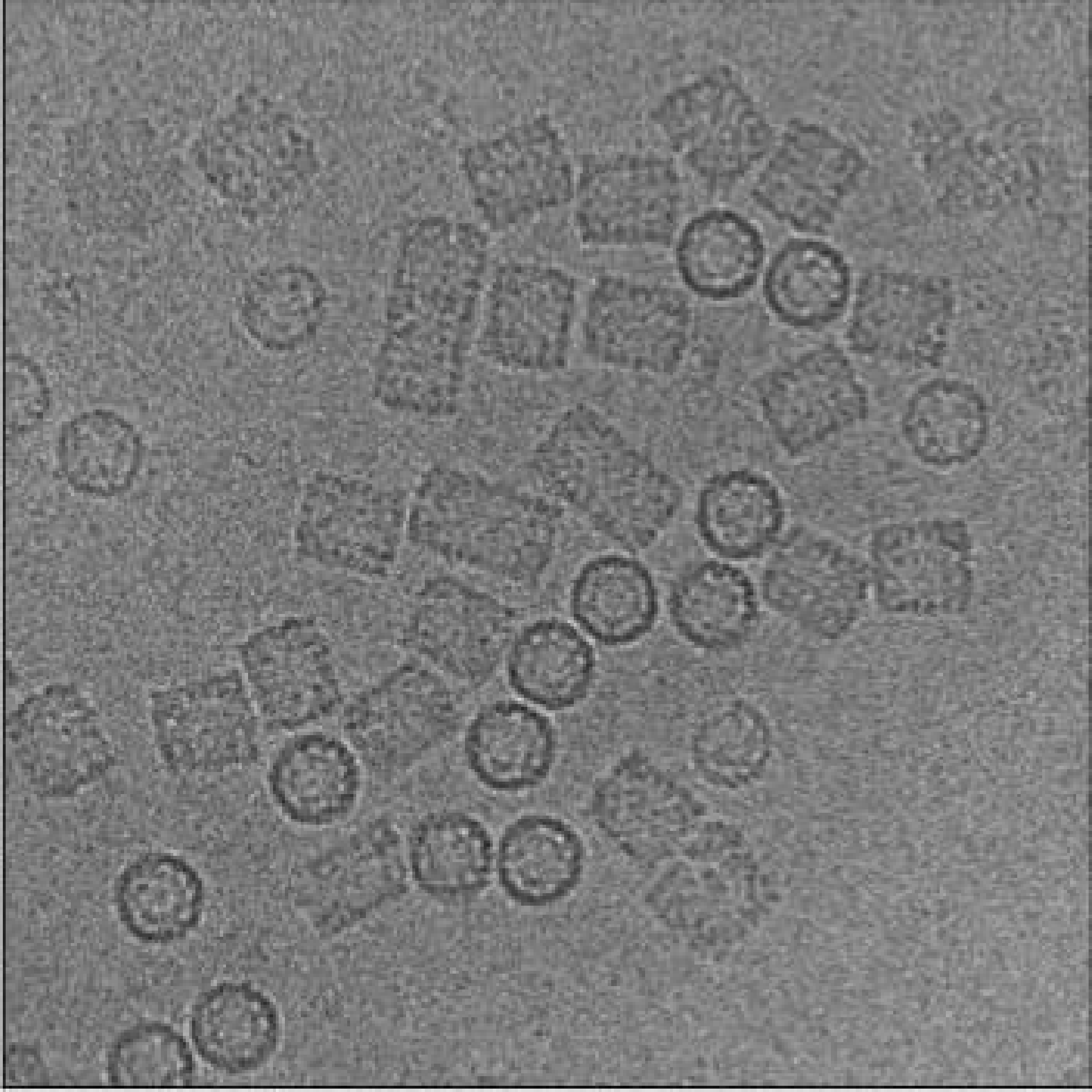}}
{\includegraphics[width=0.3\linewidth]{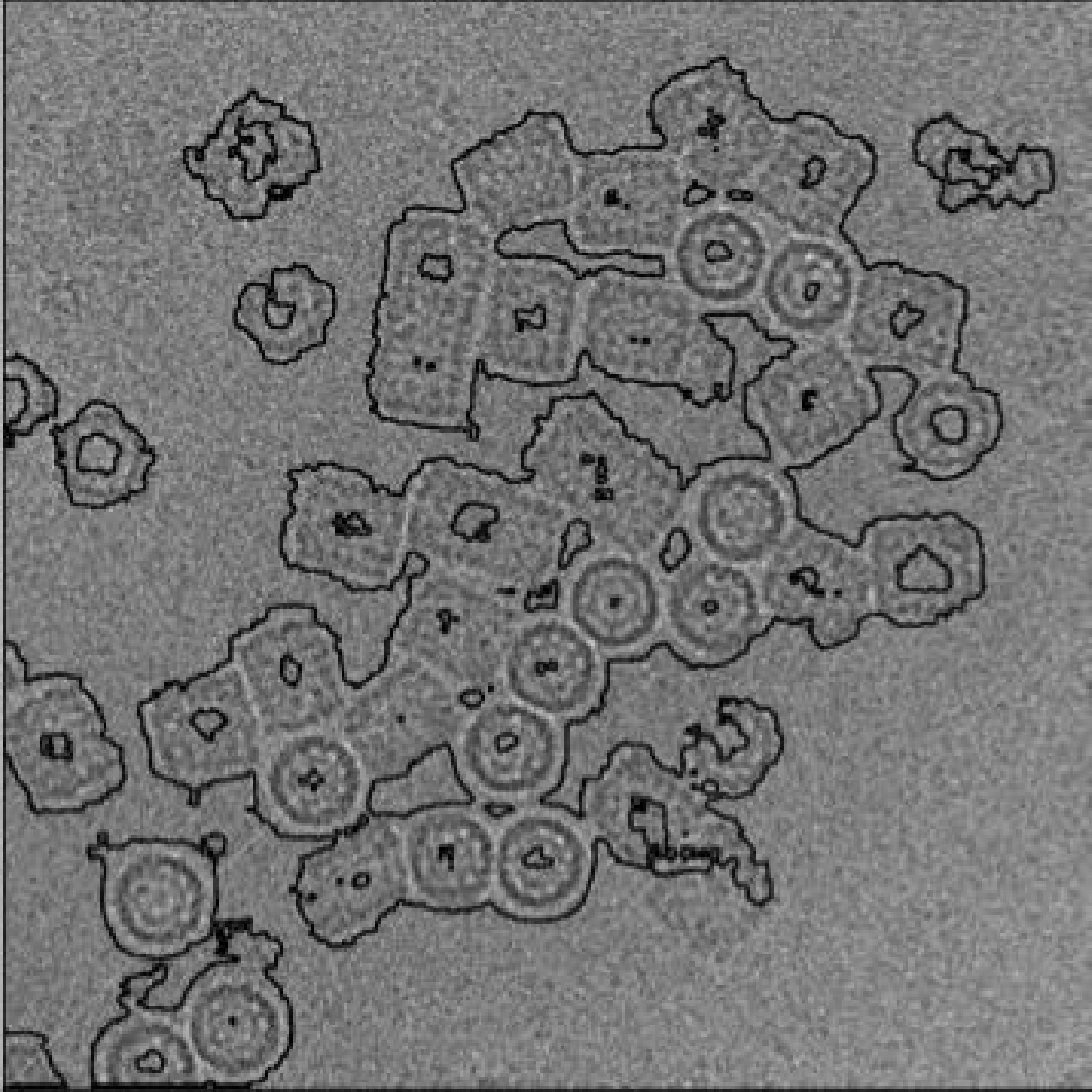}}
{\includegraphics[width=0.3\linewidth]{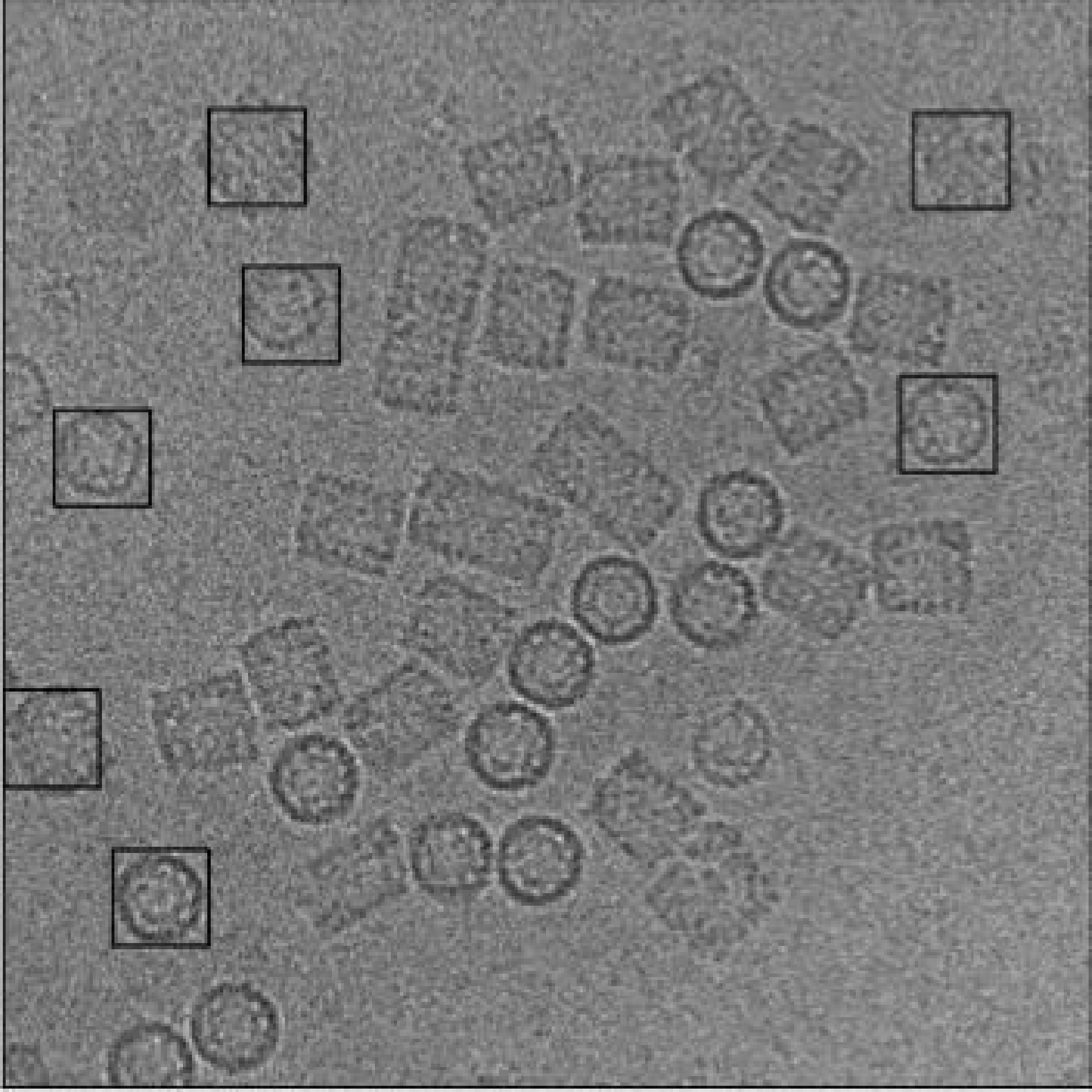}}
{\includegraphics[width=0.3\linewidth]{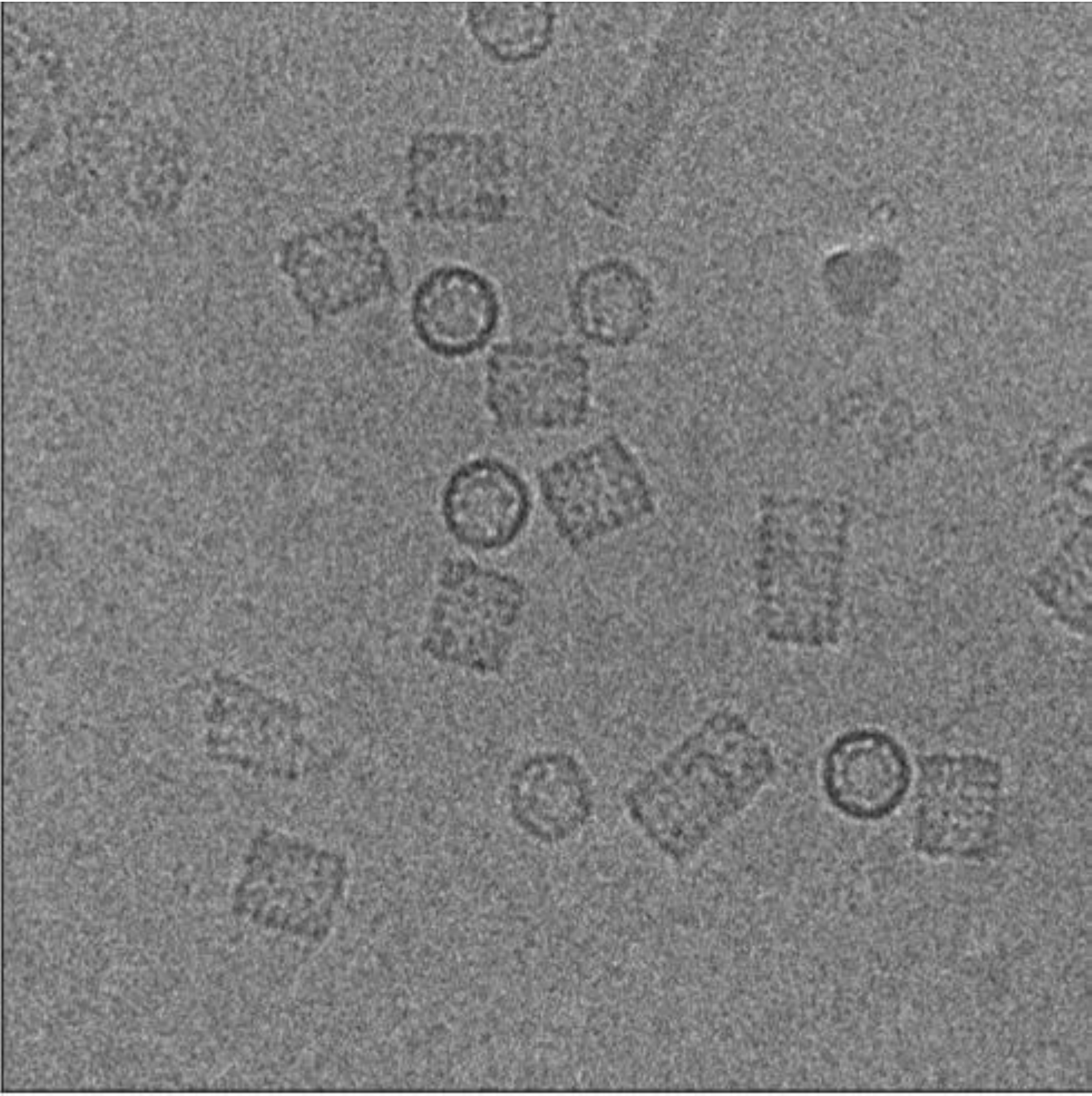}}
{\includegraphics[width=0.3\linewidth]{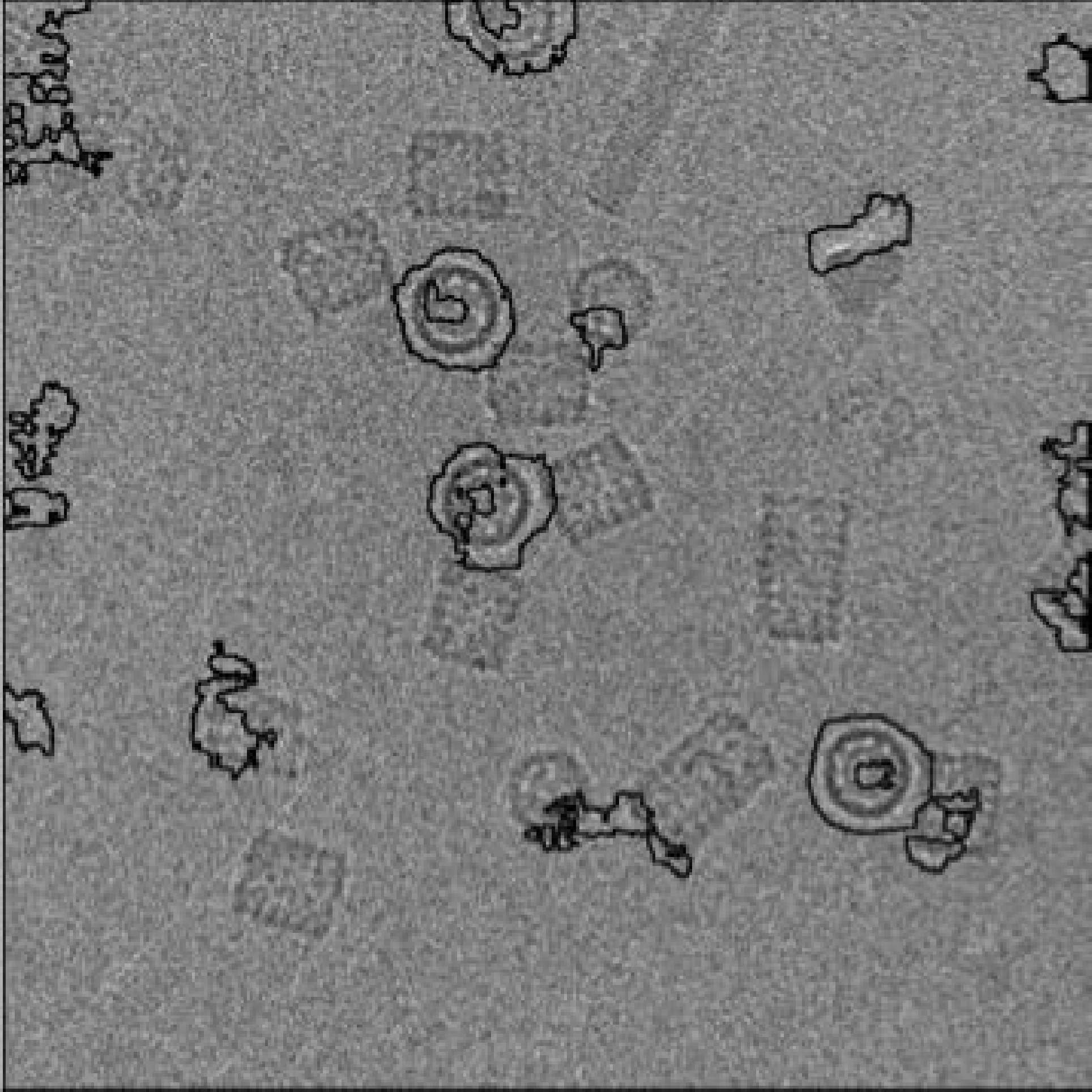}}
{\includegraphics[width=0.3\linewidth]{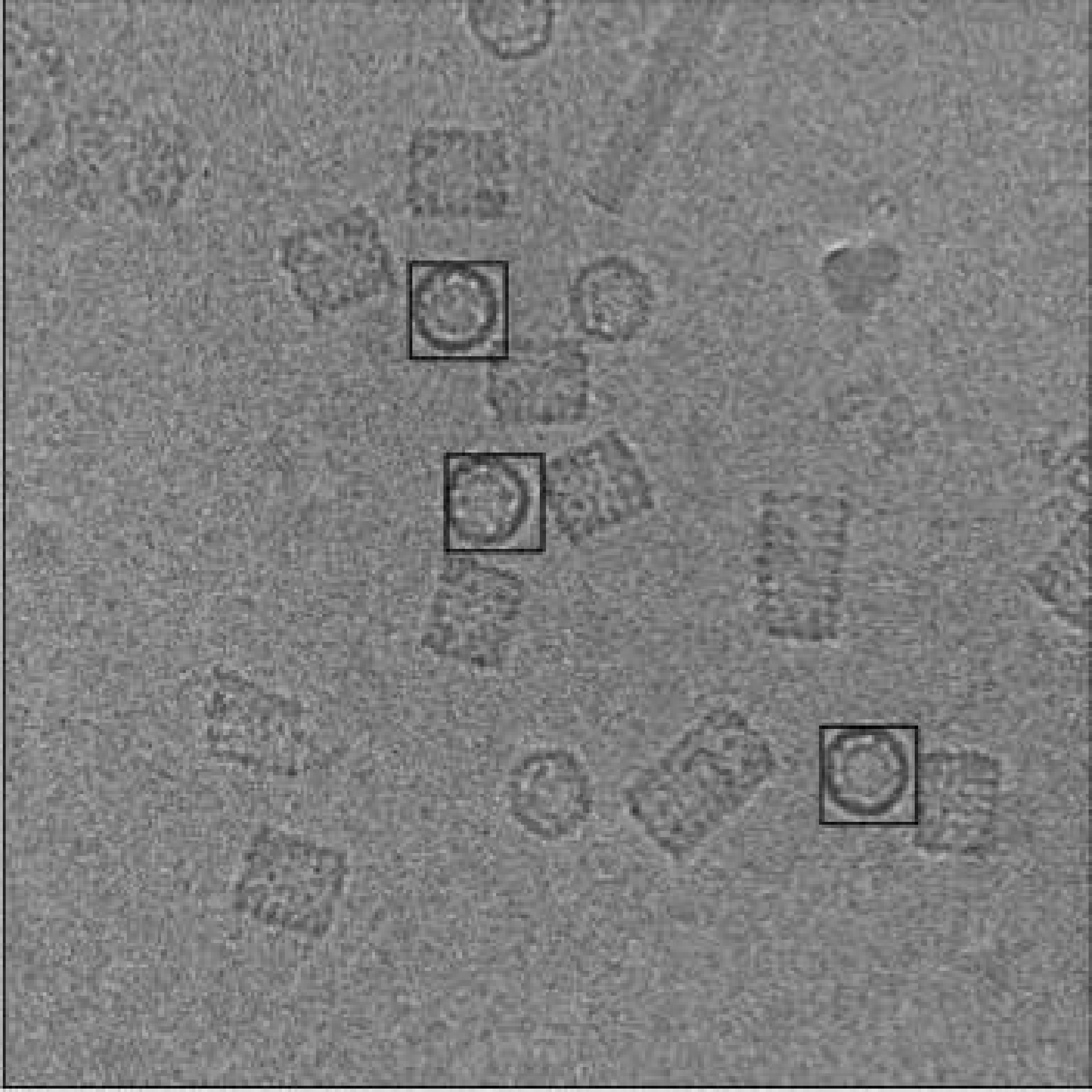}}
\caption{Result on KLH dataset. The left column contains micrographs. The middle column contains the windows classified as
particle by our classifier. The right column contains the picked particles.}
 \label{fig:klh2}
\end{figure*}

We present in Figures  \ref{fig:klh1} and \ref{fig:klh2} some results for the APPLE picker on the KLH dataset. 
We note these figures  
show two types of isoforms of KLH. These isoforms are identified in 
\citep{findEM} as KLH1 (short particles) and KLH2 (long particles). 
We aim to find the KLH1 particles.

As detailed in Section \ref{subsubsec:class}, we use only mean and variance for classifier training. 
An issue with this practice is exemplified by the hollow KLH particles. 
A window containing some regions of the particle and   
some regions of noise that are internal to the hollow particle is 
indistinguishable from a window containing some regions of the particle and 
some regions of noise that are external to the particle.  This leads the classifier to
identify a ring of pixels around the particle as belonging to the particle. 
Depending on the concentration of particles in the micrograph, 
particles may merge together in the output of the classifier. 

We use  morphological erosion to address this  problem. 
This process, detailed in Section \ref{subsubsec:pick}, 
will discard all connected components with maximum diameter 
smaller than $132$ pixels  and larger than $184$ pixels (where the diameter of the KLH particles are approximately $160$ pixels).  
In addition, it will separate adjacent particles connected by a narrow band of pixels. This practice is useful 
in cases where particle projections are close enough that the rings of pixels around each particle 
will merge, but distant enough that the merging is restricted to a narrow region between the particles.

Figure \ref{fig:klh1} contains micrographs where the particles are either completely isolated or distant enough  
that the morphological erosion can separate the pixels that were identified as belonging to each of the particles. 
This is the case in which the APPLE picker is successful despite the hollow particles. Figure \ref{fig:klh2} 
contains micrographs where the particles are clustered closely together, causing the APPLE picker to treat many particles  
as a single region and thus discard them.  It is clear that the APPLE picker is not suited to pick hollow particles that appear 
with a high concentration. We leave it to future work to solve this issue through addition of more discriminative features to the SVM classifier.

Another issue with the KLH dataset is that different micrographs have vastly different concentrations 
of particles. This makes it difficult to select a single value of $\tau_1$ that works well on all the micrographs. An example of this 
is shown in the last row of Figure \ref{fig:klh2}. When using $\tau_1=5\%$ the APPLE picker performs well on this micrograph. 

Our suggested framework processes all $82$ micrographs in under $30$ minutes on the CPU and in $4$ minutes on the GPU.

\section{Discussion}
\subsection{A Comparison Between the APPLE picker and Existing Particle Pickers}
Automatic particle pickers can be divided into two groups  \citep{DoGPicker}. The first of these groups consists of  
methods that assume templates of the particle are known a priori. This knowledge may exist due to 
user provided information, projections from some predetermined initial model, \textit{etc}. The second group imposes mathematical assumptions 
on the particle.

Obtaining user-provided templates is high in user effort. RELION \citep{relion}, for example, necessitates a user to choose 1000--2000 particle 
projections from several micrographs. This process is costly in both effort and time. On the other hand, using 
some initial model may bias the particle picking process \citep{pitfall}. 

Imposing mathematical assumptions on the particle can produce good results so long as the 
assumptions are satisfied by the particle. In contrast, 
the APPLE picker does not make assumptions on the particle other than using the 
well-established fact that projection images and  noise regions 
differ in their mean intensity and variance.

Another advantage of the APPLE picker is that its reference set  contains redundancy. 
This adds a robustness to false positives that is missing from traditional cross-correlation methods.  

Thus, the APPLE picker is a simple, robust and fast particle picker which requires low user effort and  assumes no prior knowledge of the particle 
other than its size. 

We note that there is an assumption on the \textit{artifacts} that may be violated, namely the size assumption. If this assumption is violated, 
regions containing artifacts can be mistaken for particle projections. However, this can easily be corrected in the 2D classification step.

\subsection{Selection of $\tau_1$ and $\tau_2$}

In Section \ref{sec:experimental} we present several datasets with different values of  $\tau_1$ and $\tau_2$. For the $\beta$-Galactosidase and 
T20S proteasome datasets we use the same values. In this section we explain the difference in values from the 70S ribosome and KLH dataset.

The value of $\tau_1$ determines the percentage of  query images that we believe  contain a particle. 
While this value is  different between the datasets, the actual number of query images determined by $\tau_1$ is similar for 
all datasets, and around  $500-800$. The difference is that 
the 70S ribosome dataset uses larger query images which causes each micrograph to contain less of them. The KLH micrographs are  
 much smaller than the micrographs of the other datasets and thus, once again, each micrograph contains less query images. 

\begin{figure}[H]
\centering
{\includegraphics[width=0.4\linewidth]{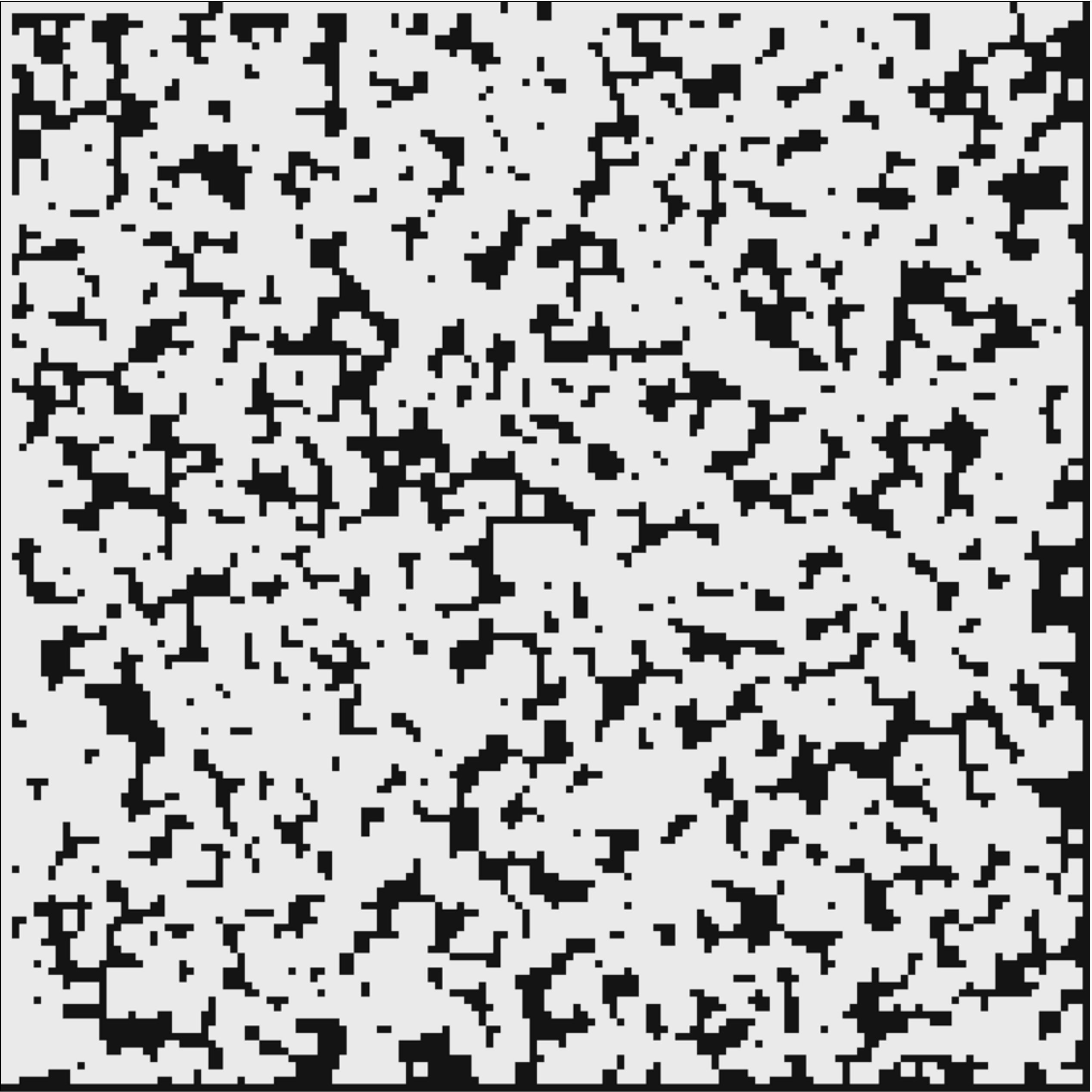}} \hspace{0.4in}
{\includegraphics[width=0.4\linewidth]{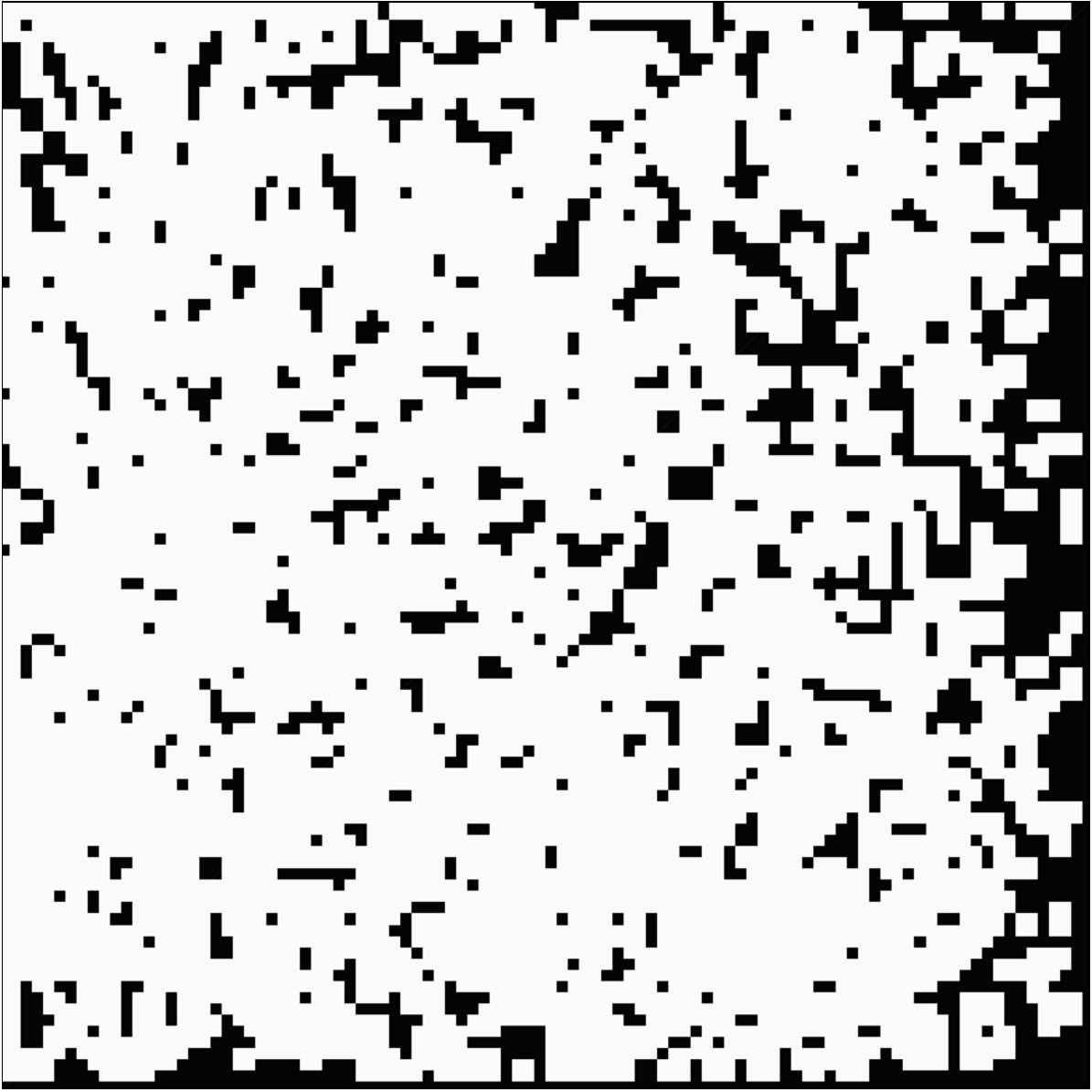}}
\caption{{Selection of $\tau_2$ for a $\beta$-Galactosidase sample micrograph (left) and a 70S ribosome sample micrograph (right). 
The white regions contain the $50\%$ of query images with the highest $k \left( \mathbf{s}_g \right)$. The black regions are 
the regions from which the training windows of noise are extracted. We note that while the $\beta$-Galactosidase sample  
will have plenty of training windows for noise, the 70S ribosome sample will not.}}
 \label{fig:last}
\end{figure}

The value of $\tau_2$ is a different matter. Where the query images are large, they tend to cover more of the micrograph.\footnote{Especially 
since there is overlap between the query images, as detailed in Section \ref{subsubsec:class}.}  
This may not leave many areas large enough to extract training windows of noise. Thus, we must use smaller values of $\tau_2$. 
{An example of this is presented in Figure \ref{fig:last} which shows (in white) the locations of the $50\%$ of windows that possess the higher 
values of $k \left( \cdot \right)$ for the $\beta$-Galactosidase and for the 70S ribosome datasets. }

In conclusion, 
for micrographs of size $4k \times 4k$ where particles are small and their concentration is similar to 
that of the micrographs we presented in Section \ref{sec:experimental}, 
we suggest using $\tau_2 = 50\%-55\%$. For larger particles we suggest using $\tau_1 \approx \tau_2$. 
{In the future, the APPLE picker's code will automatically lower $\tau_2$ until a minimal amount of  noise training windows are extracted, in which 
case this issue will no longer be a consideration for the user.}

\section{Conclusion}
In this paper we have presented  the APPLE picker, a simple and fast framework, inspired by template matching, for fully automated particle picking. 
The APPLE picker has two main classification steps. The first step determines the content of query images 
according to their response to a set of automatically chosen references.  
These results are used to train a simple 
classifier. We presented  experimental results on
four datasets, and showed the type of particles for which this framework is well suited and the reason our classifier may encounter difficulty. 
We leave it to future work to solve these issues.  
We believe that the APPLE picker brings us one step closer towards a fully automated computational pipeline for high throughput single particle analysis using cryo-EM \citep{philip}.

\section*{Acknowledgments}
The authors were partially supported by Award Number R01GM090200 from the
NIGMS, BSF Grant No. 2014401, FA9550-17-1-0291 from AFOSR, Simons Investigator Award and
Simons Collaboration on Algorithms and Geometry from Simons Foundation,
and the Moore Foundation Data-Driven Discovery Investigator Award.

The authors would  like to thank Fred Sigworth for his invaluable conversations and insight into the particle picking problem 
and for his review of the manuscripts and subsequent and helpful suggestions. The authors are also indebted to Philip R. Baldwin for sharing his expertise  
on the particle picking problem as well as his review of the APPLE picker code.

Molecular graphics and analyses were performed with the UCSF Chimera package. Chimera is developed by the Resource for Biocomputing, Visualization, and Informatics at the University of California, San Francisco (supported by NIGMS P41-GM103311).

This research was carried out, in part, while the 
first author was a visiting student research collaborator and the second author 
was a postdoctoral research associate at the Program for Applied and Computational Mathematics 
at Princeton University.

\FloatBarrier
\bibliography{particlePicking}

\begin{thebibliography}{40}
\expandafter\ifx\csname natexlab\endcsname\relax\def\natexlab#1{#1}\fi
\providecommand{\url}[1]{\texttt{#1}}
\providecommand{\href}[2]{#2}
\providecommand{\path}[1]{#1}
\providecommand{\DOIprefix}{doi:}
\providecommand{\ArXivprefix}{arXiv:}
\providecommand{\URLprefix}{URL: }
\providecommand{\Pubmedprefix}{pmid:}
\providecommand{\doi}[1]{\href{http://dx.doi.org/#1}{\path{#1}}}
\providecommand{\Pubmed}[1]{\href{pmid:#1}{\path{#1}}}
\providecommand{\bibinfo}[2]{#2}
\ifx\xfnm\relax \def\xfnm[#1]{\unskip,\space#1}\fi
\bibitem[{Aebel\'{a}ez et~al.(2011)Aebel\'{a}ez, Han, Typke, {L}im, Glaeser \&
  Malik}]{textons}
\bibinfo{author}{Aebel\'{a}ez, P.}, \bibinfo{author}{Han, B.-G.},
  \bibinfo{author}{Typke, D.}, \bibinfo{author}{{L}im, J.},
  \bibinfo{author}{Glaeser, R.~M.}, \& \bibinfo{author}{Malik, J.}
  (\bibinfo{year}{2011}).
\newblock \bibinfo{title}{Experimental evaluation of support vector
  machine-based and correlation-based approaches to automatic particle
  selection}.
\newblock {\it \bibinfo{journal}{Journal of Structural Biology}\/},  {\it
  \bibinfo{volume}{175}\/}, \bibinfo{pages}{319--328}.
\bibitem[{Baldwin et~al.(2018)Baldwin, Tan, Eng, Rice, Noble, Negro,
  Cianfrocco, Potter \& Carragher}]{philip}
\bibinfo{author}{Baldwin, P.~R.}, \bibinfo{author}{Tan, Y.~Z.},
  \bibinfo{author}{Eng, E.~T.}, \bibinfo{author}{Rice, W.~J.},
  \bibinfo{author}{Noble, A.~J.}, \bibinfo{author}{Negro, C.~J.},
  \bibinfo{author}{Cianfrocco, M.~A.}, \bibinfo{author}{Potter, C.~S.}, \&
  \bibinfo{author}{Carragher, B.} (\bibinfo{year}{2018}).
\newblock \bibinfo{title}{Big data in cryo{EM}: automated collection,
  processing and accessibility of {EM} data}.
\newblock {\it \bibinfo{journal}{Current Opinion in Microbiology}\/},  {\it
  \bibinfo{volume}{43}\/}, \bibinfo{pages}{1--8}.
\bibitem[{Bartesaghi et~al.(2015)Bartesaghi, Merk, Banerjee, Matthies, Wu,
  Milne \& Subramanian}]{betaGal2}
\bibinfo{author}{Bartesaghi, A.}, \bibinfo{author}{Merk, A.},
  \bibinfo{author}{Banerjee, S.}, \bibinfo{author}{Matthies, D.},
  \bibinfo{author}{Wu, X.}, \bibinfo{author}{Milne, J. L.~S.}, \&
  \bibinfo{author}{Subramanian, S.} (\bibinfo{year}{2015}).
\newblock \bibinfo{title}{{$2.2$ {\r A}} resolution cryo-{EM} structure of
  $\beta$-galactosidase in complex with a cell-permeant inhinitor}.
\newblock {\it \bibinfo{journal}{Science}\/},  {\it \bibinfo{volume}{348}\/},
  \bibinfo{pages}{1147--1151}.
\bibitem[{Chen \& Grigorieff(2007)}]{signature}
\bibinfo{author}{Chen, J.~Z.}, \& \bibinfo{author}{Grigorieff, N.}
  (\bibinfo{year}{2007}).
\newblock \bibinfo{title}{{SIGNATURE}: A single-particle selection system for
  molecular electron microscopy}.
\newblock {\it \bibinfo{journal}{Journal of Structural Biology}\/},  {\it
  \bibinfo{volume}{157}\/}, \bibinfo{pages}{168--173}.
\bibitem[{Chen et~al.(2013)Chen, McMullan, Faruqi, Murshudov, Short, Scheres \&
  Henderson}]{scheres1}
\bibinfo{author}{Chen, S.}, \bibinfo{author}{McMullan, G.},
  \bibinfo{author}{Faruqi, A.~R.}, \bibinfo{author}{Murshudov, G.~N.},
  \bibinfo{author}{Short, J.~M.}, \bibinfo{author}{Scheres, S.~H.}, \&
  \bibinfo{author}{Henderson, R.} (\bibinfo{year}{2013}).
\newblock \bibinfo{title}{High-resolution noise substitution to measure
  overfitting and validate resolution in 3{D} structure determination by single
  particle electron cryomicroscopy}.
\newblock {\it \bibinfo{journal}{Ultramicroscopy}\/},  {\it
  \bibinfo{volume}{135}\/}, \bibinfo{pages}{24--35}.
\bibitem[{Cortes \& Vapnik(1995)}]{svm7}
\bibinfo{author}{Cortes, C.}, \& \bibinfo{author}{Vapnik, V.}
  (\bibinfo{year}{1995}).
\newblock \bibinfo{title}{Support-vector networks}.
\newblock {\it \bibinfo{journal}{Machine Learning}\/},  {\it
  \bibinfo{volume}{20}\/}, \bibinfo{pages}{273--297}.
\bibitem[{Danev \& Baumeister(2016)}]{e10057}
\bibinfo{author}{Danev, R.}, \& \bibinfo{author}{Baumeister, W.}
  (\bibinfo{year}{2016}).
\newblock \bibinfo{title}{Cryo-{EM} single particle analysis with the volta
  phase plate}.
\newblock {\it \bibinfo{journal}{eLife}\/},  {\it
  \bibinfo{volume}{5:e13046}\/}.
\bibitem[{Downing \& Glaeser(2008)}]{downing_ctf}
\bibinfo{author}{Downing, K.~H.}, \& \bibinfo{author}{Glaeser, R.~M.}
  (\bibinfo{year}{2008}).
\newblock \bibinfo{title}{Restoration of weak phase-contrast images recorded
  with a high degree of defocus: The ``twin image" problem associated with
  {CTF} correction}.
\newblock {\it \bibinfo{journal}{Ultramicroscopy}\/},  {\it
  \bibinfo{volume}{108}\/}, \bibinfo{pages}{921--928}.
\bibitem[{Efford(2000)}]{erosion}
\bibinfo{author}{Efford, N.} (\bibinfo{year}{2000}).
\newblock {\it \bibinfo{title}{Digital Image Processing: A Practical
  Introduction Using Java}\/}.
\newblock (\bibinfo{edition}{1st} ed.).
\newblock \bibinfo{address}{Boston, MA, USA}:
  \bibinfo{publisher}{Addison-Wesley Longman Publishing Co., Inc.}
\bibitem[{Fischer et~al.(2016)Fischer, Neumann, Bock, Maracci, Wang, Paleskava,
  Konevega, Schr\"{o}der, Grubm\"{u}ller, Ficner, Rodnina \& Stark}]{n10077}
\bibinfo{author}{Fischer, N.}, \bibinfo{author}{Neumann, P.},
  \bibinfo{author}{Bock, L.~V.}, \bibinfo{author}{Maracci, C.},
  \bibinfo{author}{Wang, Z.}, \bibinfo{author}{Paleskava, A.},
  \bibinfo{author}{Konevega, A.~L.}, \bibinfo{author}{Schr\"{o}der, G.~F.},
  \bibinfo{author}{Grubm\"{u}ller, H.}, \bibinfo{author}{Ficner, R.},
  \bibinfo{author}{Rodnina, M.}, \& \bibinfo{author}{Stark, H.}
  (\bibinfo{year}{2016}).
\newblock \bibinfo{title}{The pathway to {GTP}ase activation of elongation
  factor selb on the ribosome}.
\newblock {\it \bibinfo{journal}{Nature}\/},  {\it \bibinfo{volume}{540}\/},
  \bibinfo{pages}{80--85}.
\bibitem[{Frank \& Wagenknecht(1983)}]{azimut}
\bibinfo{author}{Frank, J.}, \& \bibinfo{author}{Wagenknecht, T.}
  (\bibinfo{year}{1983}).
\newblock \bibinfo{title}{Automatic selection of molecular images from electron
  micrographs}.
\newblock {\it \bibinfo{journal}{Ultramicroscopy}\/},  {\it
  \bibinfo{volume}{12}\/}, \bibinfo{pages}{169--175}.
\bibitem[{Grant \& Grigorieff(2015)}]{unblur}
\bibinfo{author}{Grant, T.}, \& \bibinfo{author}{Grigorieff, N.}
  (\bibinfo{year}{2015}).
\newblock \bibinfo{title}{Measuring the optimal exposure for single particle
  cryo-{EM} using a $2.6$ \r{A} reconstruction of rotavirus {VP}6}.
\newblock {\it \bibinfo{journal}{eLife}\/},  {\it
  \bibinfo{volume}{4:e06980}\/}.
\bibitem[{Harauz \& Fong-Lochovsky(1989)}]{edgeDetection}
\bibinfo{author}{Harauz, G.}, \& \bibinfo{author}{Fong-Lochovsky, A.}
  (\bibinfo{year}{1989}).
\newblock \bibinfo{title}{Automatic selection of macromolecules from electron
  micrographs by component labelling and symbolic processing}.
\newblock {\it \bibinfo{journal}{Ultramicroscopy}\/},  {\it
  \bibinfo{volume}{31}\/}, \bibinfo{pages}{333--344}.
\bibitem[{Henderson(2013)}]{pitfall}
\bibinfo{author}{Henderson, R.} (\bibinfo{year}{2013}).
\newblock \bibinfo{title}{Avoiding the pitfalls of single particle
  cryo-electron microscopy: Einstein from noise}.
\newblock {\it \bibinfo{journal}{Proceedings of the National Academy of
  Sciences of the United States of America}\/},  {\it \bibinfo{volume}{110}\/},
  \bibinfo{pages}{18037--18041}.
\bibitem[{Hoang et~al.(2013)Hoang, Cavin, Schultz \& Ritchie}]{gEMpicker}
\bibinfo{author}{Hoang, T.~V.}, \bibinfo{author}{Cavin, X.},
  \bibinfo{author}{Schultz, P.}, \& \bibinfo{author}{Ritchie, D.~W.}
  (\bibinfo{year}{2013}).
\newblock \bibinfo{title}{g{EM}picker: a highly parallel {GPU}-accelerated
  particle picking tool for cryo-electron microscopy}.
\newblock {\it \bibinfo{journal}{BMC Structural Biology}\/},  {\it
  \bibinfo{volume}{13}\/}, \bibinfo{pages}{25}.
\bibitem[{Iudin et~al.(2016)Iudin, Korir, Salavert-Torres, Kleywegt \&
  Patwardhan}]{empire}
\bibinfo{author}{Iudin, A.}, \bibinfo{author}{Korir, P.},
  \bibinfo{author}{Salavert-Torres, J.}, \bibinfo{author}{Kleywegt, G.}, \&
  \bibinfo{author}{Patwardhan, A.} (\bibinfo{year}{2016}).
\newblock \bibinfo{title}{{EMPIAR}: A public archive for raw electron
  microscopy image data}.
\newblock {\it \bibinfo{journal}{Nature Methods}\/},  {\it
  \bibinfo{volume}{13}\/}.
\bibitem[{Langlois et~al.(2014)Langlois, Pallesen, Ash, Ho, Rubinstein \&
  Frank}]{templateDisk}
\bibinfo{author}{Langlois, R.}, \bibinfo{author}{Pallesen, J.},
  \bibinfo{author}{Ash, J.~T.}, \bibinfo{author}{Ho, D.~N.},
  \bibinfo{author}{Rubinstein, J.~L.}, \& \bibinfo{author}{Frank, J.}
  (\bibinfo{year}{2014}).
\newblock \bibinfo{title}{Automated particle picking for low-contrast
  macromolecules in cryo-electron microscopy}.
\newblock {\it \bibinfo{journal}{Journal of Structural Biology}\/},  {\it
  \bibinfo{volume}{186}\/}, \bibinfo{pages}{1--7}.
\bibitem[{Ludtke et~al.(1999)Ludtke, Baldwin \& Chiu}]{eman}
\bibinfo{author}{Ludtke, S.~J.}, \bibinfo{author}{Baldwin, P.~R.}, \&
  \bibinfo{author}{Chiu, W.} (\bibinfo{year}{1999}).
\newblock \bibinfo{title}{{EMAN}: Semiautomated software for high-resolution
  single-particle reconstructions}.
\newblock {\it \bibinfo{journal}{Journal of Structural Biology}\/},  {\it
  \bibinfo{volume}{128}\/}, \bibinfo{pages}{82--97}.
\bibitem[{McMullan et~al.(2014)McMullan, Faruqi, Clare \&
  Henderson}]{detectors}
\bibinfo{author}{McMullan, G.}, \bibinfo{author}{Faruqi, A.~R.},
  \bibinfo{author}{Clare, D.}, \& \bibinfo{author}{Henderson, R.}
  (\bibinfo{year}{2014}).
\newblock \bibinfo{title}{Comparison of optimal performance at 300 ke{V} of
  three direct electron detectors for use in low dose electron microscopy}.
\newblock {\it \bibinfo{journal}{Ultramicroscopy}\/},  {\it
  \bibinfo{volume}{147}\/}.
\bibitem[{Mindell \& Grigorieff(2003)}]{ctf_paper}
\bibinfo{author}{Mindell, J.~A.}, \& \bibinfo{author}{Grigorieff, N.}
  (\bibinfo{year}{2003}).
\newblock \bibinfo{title}{Accurate determination of local defocus and specimen
  tilt in electron microscopy}.
\newblock {\it \bibinfo{journal}{Journal of Structural Biology}\/},  {\it
  \bibinfo{volume}{142}\/}, \bibinfo{pages}{334--347}.
\bibitem[{Nicholson \& Glaeser(2001)}]{review}
\bibinfo{author}{Nicholson, W.~V.}, \& \bibinfo{author}{Glaeser, R.~M.}
  (\bibinfo{year}{2001}).
\newblock \bibinfo{title}{Review: Automatic particle detection in electron
  microscopy}.
\newblock {\it \bibinfo{journal}{Journal of Structural Biology}\/},  {\it
  \bibinfo{volume}{133}\/}, \bibinfo{pages}{90--101}.
\bibitem[{Ogura \& Sato(2004)}]{deep04}
\bibinfo{author}{Ogura, T.}, \& \bibinfo{author}{Sato, C.}
  (\bibinfo{year}{2004}).
\newblock \bibinfo{title}{Automatic particle pickup method using a neural
  network has high accuracy by applying an initial weight derived from
  eigenimages: {A} new reference free method for single-particle analysis}.
\newblock {\it \bibinfo{journal}{Journal of Structural Biology}\/},  {\it
  \bibinfo{volume}{145}\/}, \bibinfo{pages}{63--75}.
\bibitem[{Pettersen et~al.(2004)Pettersen, Goddard, Huang, Couch, Greenblatt,
  Meng \& Ferrin}]{chimera1}
\bibinfo{author}{Pettersen, E.}, \bibinfo{author}{Goddard, T.},
  \bibinfo{author}{Huang, C.}, \bibinfo{author}{Couch, G.},
  \bibinfo{author}{Greenblatt, D.}, \bibinfo{author}{Meng, E.}, \&
  \bibinfo{author}{Ferrin, T.} (\bibinfo{year}{2004}).
\newblock \bibinfo{title}{{UCSF} chimera -- a visualization system for
  exploratory research and analysis}.
\newblock {\it \bibinfo{journal}{Journal of Computational Chemistry}\/},  {\it
  \bibinfo{volume}{25}\/}, \bibinfo{pages}{1605--1612}.
\bibitem[{Rohou \& Grigorieff(2015)}]{ctffind}
\bibinfo{author}{Rohou, A.}, \& \bibinfo{author}{Grigorieff, N.}
  (\bibinfo{year}{2015}).
\newblock \bibinfo{title}{{CTFFIND}4: Fast and accurate defocus estimation from
  electron micrographs}.
\newblock {\it \bibinfo{journal}{Journal of Structural Biology}\/},  {\it
  \bibinfo{volume}{192}\/}, \bibinfo{pages}{216--221}.
\bibitem[{Roseman(2004)}]{findEM}
\bibinfo{author}{Roseman, A.} (\bibinfo{year}{2004}).
\newblock \bibinfo{title}{{F}ind{EM} -- {A} fast, efficient program for
  automatic selection of particles from electron micrographs}.
\newblock {\it \bibinfo{journal}{Journal of Structural Biology}\/},  {\it
  \bibinfo{volume}{145}\/}, \bibinfo{pages}{91--99}.
\bibitem[{Scheres(2012{\natexlab{a}})}]{relion3}
\bibinfo{author}{Scheres, S.~H.} (\bibinfo{year}{2012}{\natexlab{a}}).
\newblock \bibinfo{title}{A bayesian view on cryo-{EM} structure
  determination}.
\newblock {\it \bibinfo{journal}{Journal of Molecular Biology}\/},  {\it
  \bibinfo{volume}{415}\/}, \bibinfo{pages}{406--418}.
\bibitem[{Scheres(2012{\natexlab{b}})}]{relion2}
\bibinfo{author}{Scheres, S.~H.} (\bibinfo{year}{2012}{\natexlab{b}}).
\newblock \bibinfo{title}{{RELION}: Implementation of a bayesian approach to
  cryo-{EM} structure determination}.
\newblock {\it \bibinfo{journal}{Journal of Structural Biology}\/},  {\it
  \bibinfo{volume}{180}\/}, \bibinfo{pages}{519--530}.
\bibitem[{Scheres(2015)}]{relion}
\bibinfo{author}{Scheres, S.~H.} (\bibinfo{year}{2015}).
\newblock \bibinfo{title}{Semi-automated selection of cryo-{EM} particles in
  {RELION}-1.3}.
\newblock {\it \bibinfo{journal}{Journal of Structural Biology}\/},  {\it
  \bibinfo{volume}{189}\/}, \bibinfo{pages}{114--122}.
\bibitem[{Scheres \& Chen(2012)}]{scheres2}
\bibinfo{author}{Scheres, S.~H.}, \& \bibinfo{author}{Chen, S.}
  (\bibinfo{year}{2012}).
\newblock \bibinfo{title}{Prevention of overfitting in cryo-{EM} structure
  determination}.
\newblock {\it \bibinfo{journal}{Nature Methods}\/},  {\it
  \bibinfo{volume}{9}\/}, \bibinfo{pages}{853--854}.
\bibitem[{Sch\"{o}lkopf \& Smola(2001)}]{svm3}
\bibinfo{author}{Sch\"{o}lkopf, B.}, \& \bibinfo{author}{Smola, A.~J.}
  (\bibinfo{year}{2001}).
\newblock {\it \bibinfo{title}{Learning with Kernels: Support Vector Machines,
  Regularization, Optimization, and Beyond}\/}.
\newblock \bibinfo{address}{Cambridge, MA, USA}: \bibinfo{publisher}{MIT
  Press}.
\bibitem[{Shatsky et~al.(2009)Shatsky, Hall, Brenner \&
  Glaeser}]{einsteinFromNoise}
\bibinfo{author}{Shatsky, M.}, \bibinfo{author}{Hall, R.~J.},
  \bibinfo{author}{Brenner, S.~E.}, \& \bibinfo{author}{Glaeser, R.~M.}
  (\bibinfo{year}{2009}).
\newblock \bibinfo{title}{A method for the alignment of heterogeneous
  macromolecules from electron microscopy}.
\newblock {\it \bibinfo{journal}{Journal of Structural Biology}\/},  {\it
  \bibinfo{volume}{166}\/}, \bibinfo{pages}{67--78}.
\bibitem[{Sigworth(2004)}]{sigworth1}
\bibinfo{author}{Sigworth, F.~J.} (\bibinfo{year}{2004}).
\newblock \bibinfo{title}{Classical detection theory and the cryo-{EM} particle
  selection problem}.
\newblock {\it \bibinfo{journal}{Journal of Structural Biology}\/},  {\it
  \bibinfo{volume}{145}\/}, \bibinfo{pages}{111--122}.
\bibitem[{Turo\v{n}ov\'{a} et~al.(2017)Turo\v{n}ov\'{a}, Schur, Wan \&
  Briggs}]{ctf2_paper}
\bibinfo{author}{Turo\v{n}ov\'{a}, B.}, \bibinfo{author}{Schur, F.~K.},
  \bibinfo{author}{Wan, W.}, \& \bibinfo{author}{Briggs, J.~A.}
  (\bibinfo{year}{2017}).
\newblock \bibinfo{title}{Efficient 3{D-CTF} correction for cryo-electron
  tomography using {N}ova{CTF} improves subtomogram averaging resolution to 3.4
  \r{A}}.
\newblock {\it \bibinfo{journal}{Journal of Structural Biology}\/},  {\it
  \bibinfo{volume}{199}\/}, \bibinfo{pages}{187--195}.
\bibitem[{{van Heel}(1982)}]{VanHeel}
\bibinfo{author}{{van Heel}, M.} (\bibinfo{year}{1982}).
\newblock \bibinfo{title}{Detection of objects in quantum-noise-limited
  images}.
\newblock {\it \bibinfo{journal}{Ultramicroscopy}\/},  {\it
  \bibinfo{volume}{7}\/}, \bibinfo{pages}{331--341}.
\bibitem[{Voss et~al.(2009)Voss, Yoshioka, Radermacher, Potter \&
  Carragher}]{DoGPicker}
\bibinfo{author}{Voss, N.~R.}, \bibinfo{author}{Yoshioka, C.},
  \bibinfo{author}{Radermacher, M.}, \bibinfo{author}{Potter, C.~S.}, \&
  \bibinfo{author}{Carragher, B.} (\bibinfo{year}{2009}).
\newblock \bibinfo{title}{{D}o{G} picker and {T}ilt{P}icker: Software tools to
  facilitate particle selection in single particle electron microscopy}.
\newblock {\it \bibinfo{journal}{Journal of Structural Biology}\/},  {\it
  \bibinfo{volume}{166}\/}, \bibinfo{pages}{205--213}.
\bibitem[{Wang et~al.(2016)Wang, Gong, Liu, Li, Yan, Xia, Li \&
  Zeng}]{deepPickler}
\bibinfo{author}{Wang, F.}, \bibinfo{author}{Gong, H.}, \bibinfo{author}{Liu,
  G.}, \bibinfo{author}{Li, M.}, \bibinfo{author}{Yan, C.},
  \bibinfo{author}{Xia, T.}, \bibinfo{author}{Li, X.}, \&
  \bibinfo{author}{Zeng, J.} (\bibinfo{year}{2016}).
\newblock \bibinfo{title}{{D}eep{P}icker: {A} deep learning approach for fully
  automated particle picking in cryo-{EM}}.
\newblock {\it \bibinfo{journal}{Journal of Structural Biology}\/},  {\it
  \bibinfo{volume}{195}\/}, \bibinfo{pages}{325--336}.
\bibitem[{Zhang(2017)}]{gautomatch}
\bibinfo{author}{Zhang, K.} (\bibinfo{year}{2017}).
\newblock \bibinfo{title}{http://www.mrc-lmb.cam.ac.uk/kzhang/}.
\bibitem[{Zhu et~al.(2004)Zhu, Carragher, Glaeser, Fellmann, Bajaj, Bern,
  Mouche, de~Haas, Hall, Kriegman, Ludtke, Mallick, Penczek, Roseman, Sigworth,
  Volkmann \& Potter}]{bakeoff}
\bibinfo{author}{Zhu, Y.}, \bibinfo{author}{Carragher, B.},
  \bibinfo{author}{Glaeser, R.~M.}, \bibinfo{author}{Fellmann, D.},
  \bibinfo{author}{Bajaj, C.}, \bibinfo{author}{Bern, M.},
  \bibinfo{author}{Mouche, F.}, \bibinfo{author}{de~Haas, F.},
  \bibinfo{author}{Hall, R.~J.}, \bibinfo{author}{Kriegman, D.~J.},
  \bibinfo{author}{Ludtke, S.~J.}, \bibinfo{author}{Mallick, S.~P.},
  \bibinfo{author}{Penczek, P.~A.}, \bibinfo{author}{Roseman, A.~M.},
  \bibinfo{author}{Sigworth, F.~J.}, \bibinfo{author}{Volkmann, N.}, \&
  \bibinfo{author}{Potter, C.~S.} (\bibinfo{year}{2004}).
\newblock \bibinfo{title}{Automatic particle selection: {R}esults of a
  comparative study}.
\newblock {\it \bibinfo{journal}{Journal of Structural Biology}\/},  {\it
  \bibinfo{volume}{145}\/}, \bibinfo{pages}{3--14}.
\bibitem[{Zhu et~al.(2003)Zhu, Carragher, Mouche \& Potter}]{KLH}
\bibinfo{author}{Zhu, Y.}, \bibinfo{author}{Carragher, B.},
  \bibinfo{author}{Mouche, F.}, \& \bibinfo{author}{Potter, C.~S.}
  (\bibinfo{year}{2003}).
\newblock \bibinfo{title}{Automatic particle detection through efficient
  {H}ough transforms}.
\newblock {\it \bibinfo{journal}{IEEE Transactions on Medical Imaging}\/},
  {\it \bibinfo{volume}{22}\/}, \bibinfo{pages}{1053--1062}.
\bibitem[{Zhu et~al.(2016)Zhu, Ouyang \& Mao}]{deep2}
\bibinfo{author}{Zhu, Y.}, \bibinfo{author}{Ouyang, Q.}, \&
  \bibinfo{author}{Mao, Y.} (\bibinfo{year}{2016}).
\newblock \bibinfo{title}{A deep learning approach to single-particle
  recognition in cryo-electron microscopy}.
\newblock {\it \bibinfo{journal}{CoRR}\/},  {\it
  \bibinfo{volume}{abs/1605.05543}\/}.

\end{thebibliography}
\end{multicols}
\end{document}